\documentclass[11pt]{style/prism-princeton}

\makeatletter
\newcommand{\longdash}[1][2em]{%
  \makebox[#1]{$\m@th\smash-\mkern-7mu\cleaders\hbox{$\mkern-2mu\smash-\mkern-2mu$}\hfill\mkern-7mu\smash-$}}
\makeatother
\newcommand{\omitskip}{\kern-\arraycolsep}

\usepackage{booktabs}
\usepackage{caption}
\usepackage{amsmath}
\usepackage{amsfonts}
\usepackage{amssymb}
\usepackage{graphicx}
\usepackage{multirow}
\usepackage{xcolor}
\usepackage{colortbl}      
\usepackage{pifont}
\usepackage{algorithm}
\usepackage{listings}
\usepackage{subcaption}
\usepackage{enumitem}
\usepackage{tabularx}
\usepackage{array}
\usepackage{longtable}
\usepackage{placeins}
\newcolumntype{L}[1]{>{\raggedright\arraybackslash}p{#1}}
\newcolumntype{C}[1]{>{\centering\arraybackslash}p{#1}}
\providecommand{\rel}[1]{\operatorname{#1}}
\providecommand{\checkmark}{\ding{51}}                       

\definecolor{codegray}{rgb}{0.5,0.5,0.5}
\definecolor{codegreen}{rgb}{0,0.5,0}

\lstdefinelanguage{json}{
  morestring=[b]",
  morestring=[d]",
  stringstyle=\color{codegreen},
  literate=
    *{0}{{{\color{blue}0}}}{1}
     {1}{{{\color{blue}1}}}{1}
     {2}{{{\color{blue}2}}}{1}
     {3}{{{\color{blue}3}}}{1}
     {4}{{{\color{blue}4}}}{1}
     {5}{{{\color{blue}5}}}{1}
     {6}{{{\color{blue}6}}}{1}
     {7}{{{\color{blue}7}}}{1}
     {8}{{{\color{blue}8}}}{1}
     {9}{{{\color{blue}9}}}{1}
     {:}{{{\color{black}{:}}}}{1}
     {,}{{{\color{black}{,}}}}{1}
     {\{}{{{\color{black}{\{}}}}{1}
     {\}}{{{\color{black}{\}}}}}{1}
     {[}{{{\color{black}{[}}}}{1}
     {]}{{{\color{black}{]}}}}{1},
}
\lstdefinestyle{tight}{
  basicstyle=\footnotesize\ttfamily,
  keywordstyle=\color{blue},
  commentstyle=\color{codegray},
  stringstyle=\color{codegreen},
  showstringspaces=false,
  breaklines=true,
  frame=single,
  framesep=3pt,
  xleftmargin=4pt,
  aboveskip=4pt,
  belowskip=4pt,
}
\providecommand{\num}[1]{{\color{black}#1}}
\providecommand{\methodname}[1]{\texttt{Pigey}}

\author[1,2]{Liane Galanti}
\author[1]{Dhruv Shah}
\author[1,2]{Tri Dao}

\affiliation[1]{Princeton University}
\affiliation[2]{Together AI}

\begin{document}

\title{Addressing the Orchestration Gap in Generalist Robots via Physical Agency}
\abstract{
General-purpose robots need to reason about their actions, combining perception, world knowledge, planning, success detection, recovery, and low-level control.
Today's state-of-the-art models attempt to combine all these capabilities into the learned policy via large-scale pre-training. Instead, we show that these capabilities can be decomposed into a general language-conditioned policy/control agent and a high-level agent manager/orchestrator.
Rather than training policies to reason via pre-training, we build a closed-loop physical agent orchestrator that can do high-level planning, decompose the goal into achievable subgoals, command low-level motor commands, track and verify the outcome from low-level observations, and recover from failures. 
Our \textbf{P}hys\textbf{i}cal A\textbf{ge}nc\textbf{y} orchestrator (\methodname{}) can control \emph{existing} vision-language-action (VLA) policies as well as parametrized skills to solve complex reasoning tasks in the real world, without any additional data collection or post-training.
We evaluate \methodname{} extensively across simulation benchmarks and challenging real-world robotic manipulation tasks, and demonstrate significant performance improvements over existing generalist policies. On LIBERO-PRO, \methodname{} advances the state-of-the-art by over $4\times$ (12.8\% $\to$ 53.3\%) with no task-specific fine-tuning. On a real robot, \methodname{} lifts the frozen policy from near-zero to over \num{90}\% on reasoning-limited tasks. We call the difference between what frozen motor skills achieve alone and inside the agentic loop the \emph{orchestration gap}.
}



\website{https://lianegalanti.github.io/Pigey/}{lianegalanti.github.io/Pigey}
\code{https://github.com/lianegalanti/Pigey}{github.com/lianegalanti/Pigey}
\maketitle

\begin{figure}[h]
    \centering
    \includegraphics[width=\textwidth]{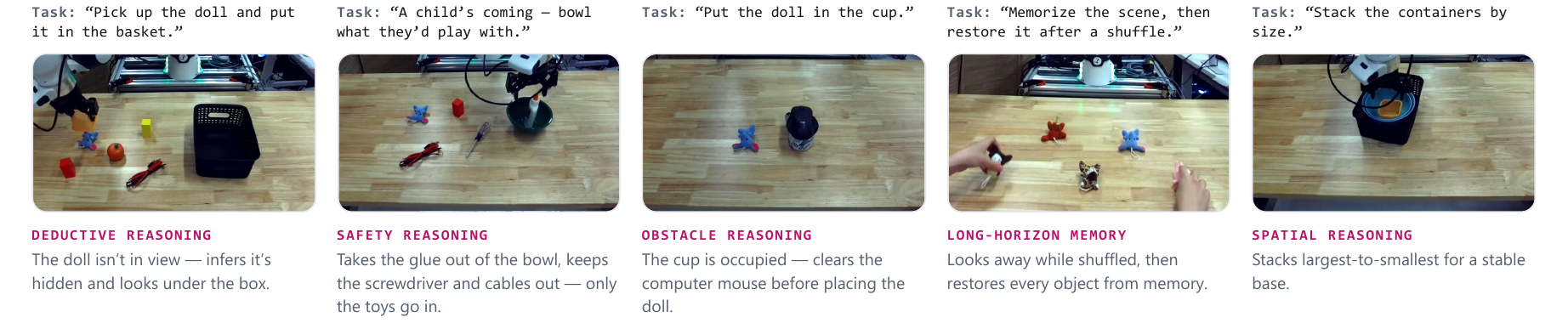}
    \caption{Example reasoning behaviors enabled by \methodname{}: obstacle reasoning (inferring a target is hidden and uncovering it), safety reasoning (identifying and removing hazardous items), clearing an occupied goal (emptying a container before filling it), long-horizon memory (memorizing a scene and restoring it after an occlusion), and spatial reasoning (stacking containers largest-to-smallest for stability). All panels are single, uncut real-robot rollouts; the orchestrator drives frozen skills with no new training.}
    \label{fig:teaser}
\end{figure}

\section{Introduction}
\label{sec:introduction}


Let's imagine instructing your home robot "a child is coming over---put the toys on the plate, and the unsafe items in the box." A capable visuomotor policy might fail: it has to tell a toy from a hazard (world knowledge), find items hidden behind others (perception), recover when a grasp slips (closed-loop control), and stop only once the table is genuinely safe---not when it appears tidy. General-purpose manipulation is a full stack---perception, world knowledge, intent reasoning, planning, success detection, recovery, and low-level control---and the hard part is rarely the motion alone. A system with strong control but weak reasoning cannot interpret abstract goals; a system with strong reasoning but weak control cannot act in the world.

The dominant route to generalization is to scale robot data. VLA policies trained on large demonstration collections~\cite{pi0, pi05, openvla, droid} learn increasingly capable visuomotor skills, and such data is essential: contact, grasping, and embodiment-specific control must be learned from interaction. But robot data is expensive, and it is not always the most direct way to teach task-level capabilities such as negation, progress tracking, and recovery. When a policy fails on ``put everything on the plate except the cup,'' ``set the bowl for a vegetarian,'' or the childproofing request above, the missing ingredient is often not the low-level motion itself, but negation, world knowledge, decomposition, progress tracking, or recognizing that a grasp failed and must be retried.

Prior work addresses parts of this stack but rarely the whole loop. VLA scaling~\cite{pi05, openvla, groot} sharpens control and grounding, yet direct prompting still asks one network to perceive, reason, plan, verify, recover, and act in a single forward pass. Code-as-policies and task planners~\cite{codeaspolicies, progprompt, saycan, voxposer} add structure, but lean on symbolic APIs, hand-built primitives, privileged simulator state, or code-level action spaces. Reasoning-VLAs~\cite{hirobot, ecot} push more deliberation into the policy itself, at the cost of additional training, and still leave success detection and recovery outside the loop. What is missing is not a better motor policy or a single reasoning module, but a \emph{process} that closes the loop---deciding what to do, checking whether it worked, and repairing when it did not.

We propose augmenting these capabilities during inference: decomposing manipulation into a low-level language-conditioned control layer and a high-level agent manager rather than baking everything into learned weights. We instantiate this as \methodname{} (\textbf{P}hys\textbf{i}cal A\textbf{ge}nc\textbf{y}), a closed-loop orchestrator. A frontier VLM plans a short concrete subgoal, selects a frozen backend to execute it, verifies the outcome from the resulting observation, and replans or recovers when verification fails---repeating until the instruction is satisfied. \methodname{} drives two frozen, complementary skills, neither trained for this work: a TAMP grasp planner built on TiPToP~\cite{tiptop} for precise pick-and-place on rigid objects, and a $\pi_{0.5}$ VLA~\cite{pi05} for deformable, contact-rich, and recovery actions; each is handed only a short executable subgoal such as ``put the red cup on the plate,'' never the abstract instruction. The loop is what carries the hard cases. Asked to ``pick up the doll and put it in the basket'' (Figure~\ref{fig:qualitative_rollouts}), \methodname{} sees no doll---only a box; rather than grasp blindly, it infers the doll is underneath, removes the box, re-perceives, finds the doll, and only then picks and places it. The grasp was never the bottleneck---the inference ``it is hidden, uncover it first'' was. That this works at all is not obvious. A general VLM, never trained for embodiment, must ground its reasoning in live camera observations tightly enough to tell a completed grasp from a failed one, a present object from a hidden one, and a finished task from an unfinished one---from general pretraining alone. We find that it can, and that the effect is a property of the loop rather than of any single model: it holds across seven different reasoners.

This isolates where generalization might be lacking. We hold the robot, cameras, scenes, demonstrations, and policy weights fixed, and vary only the inference-time orchestration: in the \emph{direct} condition, the full instruction is passed to the motor policy as one command; in the \emph{agentic} condition, the same frozen skills are reached through \methodname{}. There is no additional data collection or post-training, so changes in behavior isolate the effect of the inference-time loop rather than new motor learning.

The resulting behaviors differ sharply. In identical scenes with identical weights, direct prompting is nearly prompt-invariant---producing almost the same motion whether asked for the vegetarian item, the dangerous one, or ``everything except the cup''---whereas \methodname{} yields distinct, correct behaviors and recovers from failed or mis-grasped objects. Across \num{30} Franka FR3 tasks, the gains land squarely on reasoning-limited tasks (world knowledge, conditionals, multi-step, recovery) and leave already-easy pick-and-place untouched. The same pattern holds in LIBERO-PRO~\cite{liberopro}, where the loop lifts the frozen policy's mean success from \num{12.8}\% to \num{53.3}\% across six perturbation suites---over $4\times$, with no change to policy weights.

Robot data and reasoning thus solve different problems: demonstrations teach a policy \emph{how} to act, while \methodname{} decides \emph{when}, \emph{why}, in \emph{what order}, and with \emph{which} skill to act. Concretely, this paper contributes:
\begin{itemize}[leftmargin=1.4em, itemsep=2pt, topsep=2pt]
    \item \textbf{A full-stack framing of robot generalization}, in which perception, world knowledge, reasoning, planning, success detection, recovery, and control must work together---and failures are attributed to the missing component rather than to ``not enough data.''
    \item \textbf{\methodname{}, a closed-loop inference-time orchestrator} that supplies the task-level stack---planning, skill selection, verification, and recovery---over \emph{frozen} VLA policies and parametrized skills (here, a TAMP grasp planner and a $\pi_{0.5}$ VLA), with no additional data collection or post-training.
    \item \textbf{Demonstrating and bridging the \emph{orchestration gap}}: across \num{30} real-robot tasks and LIBERO-PRO, the same frozen skills succeed far more often inside \methodname{} than when prompted directly, with gains concentrated on reasoning-limited rather than motor-bound failures.
\end{itemize}

\section{Related Work}
\label{sec:related}

\textbf{Scaling robot data.}
VLAs scale imitation learning across larger datasets, embodiments, and tasks. $\pi$0~\cite{pi0} introduced flow matching for continuous control; $\pi_{0.5}$~\cite{pi05} adds heterogeneous co-training; OpenVLA~\cite{openvla}, GR00T~N1~\cite{groot}, and DROID~\cite{droid} push the data-scaling view of general-purpose control. This learns genuinely better \emph{motor} behavior---and remains essential for grounded execution. But data alone is an indirect lever for the rest of the stack: another demonstration of a scene does not teach negation, world knowledge, decomposition, progress tracking, or recovery, and direct prompting forces a single network to perceive, reason, plan, verify, recover, and act in one forward pass. Capabilities that fail for entirely different reasons are collapsed into one objective, so when the policy fails it is unclear what is even missing.

\textbf{Planning and code-as-policy agents.}
A second line adds task-level structure above control. SayCan~\cite{saycan} scores affordances; Code-as-Policies~\cite{codeaspolicies} and ProgPrompt~\cite{progprompt} emit executable programs; Inner Monologue~\cite{innermonologue} folds in feedback; VoxPoser~\cite{voxposer} synthesizes 3D value maps; and embodied coding agents~\cite{capx} add structured feedback and test-time computation. These demonstrate the value of planning and feedback, but many rely on symbolic APIs, hand-designed primitives, privileged simulator state, or code-level action spaces. In contrast, our backends are pixel-conditioned robot skills, and the loop operates through the same observations used for execution. We compare against this line in simulation (CaP-Agent0~\cite{capx}) and show our agent surpasses it precisely because our backends are learned, pixel-conditioned skills closed in a verify-and-recover loop, not hand-written code.

\textbf{Hierarchies, reasoning VLAs, and capability modules.}
A third line wires reasoning into learned policies. Hi~Robot~\cite{hirobot} trains a VLM to decompose tasks for $\pi$0; GR00T~N1~\cite{groot} and Gemini Robotics~\cite{geminirobotics} train dual-system VLM-action architectures; Steerable Policies~\cite{steerable} train on richer command structure; and others bolt on one capability at a time---MemER~\cite{memer} for memory, ECoT~\cite{ecot} for embodied chain-of-thought, RoboMonkey~\cite{robomonkey} for verification, learned reward VLMs~\cite{liang2026robometer, ma2025vision} for scoring. TiPToP~\cite{tiptop} pairs Gemini Robotics-ER grounding with classical TAMP; we build directly on it, but use it as one \emph{frozen} backend the agent can call, not as the whole system. The common cost across this line is the same: each path either requires additional training (more robot data, fine-tuning, a robotics-specialized reasoner) or supplies a single bespoke module, leaving the rest of the closed loop unaddressed.


\textbf{Operational profile.}
%
Prior systems add individual capabilities around a motor policy---\emph{context} (memory of earlier observations), explicit task \emph{state} (what is held, placed, or remaining), \emph{retry} after a detected failure, and outcome \emph{verification}---and many train a dedicated module for each; those modules deliver real benefits we do not seek to replicate. Our claim is narrower: a frozen frontier VLM, used as an orchestrator, supplies many of the same functions at inference time without additional training. Table~\ref{tab:comparison} characterizes what each system requires operationally, rather than scoring them.
\begin{table}[t]
\caption{Operational profile across modular manipulation systems. \emph{Auditability}: whether the decision process is exposed as explicit tool calls, partially inspectable modules, or mostly black-box model computation. \emph{Added training}: training required on top of a base VLA. Entries reflect our reading of each system's published description.}
\label{tab:comparison}
\centering
\small
\begin{tabular}{@{}lccccll@{}}
\toprule
\textbf{System} &
\textbf{Context} &
\textbf{State} &
\textbf{Retry} &
\textbf{Verify} &
\textbf{Auditability} &
\textbf{Added training} \\
\midrule
$\pi_{0.5}$ alone~\cite{pi05}        & -- & -- & -- & -- & Black-box & VLA only \\
Hi Robot~\cite{hirobot}           & -- & -- & -- & Partial & Partial & High-level planner \\
GR00T~N1~\cite{groot}             & -- & -- & -- & -- & Black-box & Joint VLM+VLA \\
Gemini Robotics~\cite{geminirobotics} & -- & Partial & -- & Partial & Black-box & End-to-end VLA \\
TiPToP~\cite{tiptop}              & -- & Partial & -- & Partial & Explicit & None (planner) \\
CaP-Agent0~\cite{capx}            & Partial & Partial & Partial & -- & Explicit & None / code policy \\
ECoT~\cite{ecot}                  & -- & -- & -- & -- & Explicit & CoT into VLA \\
MemER~\cite{memer}                & \checkmark & -- & -- & -- & Partial & Memory module \\
RoboMonkey~\cite{robomonkey}      & -- & -- & \checkmark & \checkmark & Partial & Verifier \\
RoboMeter~\cite{liang2026robometer} & -- & -- & Partial & \checkmark & Partial & Reward critic \\
\rowcolor{magenta!25}\methodname{} (\textbf{Ours})                     & \checkmark & \checkmark & \checkmark & \checkmark & \textbf{Explicit} & \textbf{None} \\
\bottomrule
\end{tabular}
\end{table}

\section{Orchestrating Robots via Physical Agency}
\label{sec:method}

Across prior approaches---scaling robot data, planning and code-as-policy agents, and reasoning-VLAs---one piece stays missing: a complete \emph{closed-loop process} that decides what to do, checks whether it worked, and repairs when it did not.  We add it at inference time, with no new training, as \methodname{}: a frontier VLM that plans, calls a frozen motor backend (a TAMP planner or a $\pi_{0.5}$ VLA), verifies the outcome, and recovers. This is what lets a frozen policy succeed on instructions it cannot follow when prompted directly---the \emph{orchestration gap}.

\subsection{Overview}
\begin{figure}[t]
\centering
\begin{minipage}[t]{0.49\textwidth}
  \centering
  \vspace{0pt}
  \includegraphics[width=\linewidth]{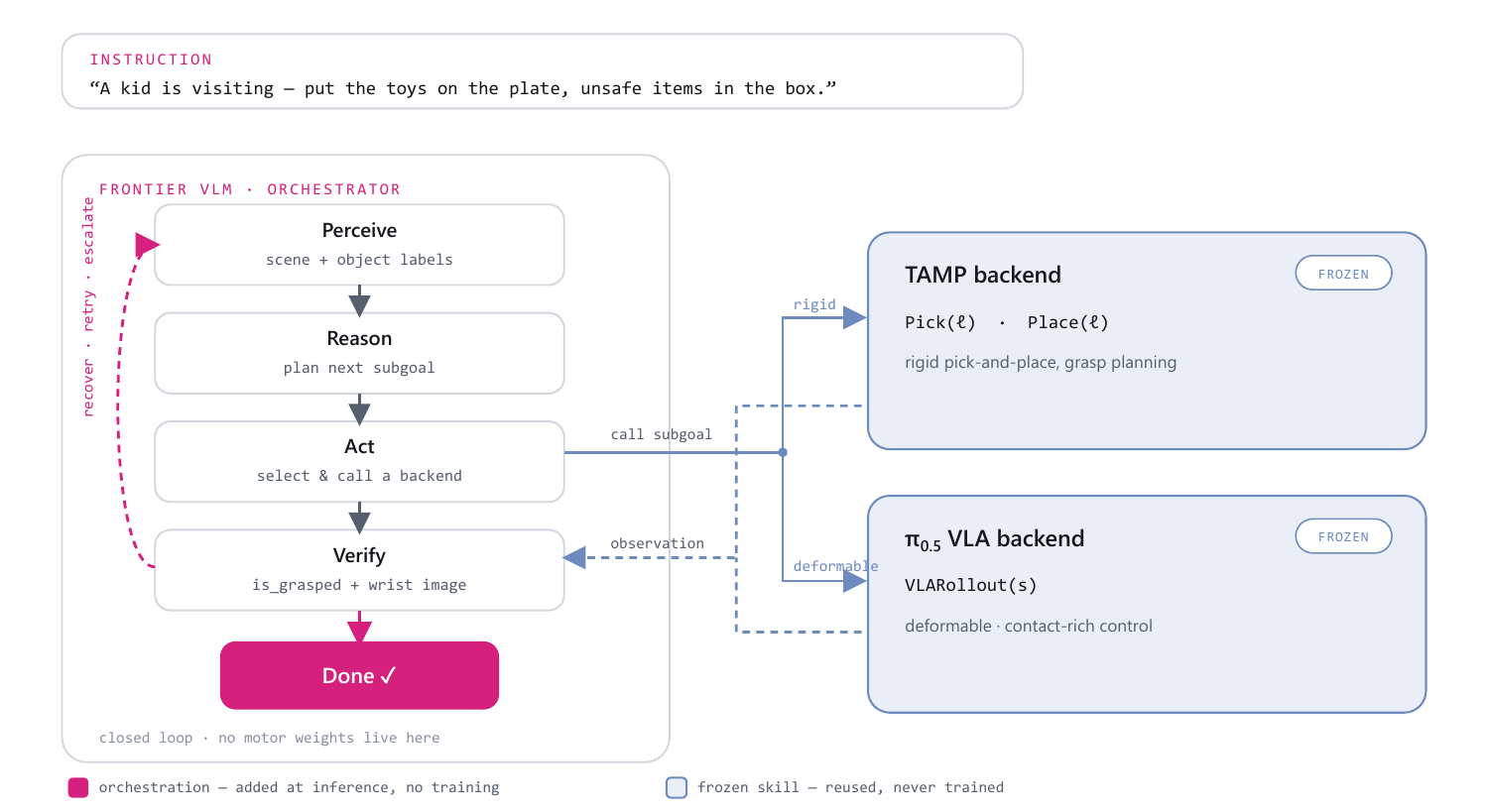}
  \captionof{figure}{A frontier VLM runs a closed loop---observe, reason, act,
  verify---over two \emph{frozen} motor backends, routing each subgoal to TAMP
  (rigid pick-and-place) or the $\pi_{0.5}$ VLA (deformable / recovery), verifying
  the outcome, and recovering on failure.}
  \label{fig:architecture}
\end{minipage}\hfill
\begin{minipage}[t]{0.48\textwidth}
  \centering
  \vspace{0pt}
  \footnotesize
  \begin{algorithm}[H]
  \caption{Closed-loop orchestration}\label{alg:loop}
  $h \leftarrow$ instruction $+$ initial observation\;
  \For{$t = 1,\ldots,T_{\max}$}{
      $a \leftarrow \phi(h,\mathcal{T})$ \tcp*{select next tool call}
      \uIf{$a = \textsc{Done}$}{
          \Return success iff task predicate holds\;
      }
      \uElseIf{$a = \textsc{Perceive}$}{
          append observation $+$ labels to $h$\;
      }
      \uElseIf{$a = \textsc{Pick}/\textsc{DropAbove}(\ell)$}{
          run TAMP; append views $+$ grasp/plan flags to $h$\;
      }
      \ElseIf{$a = \textsc{VLARollout}(s)$}{
          run VLA on $s$; append views $+$ flags to $h$\;
      }
  }
  \Return failure\;
  \end{algorithm}
\end{minipage}
\end{figure}

Figure~\ref{fig:architecture} summarizes the system. A frontier VLM $\phi$ runs as a closed-loop agent on a fixed instruction $I$: at each step it reads the current observation and interaction history, emits one tool call, incorporates the result, and decides again, until it declares the task complete or a fixed budget of $T_{\max}$ tool calls is reached. The agent supplies the task-level loop: it decomposes $I$ into short subgoals, maintains memory of what it has done, selects which frozen motor backend executes each subgoal, verifies the outcome from sensor and visual feedback, and recovers when verification fails. It never emits motor commands itself; all motion is delegated. We contrast this \emph{agentic} condition with \emph{direct} prompting, where $I$ is handed to a single motor policy as one command. (We use agent and orchestrator interchangeably.) We measure the resulting orchestration gap: the increase in success rate when the same frozen motor policy is invoked through the agent rather than prompted directly.

Formally, the agent maps the observation $o\in\mathcal{O}$, history $h\in\mathcal{H}$, and instruction $I\in\mathcal{I}$ to a tool call, $\phi:\mathcal{O}\times\mathcal{H}\times\mathcal{I}\rightarrow\mathcal{T}$; each frozen backend $\pi_b:\mathcal{O}\times\mathcal{S}\rightarrow\mathcal{A}_{\mathrm{motor}}$, $b\in\{\textsc{tamp},\textsc{vla}\}$, executes a short subgoal $s\in\mathcal{S}$ (e.g.\ ``put the red cup on the plate'') and returns control. The backends receive only $s$, never $I$.




\subsection{Components of \methodname{}}
\textbf{Tools.} The agent acts through five tools: \textsc{Perceive} (return camera views, robot state, and the set of detected object labels); \textsc{Pick}$(\ell)$ and \textsc{DropAbove}$(\ell)$ (grasp / place a labeled object via the TAMP backend); \textsc{VLARollout}$(s)$ (execute subgoal $s$ via the VLA backend); and \textsc{Done} (terminate). The full tool JSON schemas and the agent prompt template are given in Appendices~\ref{app:schemas} and~\ref{app:prompt}.

\textbf{Observation and grounding.} After every call the agent receives a wrist image, end-effector pose, gripper aperture, a binary $\textsc{is\_grasped}$ flag read from the gripper-width sensor, and the set of object labels returned by the open-vocabulary detector. These labels are the agent's \emph{vocabulary}: every \textsc{Pick}/\textsc{DropAbove} argument must be one of them, which forces the agent to ground semantic categories from the instruction (``unsafe,'' ``vegetarian,'' ``the smallest'') onto concrete detected objects rather than inventing names. Backends also return \emph{typed} failures---no grasp found, motion-planning failure, unreachable, step-budget exhausted---so the agent learns \emph{why} a step failed, not merely that it did. The full observation channel and the typed failure annotations are detailed in Appendix~\ref{app:observations}.

\textbf{Memory.} The agent carries a running record of the episode---subgoals attempted and their outcomes, what is currently held, and which objects have already been placed. This is what makes long-horizon tasks tractable: a ten- to twenty-step sort or childproofing requires tracking which items remain, not re-moving completed ones, and recognizing when the goal state is reached.

\subsection{Grounding \methodname{} in Actions}
Both backends are pre-existing and frozen---neither is trained or fine-tuned for this work---and each is reached only through a short subgoal.
\begin{itemize}[leftmargin=1.2em]
    \item \textbf{TAMP backend} (\textsc{Pick}/\textsc{DropAbove}): we build on TiPToP~\cite{tiptop} for open-vocabulary grounding, grasp prediction, and collision-free motion planning. TiPToP executes a full pick-and-place as a \emph{single open-loop plan}: once planned it is blind to execution, so a slipped or mislocalized grasp still proceeds to placement and the task fails with no recourse. We instead expose grasping and placement as \emph{two separate tools} the agent calls independently, inserting verification between them---the agent commits to a place only after the grasp is sensor-verified. This converts TiPToP's open-loop pick-and-place into a closed loop and is a direct source of our gains over it.
    \item \textbf{VLA backend} (\textsc{VLARollout}): a frozen $\pi_{0.5}$~\cite{pi05} policy runs closed-loop visuomotor control from a short subgoal. It handles deformable, contact-rich, and cluttered cases, and serves as the recovery path when TAMP cannot plan (inference-loop details in Appendix~\ref{app:vla}).
\end{itemize}
The two are complementary by construction: TAMP gives geometric precision and a verifiable grasp signal; the VLA gives closed-loop robustness where geometric planning breaks.

\subsection{Verification}
Verification is what separates the agent from open-loop prompting, and it draws on two complementary signals. The first is \emph{deterministic}: a \textsc{Pick} counts as successful only if $\textsc{is\_grasped}$ is true at the gripper-width sensor, and reach/plan failures are surfaced as explicit flags. The second is \emph{visual}: the post-action wrist image is returned to $\phi$, which confirms the intended object is in the jaws and gone from the table. The two are combined conservatively---if a backend reports success but the sensor reads an empty gripper, the step is overridden to a failure, so an optimistic backend cannot mislead the agent. Only a verified outcome advances the plan; an unverified one triggers recovery. In particular, a place (\textsc{DropAbove}) is issued only after the preceding grasp is verified, so the agent never transports and releases an object it failed to grasp.

\subsection{Planning and Recovery}

\methodname{} picks a backend per subgoal, verifies, and escalates on failure (Appendix~\ref{app:routing}); key cases:
\begin{itemize}[leftmargin=1.2em]
    \item \textbf{Path planning by object type.} Rigid, table-resting targets go to TAMP; deformable or cable-like objects, and objects inside a container or stacked on another object, go straight to the VLA---geometric grasp planning is unreliable for these.
    \item \textbf{Verify, retry, escalate.} An unverified \textsc{Pick} is retried once (re-perceiving and re-planning at the object's pose); a second failure escalates to the VLA. Escalation is \emph{bidirectional}: if a VLA rollout makes no progress, the agent falls back to a TAMP \textsc{Pick} on a freshly perceived scene.
    
    \item \textbf{Recover.} On a wrong-object grasp the agent returns the object to the table (never the destination) and retries the intended one; when a target is hidden it treats visible objects as occluders and uncovers it; when the destination holds items that do not belong in the goal state, it clears them first.
\end{itemize}

\begin{figure*}[t]
    \centering
    \setlength{\tabcolsep}{1.5pt}
    \renewcommand{\arraystretch}{0.9}
 
    \newcommand{\framew}{0.235\textwidth}
    \newcommand{\capbox}[1]{\parbox[t]{\framew}{\centering\scriptsize\textit{#1}}}
 

    \begin{subfigure}{\textwidth}
        \centering
        \caption*{\small \textit{Task:``Pick up the doll and put it in the basket''}}
            \vspace{1pt}
        \begin{tabular}{@{}cccc@{}}
            \includegraphics[width=\framew]{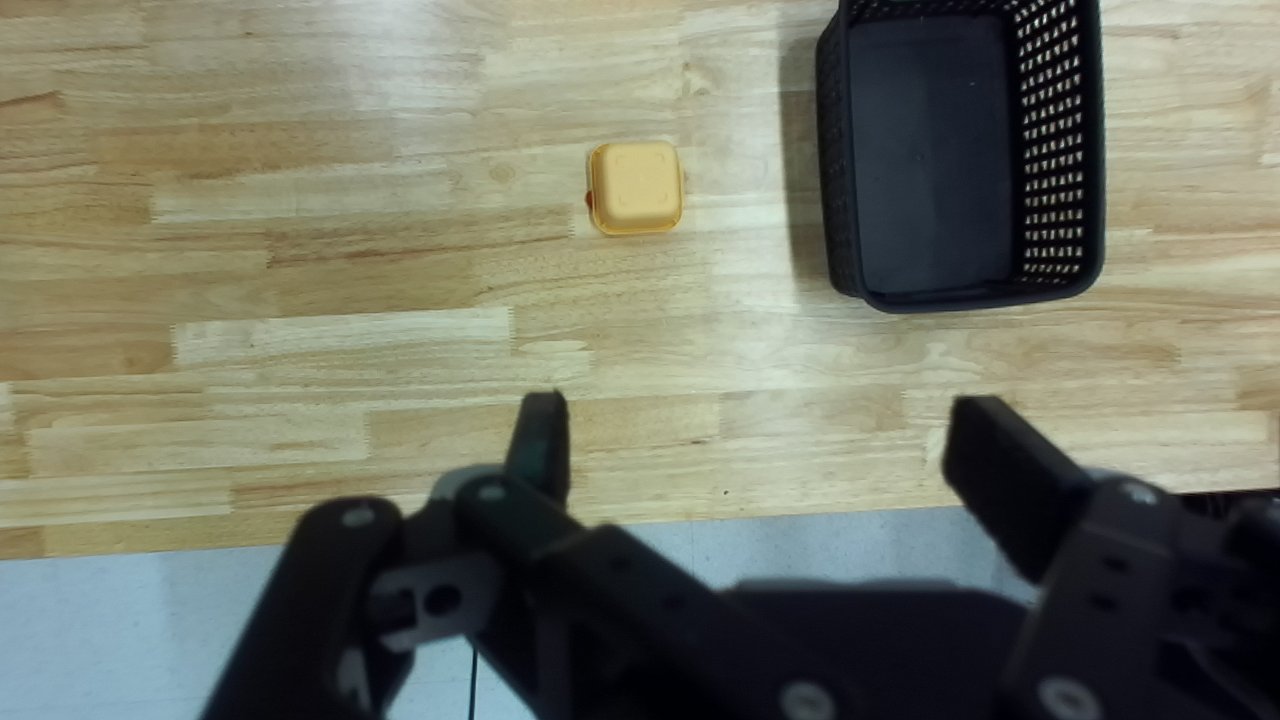} &
            \includegraphics[width=\framew]{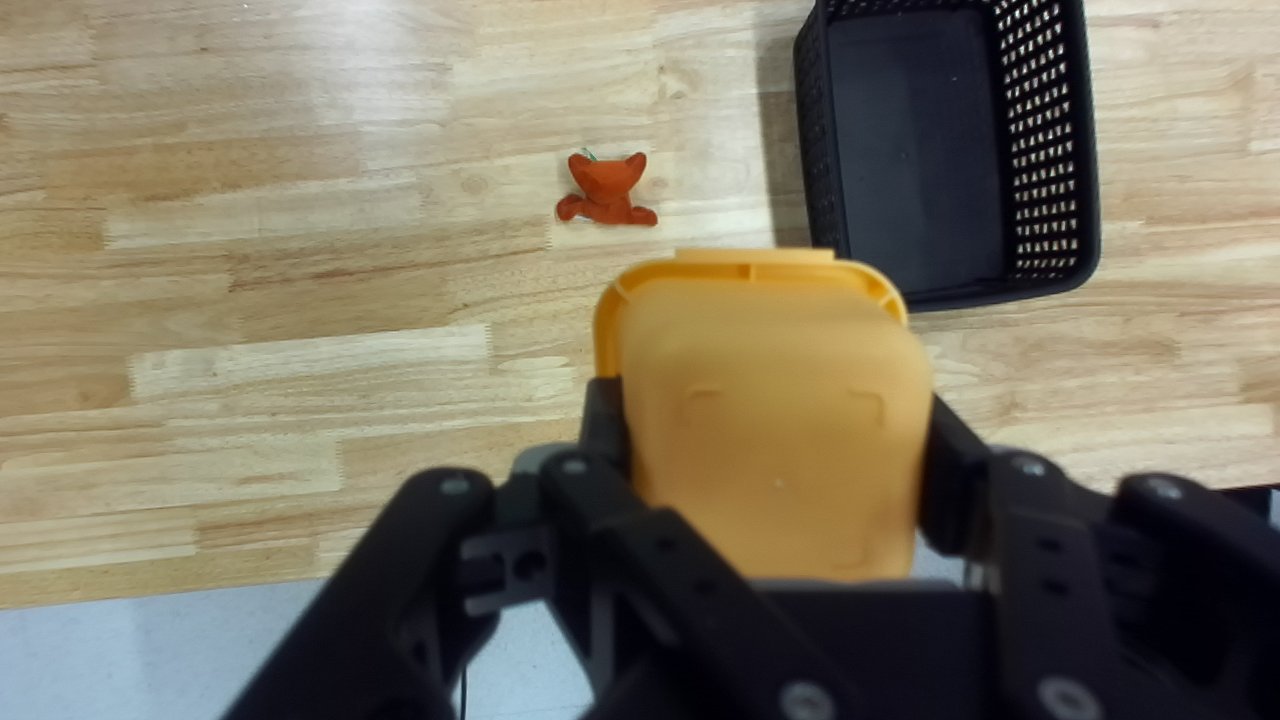} &
            \includegraphics[width=\framew]{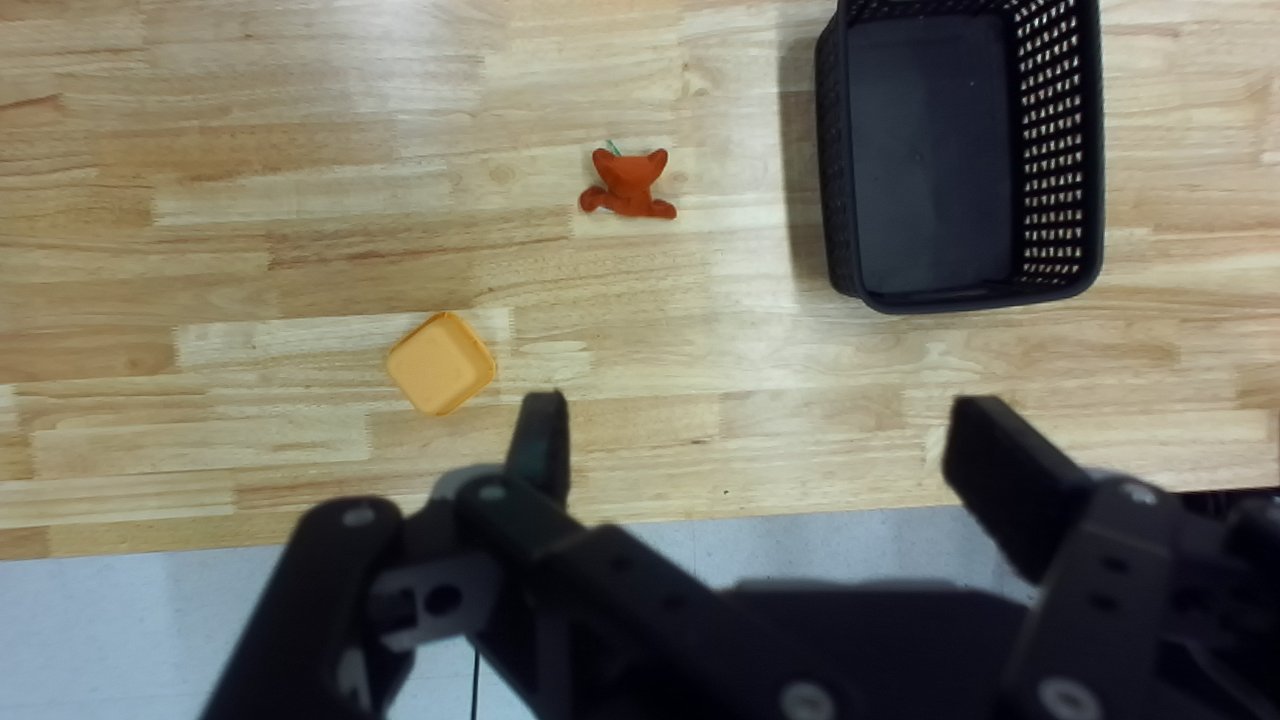} &
            \includegraphics[width=\framew]{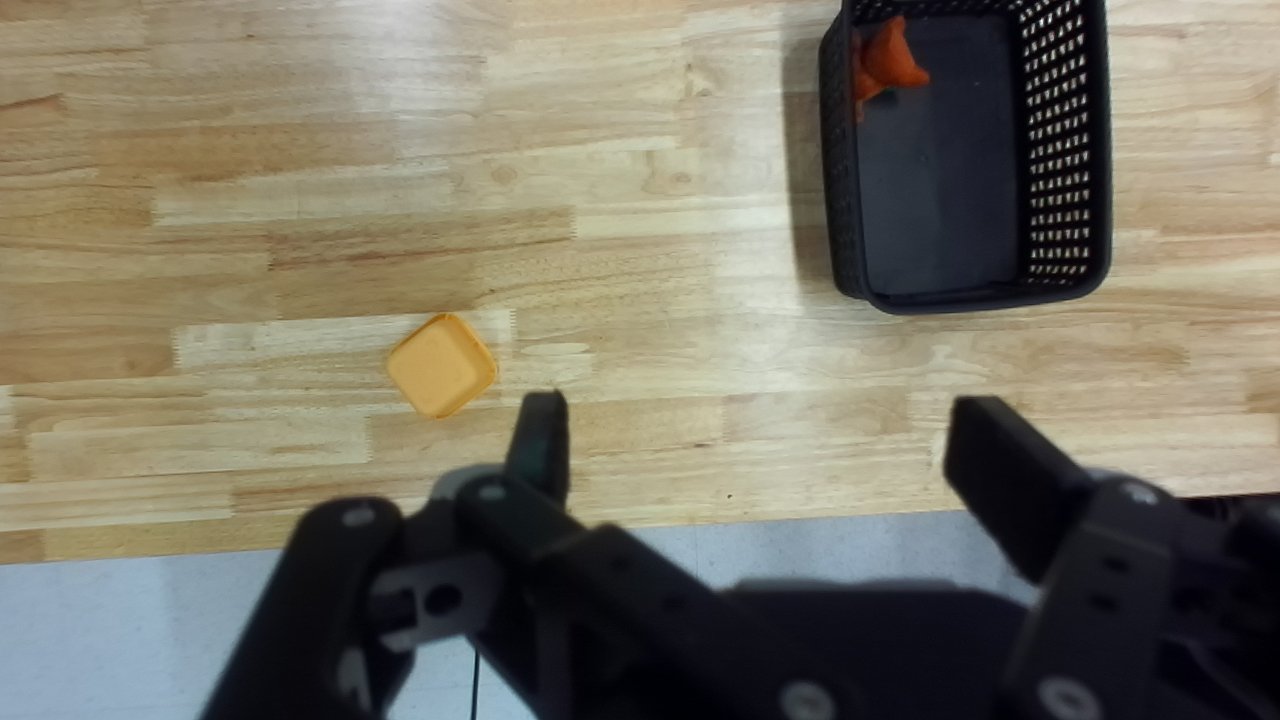} \\[2pt]
            \capbox{``Where is the doll? Might be hidden''} &
            \capbox{``Pick the container. Here's the doll!''} &
            \capbox{``Pick up the doll''} &
            \capbox{``Place doll in the basket''} \\
        \end{tabular}
        
    \end{subfigure}
        \vspace{2pt}

 \hrule

    \vspace{4pt}
 
    \begin{subfigure}{\textwidth}
        \centering
         \caption*{\small \textit{Task:``A child is coming over - put the items they would want to play with on the plate''}}
                     \vspace{1pt}

        \begin{tabular}{@{}cccc@{}}
            \includegraphics[width=\framew]{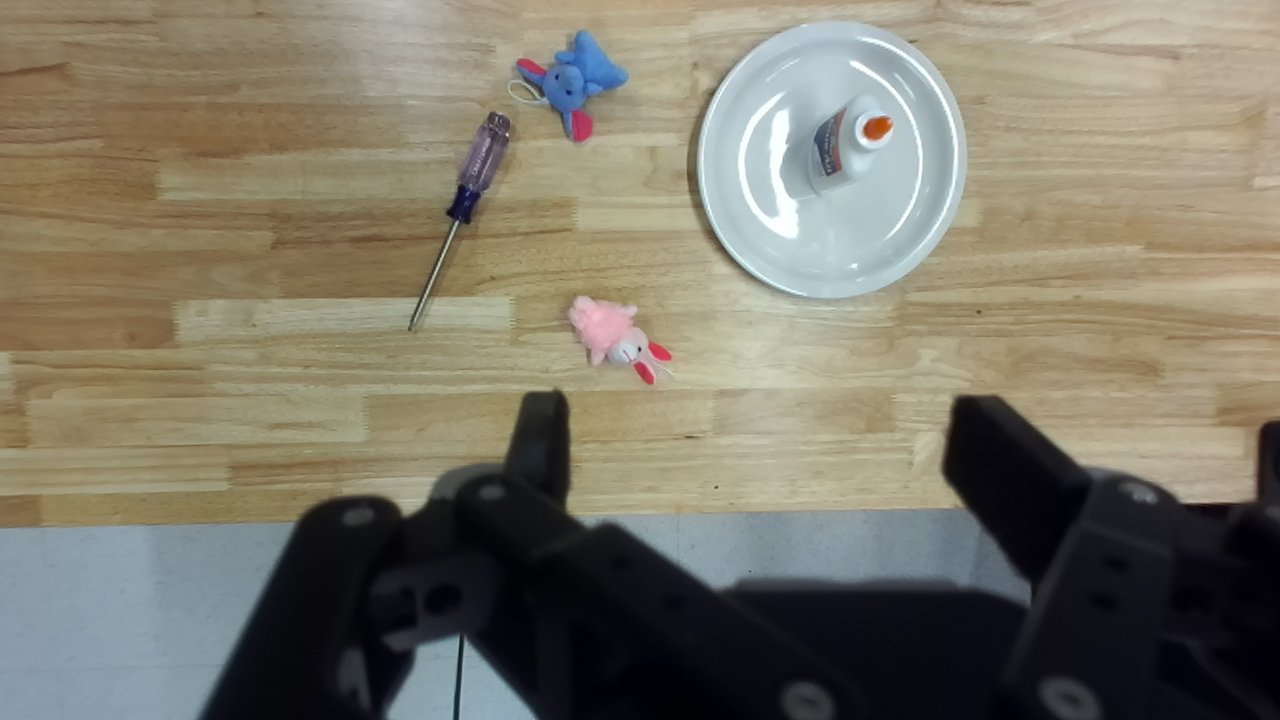} &
            \includegraphics[width=\framew]{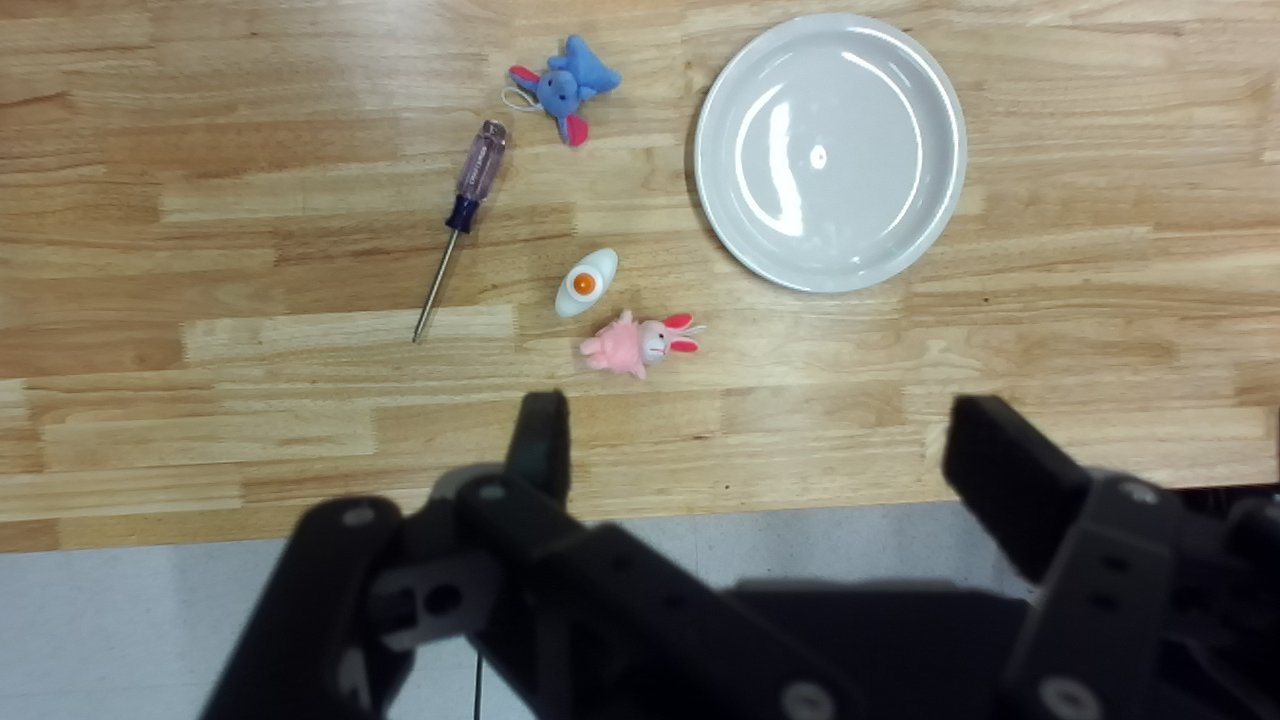} &
            \includegraphics[width=\framew]{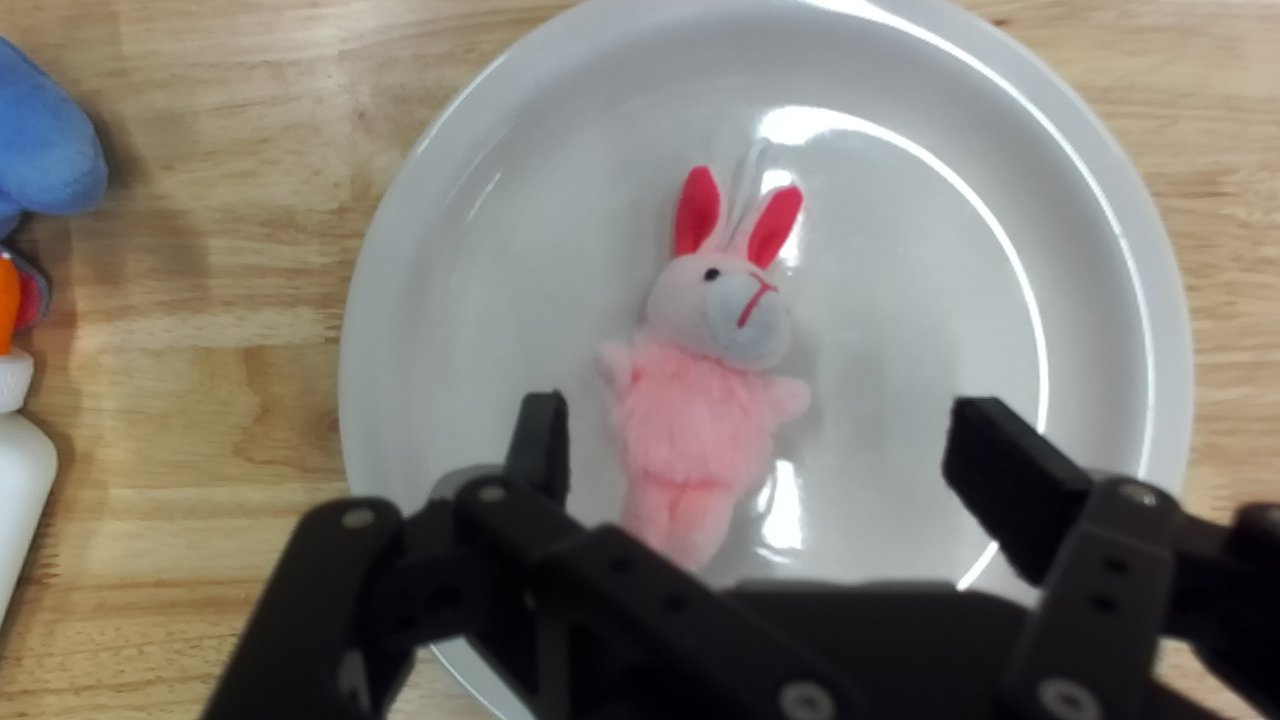} &
            \includegraphics[width=\framew]{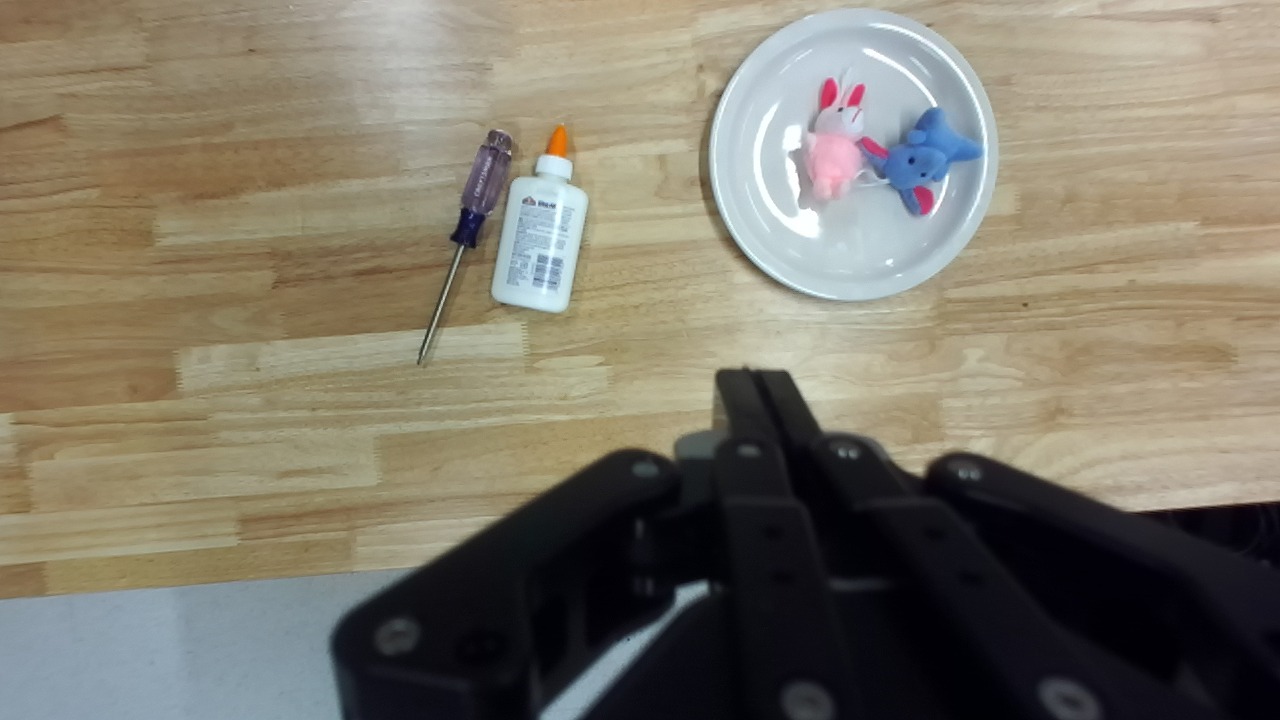} \\[2pt]
            \capbox{``Only the toys are kid-safe''} &
            \capbox{``Glue's on the plate --- unsafe; clear it''} &
            \capbox{``Place the bunny on the plate''} &
            \capbox{``Place the elephant on the plate''} \\
        \end{tabular}
       
    \end{subfigure}
        \vspace{2pt}

 \hrule

    \vspace{4pt}
    \begin{subfigure}{\textwidth}
        \centering
         \caption*{\small \textit{Task:``A vegetarian guest is coming for dinner. Set the bowl for them.''}}
         \vspace{1pt}
        \begin{tabular}{@{}cccc@{}}
            \includegraphics[width=\framew]{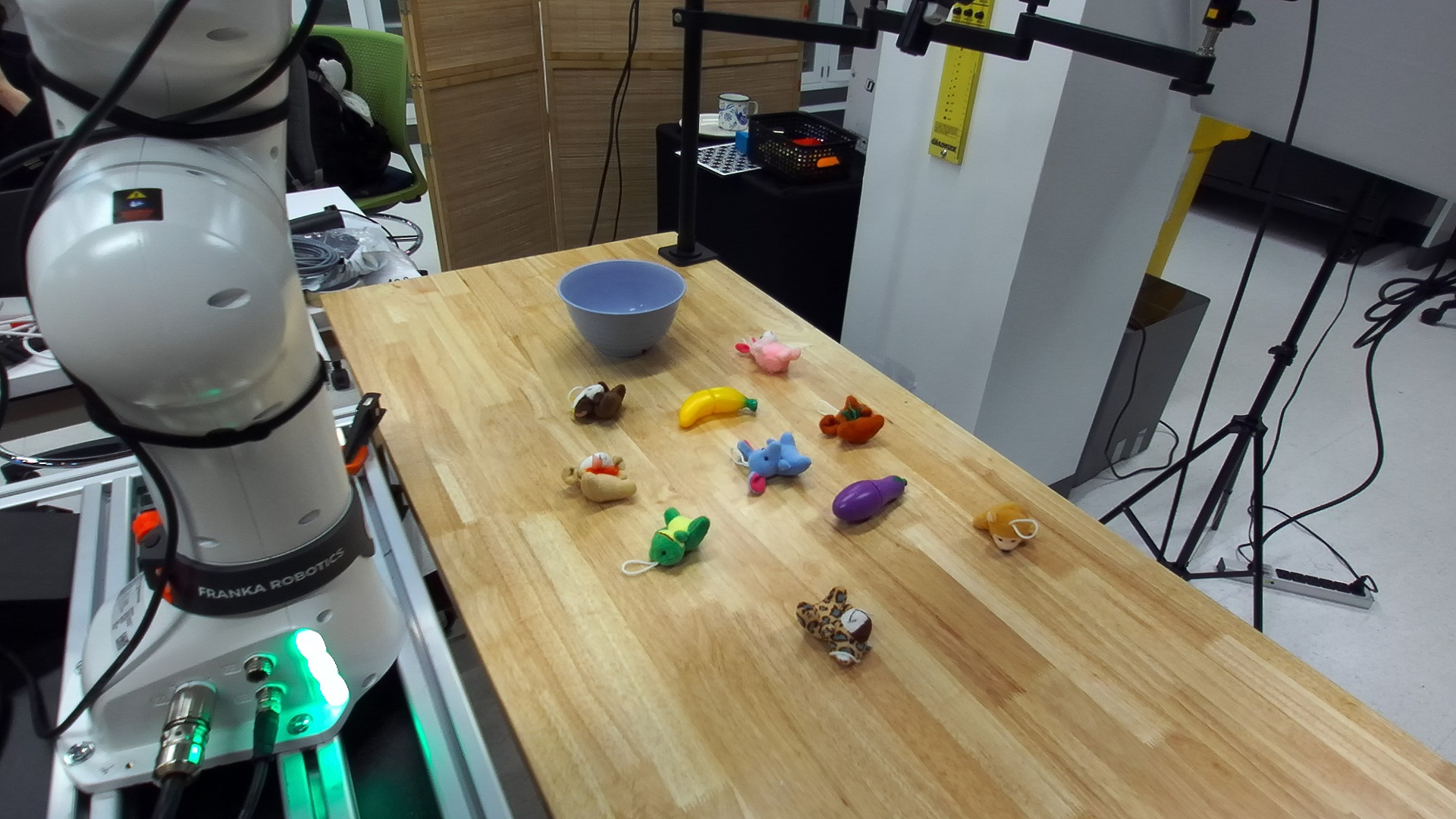} &
            \includegraphics[width=\framew]{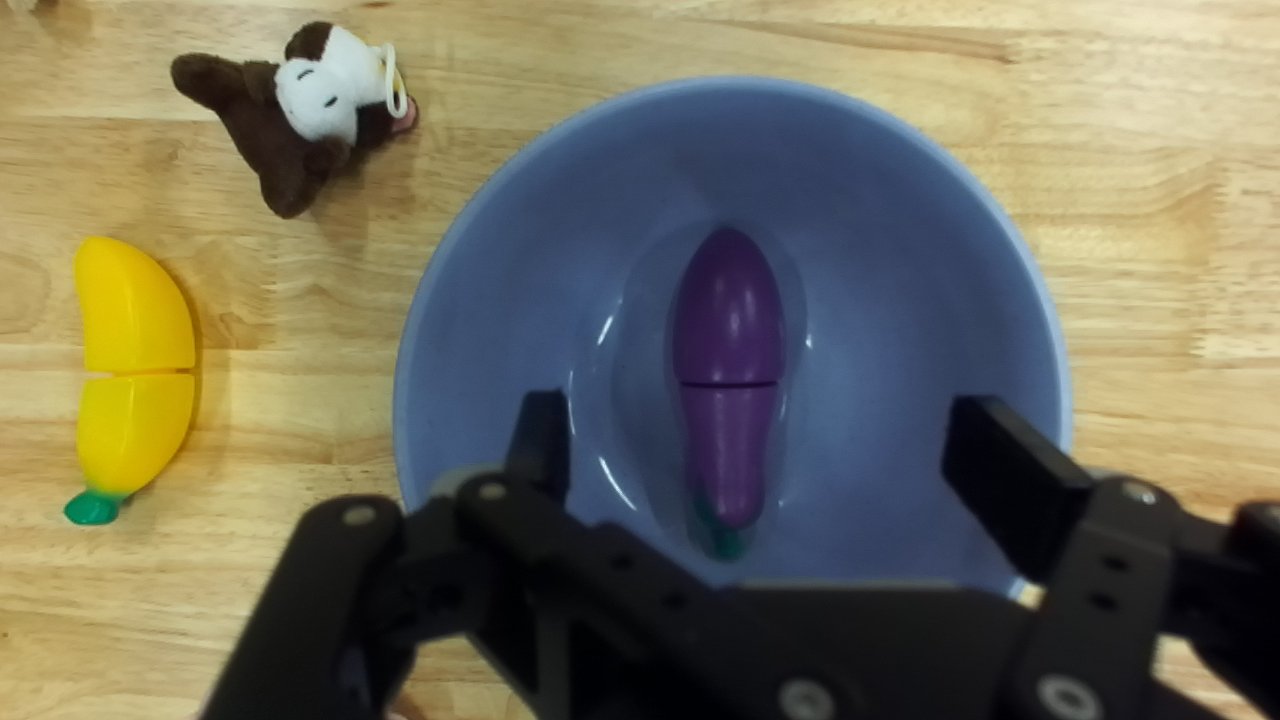} &
            \includegraphics[width=\framew]{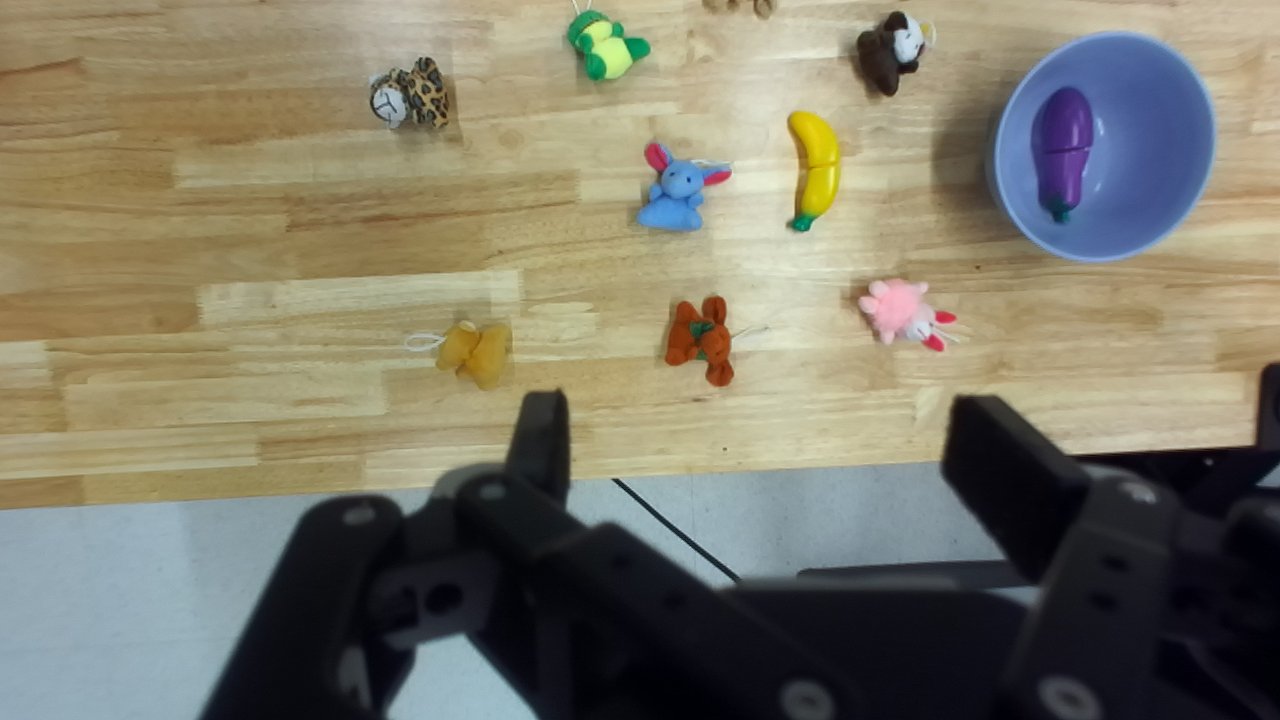} &
            \includegraphics[width=\framew]{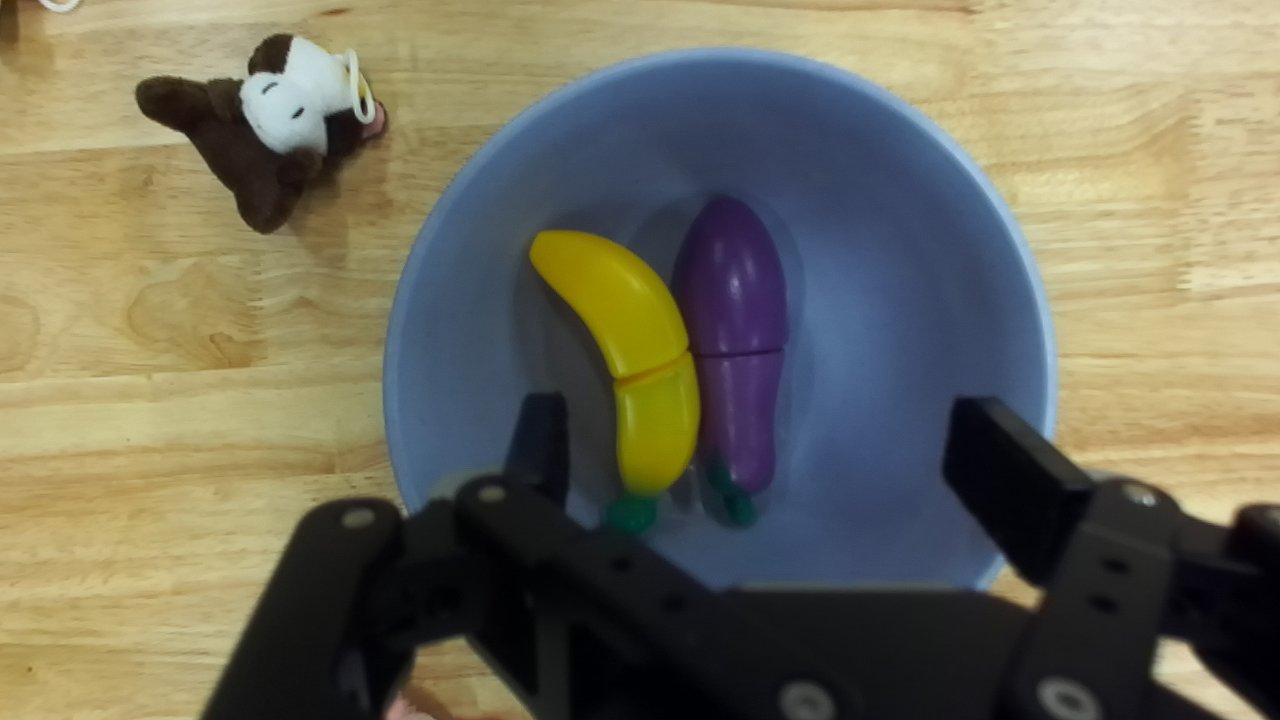} \\[2pt]
            \capbox{``Only banana \& eggplant are vegetarian''} &
            \capbox{``Put the eggplant in the bowl''} &
            \capbox{``I need to pick up the banana''} &
            \capbox{``Place banana in the bowl''} \\
        \end{tabular}
       
    \end{subfigure}

    \caption{One task per row, with the instruction above and \methodname{}'s per-frame reasoning below each frame.}

    \label{fig:qualitative_rollouts}
\end{figure*}

Re-perceiving before each grasp---rather than committing to one upfront plan---is what makes this robust to a changing world. If the target has \emph{moved} since it was last seen (nudged by a previous action or by the approach itself), the agent re-plans against its current pose instead of grasping where it used to be (Figure~\ref{fig:error_correction}); and if the target is \emph{not yet visible}, the agent removes occluders and re-perceives until it appears before grasping (Figure~\ref{fig:qualitative_rollouts}). Both are out of reach for an open-loop plan, which is blind once computed---concrete cases where closing the loop turns a guaranteed failure into a success.


\section{Experimental Setup}
\label{sec:experiments}

We evaluate the \methodname{} agent and baselines on the LIBERO-PRO simulation benchmark and the DROID real-world platform.

\paragraph{Tasks and Benchmarks.} We test \methodname{} on the DROID tabletop manipulation setup~\cite{droid}.
The agent routes each subgoal to one of two frozen backends---closed-loop TAMP (\textsc{Pick}/\textsc{DropAbove}) or the $\pi_{0.5}$-DROID VLA (\textsc{VLARollout})---across \num{30} tasks. The tasks are \emph{capability probes}: each isolates a part of the task-level stack (world knowledge, conditional logic, same-scene prompt variation, multi-step reasoning, long-horizon memory, spatial reasoning, distractor rejection, obstacle handling, error recovery), so a failure can be attributed to missing reasoning or a missing closed loop rather than to motor incompetence (the full per-subset task list in~Appendix~\ref{app:tasks}). In simulation, we report performance on the LIBERO-PRO benchmark~\cite{liberopro} with $\pi_{0.5}$-LIBERO as the base policy, perturbing objects, spatial relations, and goals. The agent can call seven tools:
\textsc{Perceive}, \textsc{Grasp}, \textsc{Place},
\textsc{VLARollout}, \textsc{VerifyCandidate},
\textsc{GoHome}, and \textsc{Release}.
Only \textsc{VLARollout} invokes the frozen learned policy; the remaining
manipulation tools use analytic controllers. Appendix~\ref{app:sim-tools}
summarizes the complete interface. Hardware, software, and the scoring protocol are specified in Appendices~\ref{app:hardware} and~\ref{app:scoring}.

\paragraph{Baselines and Metrics.} Every comparison holds the learned weights fixed and changes only the inference-time process. On hardware we compare against the two motor backends used \emph{directly}: direct $\pi_{0.5}$ prompting (the VLA with no agent) and TiPToP (the same grounding and motion planning we use, but executed as an open-loop pick-and-place). Comparing our agent to each isolates one contribution---against direct $\pi_{0.5}$, the reasoning the agent supplies; against TiPToP, the value of closing the loop around TAMP. In simulation we compare raw $\pi$0, raw $\pi_{0.5}$, and CaP-Agent0, and sweep the reasoner across nine frontier VLMs. Tasks are scored as binary success: 5 trials each on the real robot with Claude Opus 4.7 as the reasoner, 10 per sim task per reasoner (Appendix~\ref{app:ablations}).
\section{Results}

\label{sec:results}

Our agent turns two frozen, individually imperfect motor backends into a more capable system without additional robot data or motor-policy training. The gains come from two sources that we isolate in turn: task-level reasoning over the VLA, and closed-loop verification around TAMP. Both are supplied at inference time: the policy weights are fixed, and no new demonstrations are collected. Prior work often obtains these capabilities by training additional modules or scaling robot data---memory through policy fine-tuning~\cite{memer}, reasoning through embodied chain-of-thought training~\cite{ecot}, verification through learned reward or verifier models~\cite{robomonkey, liang2026robometer}, decomposition through trained high-level planners~\cite{hirobot}, and broad competence through larger robot datasets or dual-system architectures~\cite{groot,geminirobotics} (Table~\ref{tab:comparison}). Our experiment asks how much of this capability can instead be recovered by changing only the inference-time process around frozen motor skills. Orchestration is not cost-free: it spends inference compute and adds latency (Section~\ref{sec:discussion}). But it avoids the cost that dominates robot learning: collecting new robot data and retraining policies. 

\begin{table}[t]
\centering
\begin{minipage}[t]{0.50\textwidth}
\centering
\captionof{table}{LIBERO-PRO success rate (\%) across six perturbation suites. The same frozen $\pi_{0.5}$-LIBERO weights are used throughout; only the inference-time process changes.}
\label{tab:liberopro_models}
\resizebox{\textwidth}{!}{%
\begin{tabular}{@{}lccccccc@{}}
\toprule
\textbf{Method} & \textbf{Obj.} & \textbf{Obj.} & \textbf{Sp.} & \textbf{Sp.} & \textbf{Goal} & \textbf{Goal} & \textbf{Mean} \\
 & \textbf{swap} & \textbf{task} & \textbf{swap} & \textbf{task} & \textbf{swap} & \textbf{task} & \\
\midrule
$\pi_0$-LIBERO ~\cite{pi0}           & \num{0}  & \num{0}  & \num{0}  & \num{0}  & \num{0}  & \num{0}  & \num{0}    \\
$\pi_{0.5}$-LIBERO~\cite{pi05}      & \num{17} & \num{1}  & \num{20} & \num{1}  & \num{38} & \num{0}  & \num{12.8} \\
CaP-Agent0~\cite{capx}       & \num{22} & \num{18} & \num{12} & \num{14} & \num{26} & \num{17} & \num{18.2}   \\
\rowcolor{magenta!25}\textbf{\methodname{} (ours)} & \num{54} & \num{54} & \num{66} & \num{80} & \num{44} & \num{22} & \textbf{\num{53.3}} \\
\bottomrule
\end{tabular}%
}
\end{minipage}
\hfill
\begin{minipage}[t]{0.48\textwidth}
\centering
\captionof{table}{Real-robot success rate (\%) by capability probe. All learned motor behavior uses the same $\pi_{0.5}$-DROID weights; only the inference-time process changes.}
\label{tab:real-robot}
\resizebox{\textwidth}{!}{%
\begin{tabular}{@{}lccc>{\columncolor{magenta!25}}c@{}}
\toprule
\textbf{Category} & \textbf{Tasks} & \textbf{$\pi_{0.5}$-DROID}~\cite{pi05} & \textbf{TiPToP}~\cite{tiptop} & \textbf{\methodname{}} (\textbf{ours}) \\
\midrule
Simple pick-place          & 4  & 95 & 80 & 100 \\
World knowledge            & 4  & 0  & 90 & 100 \\
Conditional logic          & 4  & 0  & 95 & 100 \\
Multi-step reasoning       & 4  & 0  & 25 & 100 \\
Spatial reasoning          & 4  & 20 & 75 & 100 \\
Obstacle/Safety reasoning  & 4  & 0  & 0  & 90  \\
Error recovery             & 4  & 10 & 0  & 90  \\
Long-horizon memory        & 2  & 0  & 0  & 100 \\
\textbf{Overall}           & 30 & 16.7 & 48.7 & \textbf{97.3} \\
\bottomrule
\end{tabular}%
}
\end{minipage}

\end{table}

\subsection{LIBERO-PRO Simulation Benchmark}

We evaluate \methodname{} on LIBERO-PRO~\cite{liberopro}, a larger-scale benchmark of complex tabletop manipulation tasks. Unlike the original LIBERO benchmark~\cite{libero}, which evaluates a relatively narrow set of configurations close to training distribution, LIBERO-PRO tests a wider range of out-of-distribution scenarios and robustness to perturbations in object states, instructions, and environments.
With even minor perturbations, the state-of-the-art $\pi_{0.5}$-LIBERO policy degrades significantly, suggesting that the policy might be overfitting to the LIBERO training set.
CaP-Agent0 also struggles with \num{18}\% success (Table~\ref{tab:liberopro_models}). Under a zero-shot protocol with no task-specific
memory, reference-seed exploration, or fine-tuning, \methodname{}
achieves the highest reported mean across the six evaluated
LIBERO-PRO perturbation suites, improving the frozen
$\pi_{0.5}$-LIBERO baseline from 12.8\% to 53.3\%. While LIBERO-PRO requires minimal reasoning, we find that a combination of \methodname{}'s error recovery and the tools we provide it (Appendix~\ref{app:schemas}) can orchestrate the weak base policy to success.

\subsection{\methodname{} v/s Vanilla VLA on DROID: The Orchestration Gap}

In the real-world, we evaluate \methodname{} against the state-of-the-art $\pi_{0.5}$ VLA. As illustrated in Table~\ref{tab:real-robot}, \methodname{} significantly improves over the vanilla VLA, combining existing action capabilities with high-level reasoning. On the reasoning-limited probes, the raw VLA averages \num{4.6}\% while \methodname{} reaches \num{96.9}\% with the same VLA, suggesting that the primary bottleneck lies in instruction interpretation, decomposition, memory, and recovery, rather than low-level control. On simple pick-and-place tasks, the absolute increase on the control tasks is 5 percentage points (\num{95}$\rightarrow$\num{100}\%). Orchestration supplies the missing reasoning; it did not reduce success on the control tasks evaluated here.



\subsection{\methodname{} v/s TAMP on DROID: Orchestrating a Visuomotor Policy}

Another axis independent of high-level reasoning comes from \emph{how} the agent uses low-level primitives and policies. TiPToP~\cite{tiptop} uses the same open-vocabulary grounding and motion planning we do, but runs a pick-and-place as a single open-loop primitive: it is blind to execution, and a slip or mislocalized grasp can lead to failure. We instead expose grasping and placement as two tools and verify the grasp between them, committing to a place only once the grasp is sensor-confirmed.

The cost of staying open-loop is visible even where grounding is trivial. On the simple pick-and-place control, TiPToP reaches only \num{80}\%---\emph{below} the raw VLA---because subtle control mistakes can lead to failure, while our closed loop detects the empty gripper and retries, achieving \num{100}\% (Table~\ref{tab:real-robot}). This gap widens when we introduce adversarial perturbations: if the target is nudged after planning, an open-loop plan grasps where the object used to be, and if it is initially hidden, the plan never adapts. Our agent observes before each grasp, re-plans against the object's current pose, and uncovers occluded targets before grasping (Figure~\ref{fig:error_correction}). These failures are unavoidable for an open-loop pick-and-place plan, and they account for our margin over TiPToP on the multi-step, obstacle, and recovery probes.

Additional results and ablations are provided in Appendices~\ref{app:results} and~\ref{app:ablations}.

\section{Conclusion}
\label{sec:discussion}
In this work, we presented \methodname{}, a framework that augments the low-level capabilities of generalist policies with a frontier VLM agent to address the \emph{orchestration gap} in robotics. We show that much of the task-level stack---decomposition, memory, routing, verification, and recovery---can be supplied at inference time by a physical agent orchestrator over \emph{frozen} motor backends, with no additional robot-data collection or policy post-training. Framing the problem this way exposes an \emph{orchestration gap}: a large share of what looks like missing motor capability is in fact missing \emph{use} of an already-capable policy. Concretely, a frozen VLA recovers much of its missing capability on reasoning-limited tasks when driven through \methodname{} rather than prompted directly, and splitting an open-loop pick-and-place into separately verified steps recovers tasks that open-loop planning fails by construction. The practical implication is a sequencing rule for robot learning: before spending scarce robot data to teach a policy to reason, measure how much of that capability an inference-time agent already recovers around the frozen policy.

\noindent \textbf{Future work.}
\methodname{}'s structured inference traces also suggest a path from test-time orchestration to policy improvement. Each episode records observations, subgoals, backend choices, verification signals, recovery decisions, and final outcomes. Future work could filter successful, well-verified traces and use them to distill task decomposition, routing, and recovery behavior into smaller orchestrators or task-level policy adapters through behavior cloning, offline reinforcement learning, or targeted fine-tuning. The main challenge is selecting reliable supervision: false successes, verifier errors, or accidental behaviors could otherwise be propagated into the learned policy.

\noindent \textbf{Limitations.}
\methodname{} is limited by the skills it orchestrates: while it can compensate for weak policies, there is a ceiling set by the ``support'' of the low-level tools. Verification is also imperfect---partial observability from occlusions can hide a poor grasp, allowing a false success to propagate downstream. Finally, because \methodname{} relies on API calls to a frontier model for orchestration, it adds per-step latency and cost, making it challenging for low-latency or high-speed applications. 

\section*{Acknowledgements}
{
This research was partially supported by Microsoft Research, the Schmidt Sciences AI2050 fellowship, the Google ML and Systems Junior Faculty Awards, and the Google Research Scholar program, with compute support from the Gemini Academic Program. The authors also thank Elad Hazan, Anirudha Majumdar, Tomer Galanti, Nadav Timor, and Mingtong Zhang for helpful discussions.
}
\clearpage
\bibliographystyle{unsrt}
\bibliography{references}   

\clearpage
\beginappendix{
\section{Full Problem Setup}
\label{app:problem}

We study general-purpose tabletop manipulation from natural-language instructions. The agent observes the scene through cameras, receives a language instruction, and must execute physical actions until the instruction is satisfied. The instructions extend beyond direct pick-and-place: they include category reasoning, conditional logic, negation, comparison, counting, multi-step sequencing, long-horizon progress tracking, and recovery from failed grasps.

Let $\mathcal{I}$ be the space of natural-language instructions and $\mathcal{O}$ the space of observations available to the agent. Let $\mathcal{A}_{\mathrm{motor}}$ denote the continuous motor-action space of the robot. A VLA policy is a function
\[
\pi : \mathcal{O} \times \mathcal{S} \rightarrow \mathcal{A}_{\mathrm{motor}},
\]
where $\mathcal{S} \subset \mathcal{I}$ is the subspace of short, concrete, visually grounded subgoals on which the policy is reliable. A user instruction $I \in \mathcal{I}$ is solved if the final world state satisfies a task-specific success predicate $\mathbb{1}_{I}(\mathrm{scene}) = 1$.

The standard recipe collapses interpretation, planning, and execution into a single VLA call: $\pi(o,I)$. This works when $I$ is already VLA-legible, but degrades when $I$ requires intermediate reasoning. We instead factor interpretation, planning, and verification into an inference-time process operating over the same observation channel.

\section{Task Suite}
\label{app:tasks}

We evaluate across the capability probes in Table~\ref{tab:task-summary};
per-task instructions, scene contents, and success predicates follow,
each row showing a thumbnail of the task's initial state. Predicates use $\rel{on}(x,r)$ (object $x$ rests on receptacle
$r$), $\rel{in}(x,c)$ (object $x$ inside container $c$),
$\rel{grasped}(x)$ (stable grasp achieved), $\rel{color}(\cdot)$ and
$\rel{size}(\cdot)$ for attributes, and $\rel{count}(\cdot)$ for
cardinality. Unless stated otherwise, success additionally requires that
non-target objects are not displaced.

\begin{table}[h]
\centering
\begin{tabular}{lr}
\toprule
Capability probe & \# Tasks \\
\midrule
Pick-and-place (control)    & 4 \\
World knowledge             & 4 \\
Conditional logic           & 4 \\
Multi-step reasoning        & 4 \\
Spatial reasoning           & 4 \\
Obstacle reasoning          & 4 \\
Error recovery              & 4 \\
Long-horizon memory         & 2 \\
\midrule
Total & 30 \\
\bottomrule
\end{tabular}
\caption{Task suite composition.}
\label{tab:task-summary}
\end{table}


\subsection{Pick-and-place (control)}
\label{app:tasks:pnp}
\emph{Simple single-step control condition: a named object to a named
receptacle.}

\begin{longtable}{@{} L{0.04\textwidth} C{0.13\textwidth} L{0.23\textwidth} L{0.22\textwidth} L{0.26\textwidth} @{}}
\caption{Pick-and-place control tasks ($N=4$).}\label{tab:tasks-pnp}\\
\toprule
ID & Init. & Instruction (verbatim) & Scene contents & Success predicate \\
\midrule
\endfirsthead
\caption[]{Pick-and-place control tasks (continued).}\\
\toprule
ID & Init. & Instruction (verbatim) & Scene contents & Success predicate \\
\midrule
\endhead
\midrule \multicolumn{5}{r@{}}{\footnotesize continued on next page}\\
\endfoot
\bottomrule
\endlastfoot
PP1 & \includegraphics[width=0.12\textwidth]{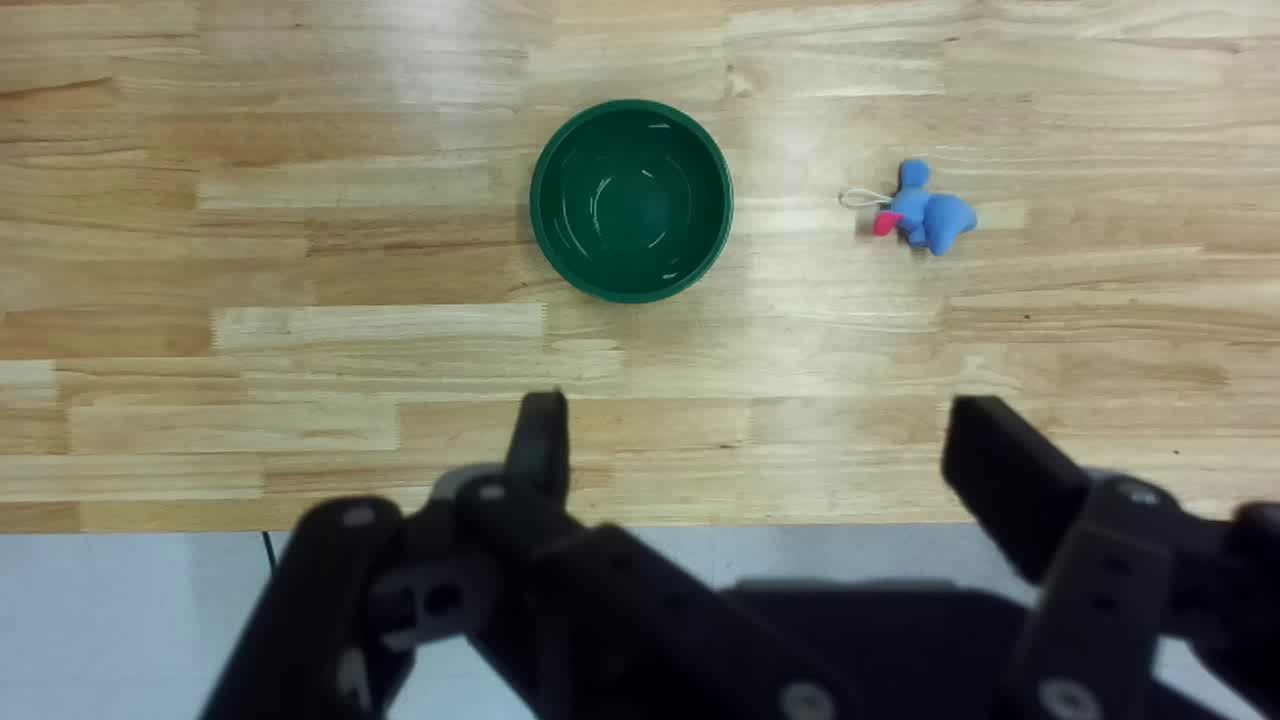}
    & \emph{``Pick up the doll and put it in the bowl.''}
    & doll, bowl
    & $\rel{in}(\text{doll}, \text{bowl})$ \\
\addlinespace
PP2 & \includegraphics[width=0.12\textwidth]{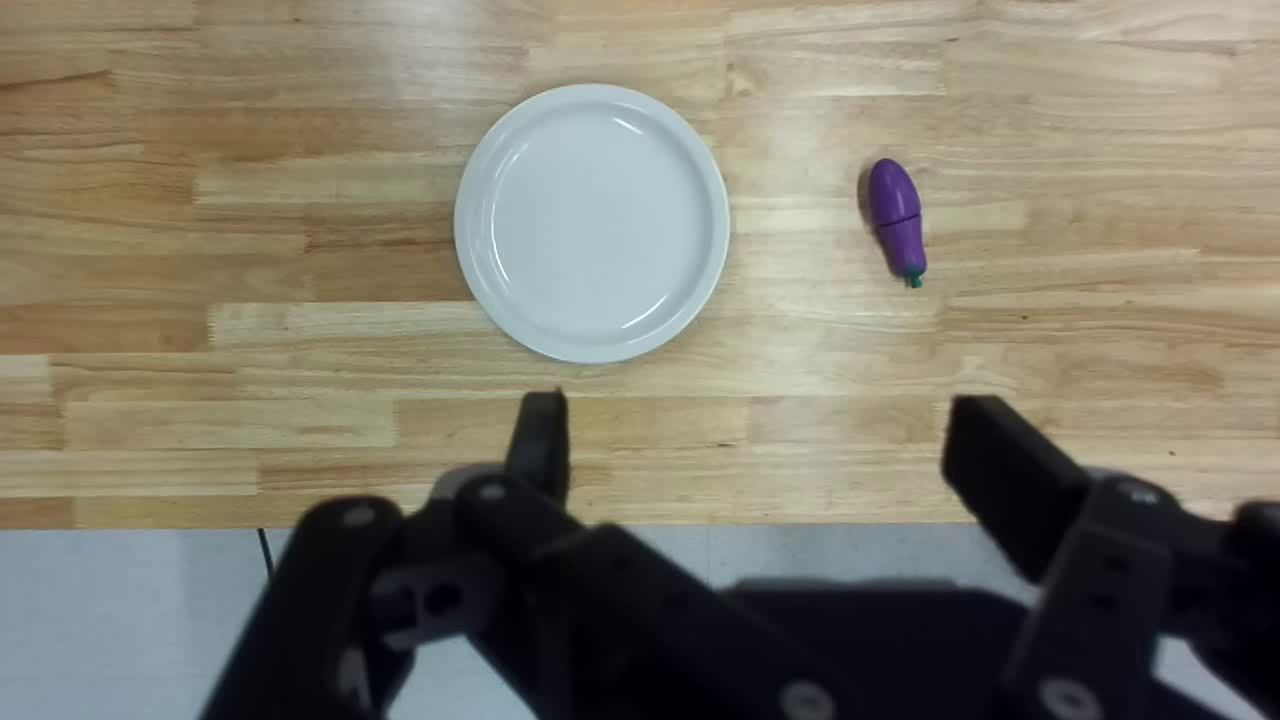}
    & \emph{``Pick up the eggplant and put it on the plate.''}
    & eggplant, plate
    & $\rel{on}(\text{eggplant}, \text{plate})$ \\
\addlinespace
PP3 & \includegraphics[width=0.12\textwidth]{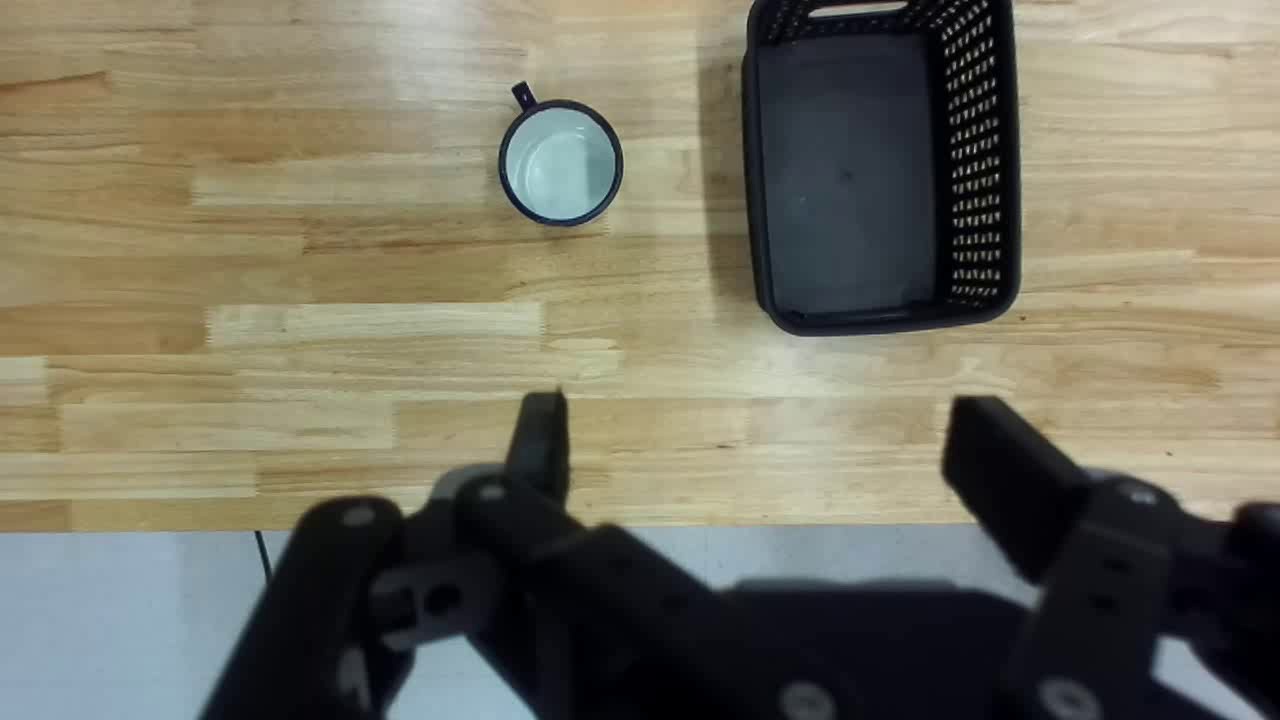}
    & \emph{``Pick up the cup and put it in the basket.''}
    & cup, basket
    & $\rel{in}(\text{cup}, \text{basket})$ \\
\addlinespace
PP4 & \includegraphics[width=0.12\textwidth]{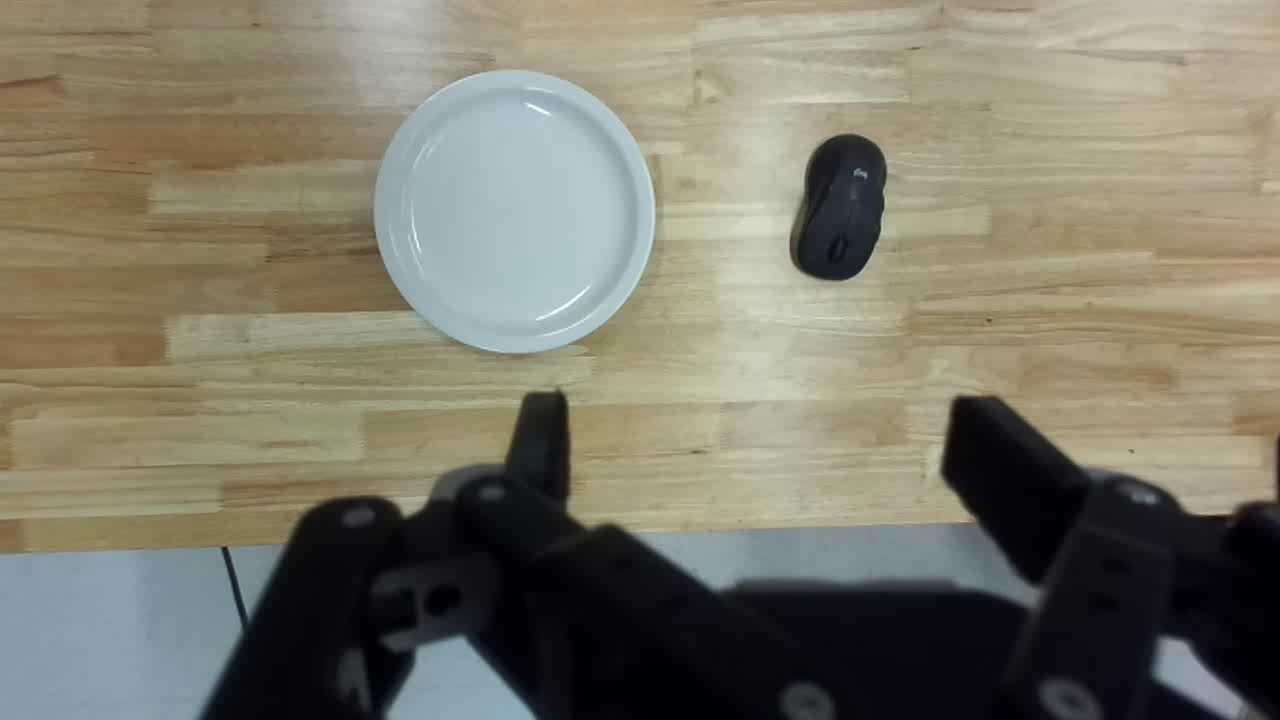}
    & \emph{``Pick up the mouse and put it on the plate.''}
    & mouse, plate
    & $\rel{on}(\text{mouse}, \text{plate})$ \\
\end{longtable}

\subsection{World knowledge}
\label{app:tasks:world}
\emph{Probes whether the policy resolves a referring expression using
facts not stated in the scene. The four prompts share one scene; the
target differs.}

\begin{longtable}{@{} L{0.04\textwidth} C{0.13\textwidth} L{0.23\textwidth} L{0.22\textwidth} L{0.26\textwidth} @{}}
\caption{World-knowledge tasks ($N=4$).}\label{tab:tasks-world}\\
\toprule
ID & Init. & Instruction (verbatim) & Scene contents & Success predicate \\
\midrule
\endfirsthead
\caption[]{World-knowledge tasks (continued).}\\
\toprule
ID & Init. & Instruction (verbatim) & Scene contents & Success predicate \\
\midrule
\endhead
\midrule \multicolumn{5}{r@{}}{\footnotesize continued on next page}\\
\endfoot
\bottomrule
\endlastfoot
WK1 & \multirow{4}{*}{\includegraphics[width=0.12\textwidth]{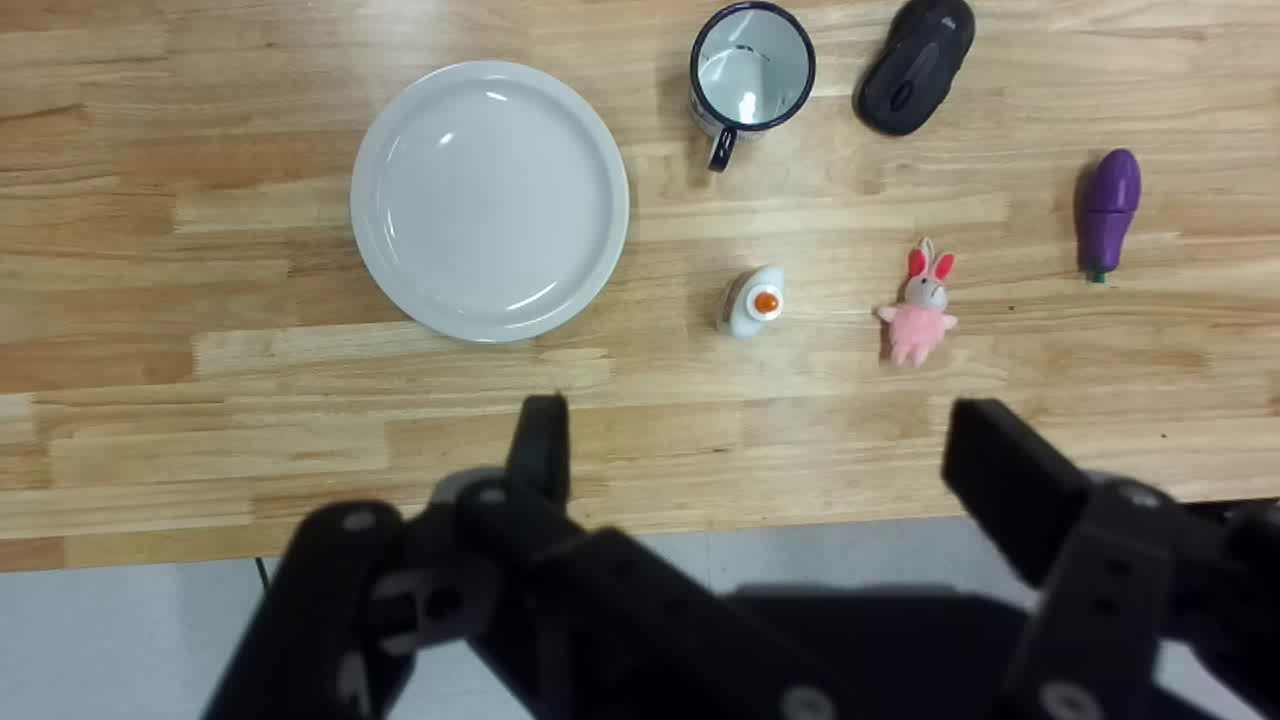}}
    & \emph{``Pick up something you would put in ratatouille and put it on the plate.''}
    & doll, eggplant, glue, cup, mouse, plate
    & $\rel{on}(\text{eggplant}, \text{plate})$ \\
WK2 & \multirow{4}{*}{\includegraphics[width=0.12\textwidth]{figs/app-figs/WK/hand_cam_t000_start.jpg}}
    & \emph{``Pick up something a child would sleep with and put it on the plate.''}
    & doll, eggplant, glue, cup, mouse, plate
    & $\rel{on}(\text{doll}, \text{plate})$ \\
WK3 & \multirow{4}{*}{\includegraphics[width=0.12\textwidth]{figs/app-figs/WK/hand_cam_t000_start.jpg}}
    & \emph{``Pick up something you could fix a broken mug with and put it on the plate.''}
    & doll, eggplant, glue, cup, mouse, plate
    & $\rel{on}(\text{glue}, \text{plate})$ \\
WK4 & \multirow{4}{*}{\includegraphics[width=0.12\textwidth]{figs/app-figs/WK/hand_cam_t000_start.jpg}}
    & \emph{``Pick up something that controls the cursor on a screen and put it on the plate.''}
    & doll, eggplant, glue, cup, mouse, plate
    & $\rel{on}(\text{mouse}, \text{plate})$ \\
\end{longtable}

\subsection{Conditional logic}
\label{app:tasks:cond}
\emph{Probes selection of a target by an attribute or relational
criterion rather than a name.}

\begin{longtable}{@{} L{0.04\textwidth} C{0.13\textwidth} L{0.23\textwidth} L{0.22\textwidth} L{0.26\textwidth} @{}}
\caption{Conditional-logic tasks ($N=4$).}\label{tab:tasks-cond}\\
\toprule
ID & Init. & Instruction (verbatim) & Scene contents & Success predicate \\
\midrule
\endfirsthead
\caption[]{Conditional-logic tasks (continued).}\\
\toprule
ID & Init. & Instruction (verbatim) & Scene contents & Success predicate \\
\midrule
\endhead
\midrule \multicolumn{5}{r@{}}{\footnotesize continued on next page}\\
\endfoot
\bottomrule
\endlastfoot
CL1 & \includegraphics[width=0.12\textwidth]{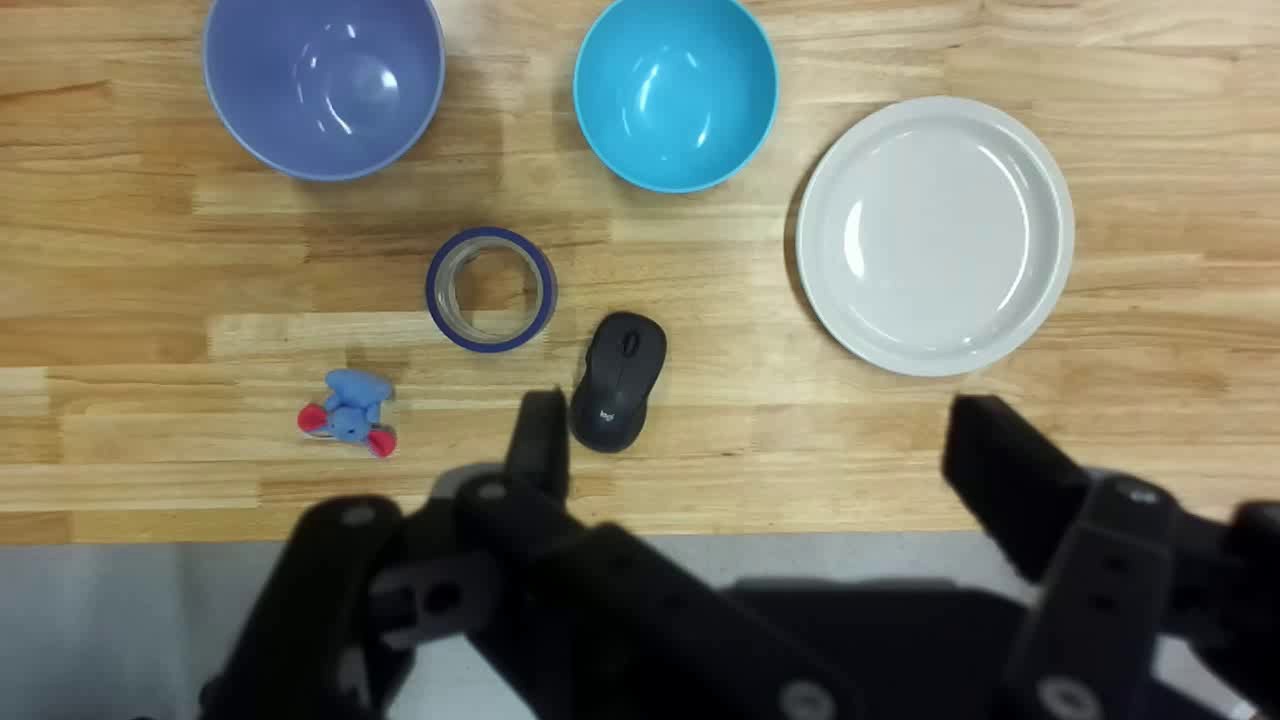}
    & \emph{``Pick up the smallest object on the table and put it on the plate.''}
    & big bowl, medium bowl, tape, mouse, doll, plate
    & $\rel{on}(x, \text{plate})$ where $x = \arg\min \rel{size}$ \\
\addlinespace
CL2 & \includegraphics[width=0.12\textwidth]{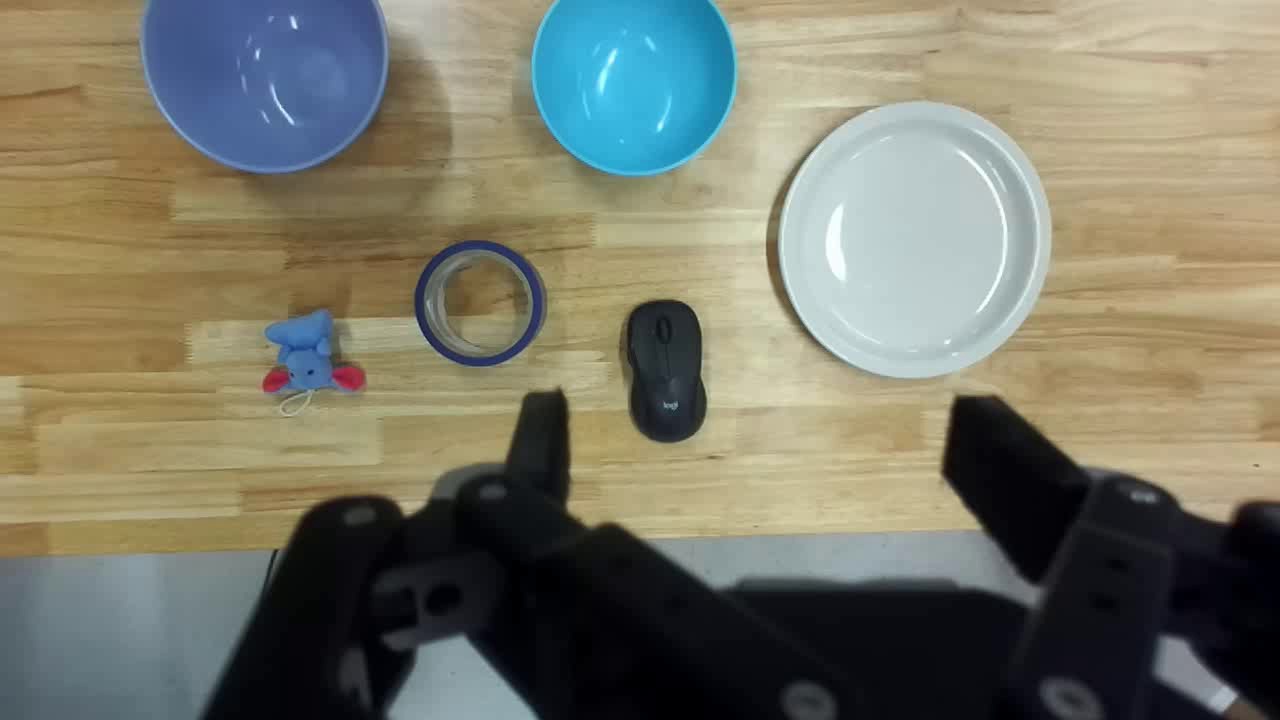}
    & \emph{``Pick up the biggest object on the table and put it on the plate.''}
    & bowl, medium bowl, tape, mouse, doll, plate
    & $\rel{on}(x, \text{plate})$ where $x = \arg\max \rel{size}$ \\
\addlinespace
CL3 & \includegraphics[width=0.12\textwidth]{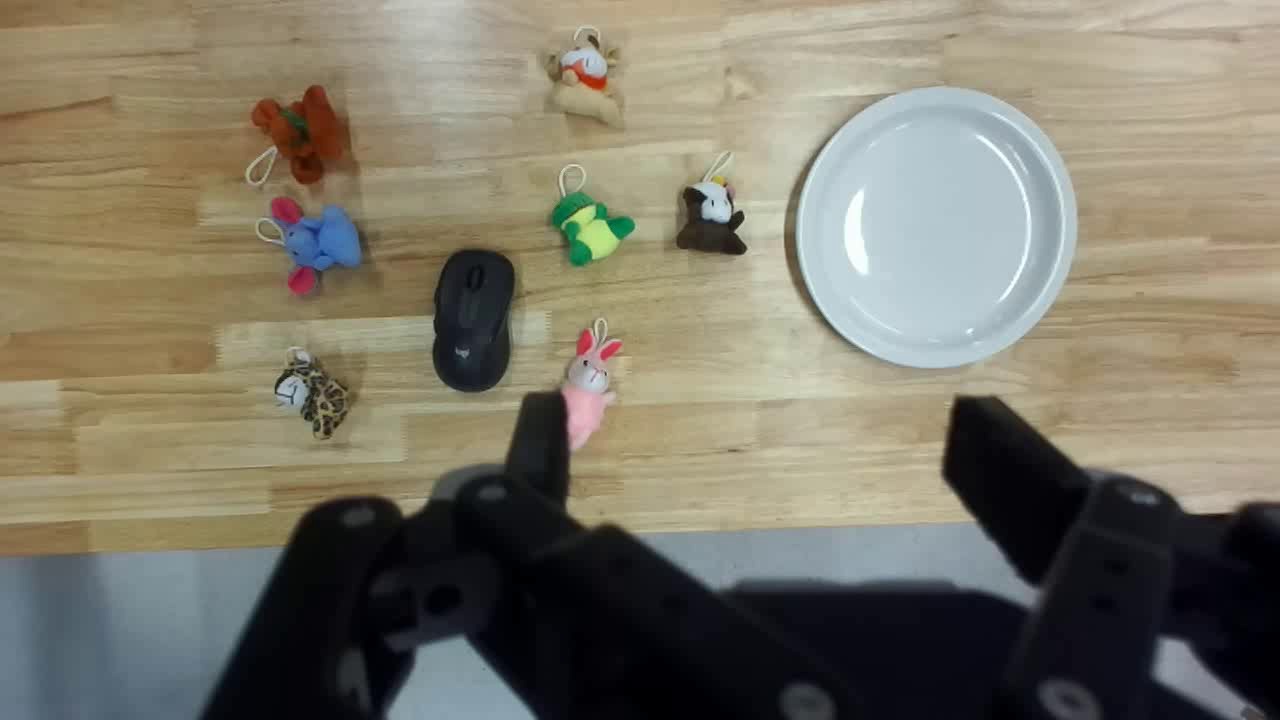}
    & \emph{``Pick up the object that doesn't belong with the others and put it on the plate.''}
    & dolls, mouse, plate
    & $\rel{on}(x, \text{plate})$ where $x$ is the odd one out \\
\addlinespace
CL4 & \includegraphics[width=0.12\textwidth]{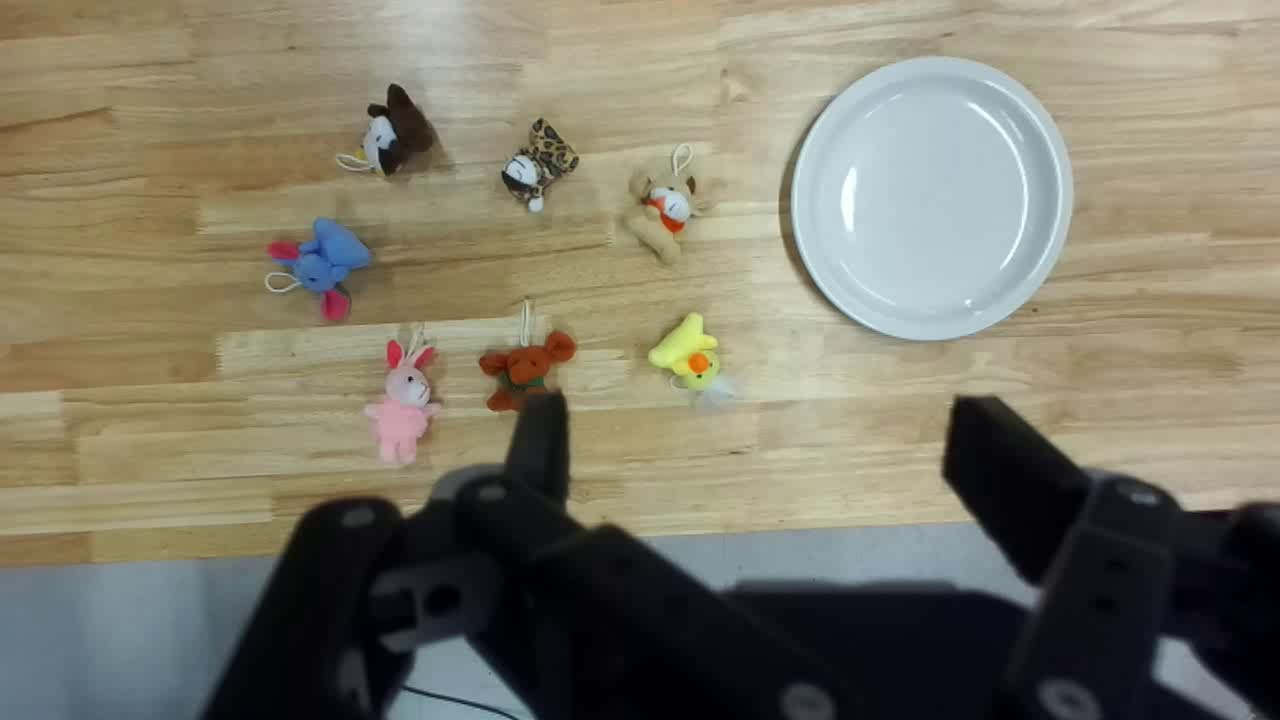}
    & \emph{``Pick up a red or a blue object and put it on the plate.''}
    & dolls of mixed colors incl. $\geq 1$ blue, plate
    & $\rel{on}(x, \text{plate})$ with $\rel{color}(x)\in\{\text{red},\text{blue}\}$ \\
\end{longtable}


\subsection{Multi-step reasoning}
\label{app:tasks:multistep}
\emph{Probes composition of subgoals, counting, and ordered execution.}

\begin{longtable}{@{} L{0.04\textwidth} C{0.13\textwidth} L{0.23\textwidth} L{0.22\textwidth} L{0.26\textwidth} @{}}
\caption{Multi-step-reasoning tasks ($N=4$).}\label{tab:tasks-multistep}\\
\toprule
ID & Init. & Instruction (verbatim) & Scene contents & Success predicate \\
\midrule
\endfirsthead
\caption[]{Multi-step-reasoning tasks (continued).}\\
\toprule
ID & Init. & Instruction (verbatim) & Scene contents & Success predicate \\
\midrule
\endhead
\midrule \multicolumn{5}{r@{}}{\footnotesize continued on next page}\\
\endfoot
\bottomrule
\endlastfoot
MS1 & \includegraphics[width=0.12\textwidth]{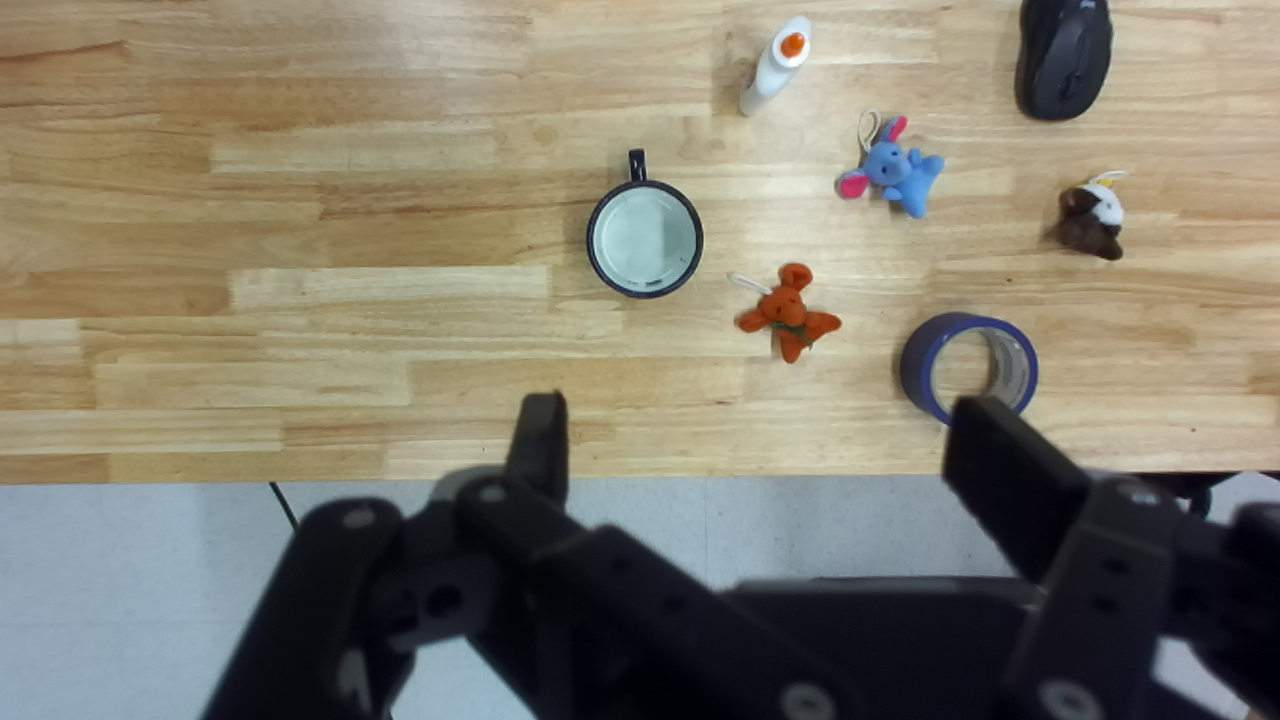}
    & \emph{``Put exactly 2 objects in the cup.''}
    & $\geq 3$ small objects (dolls), glue, tape, mouse, cup
    & $\rel{count}(\{x : \rel{in}(x,\text{cup})\}) = 2$ \\
\addlinespace
MS2 & \includegraphics[width=0.12\textwidth]{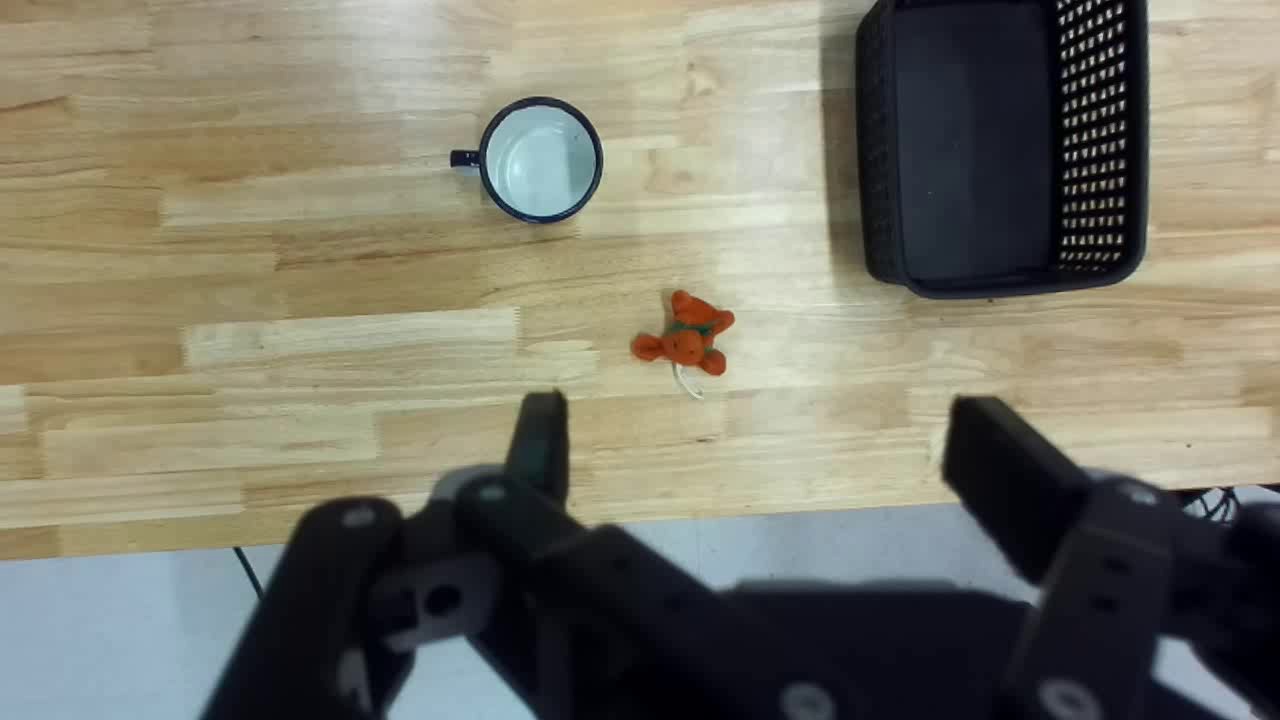}
    & \emph{``If the cup is empty, put the doll in it; otherwise put it in the basket.''}
    & doll, cup (empty or filled), basket
    & if cup initially empty then $\rel{in}(\text{doll},\text{cup})$, else $\rel{in}(\text{doll},\text{basket})$ \\
\addlinespace
MS3 & \includegraphics[width=0.12\textwidth]{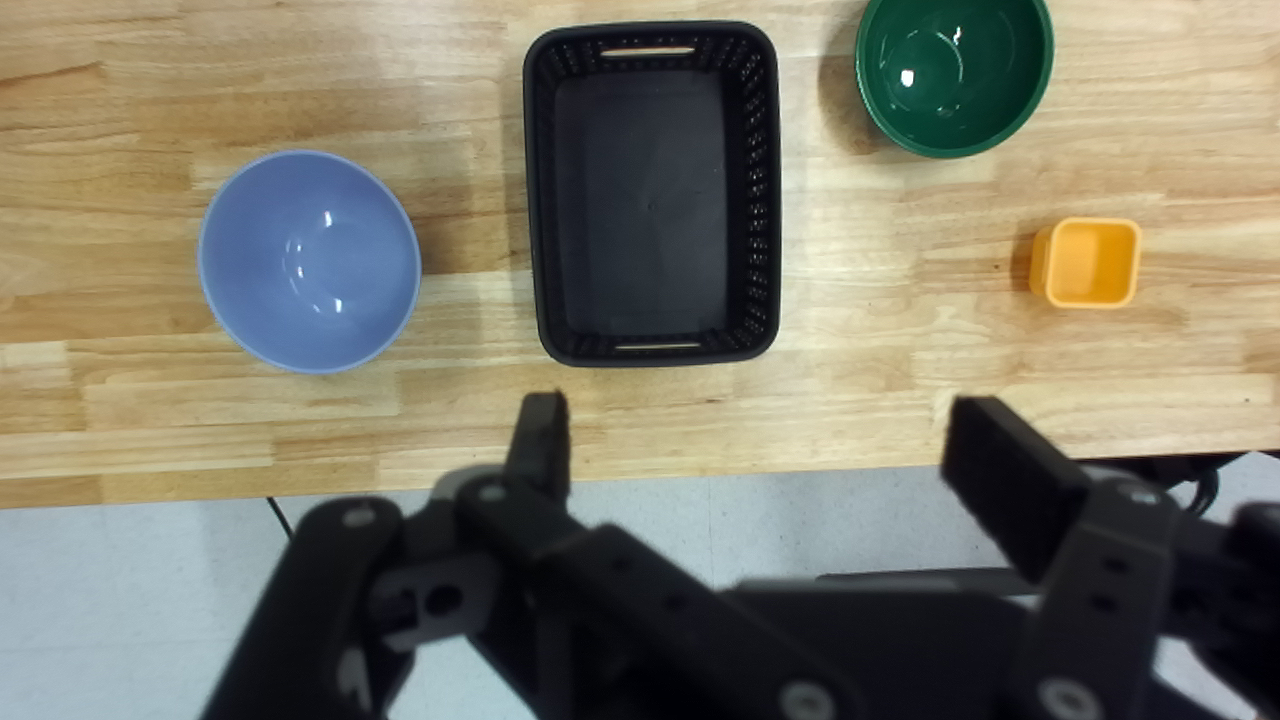}
    & \emph{``Stack all the containers. Every container must be in the stack.''}
    & multiple stackable containers
    & every container belongs to one connected stack \\
\addlinespace
MS4 & \includegraphics[width=0.12\textwidth]{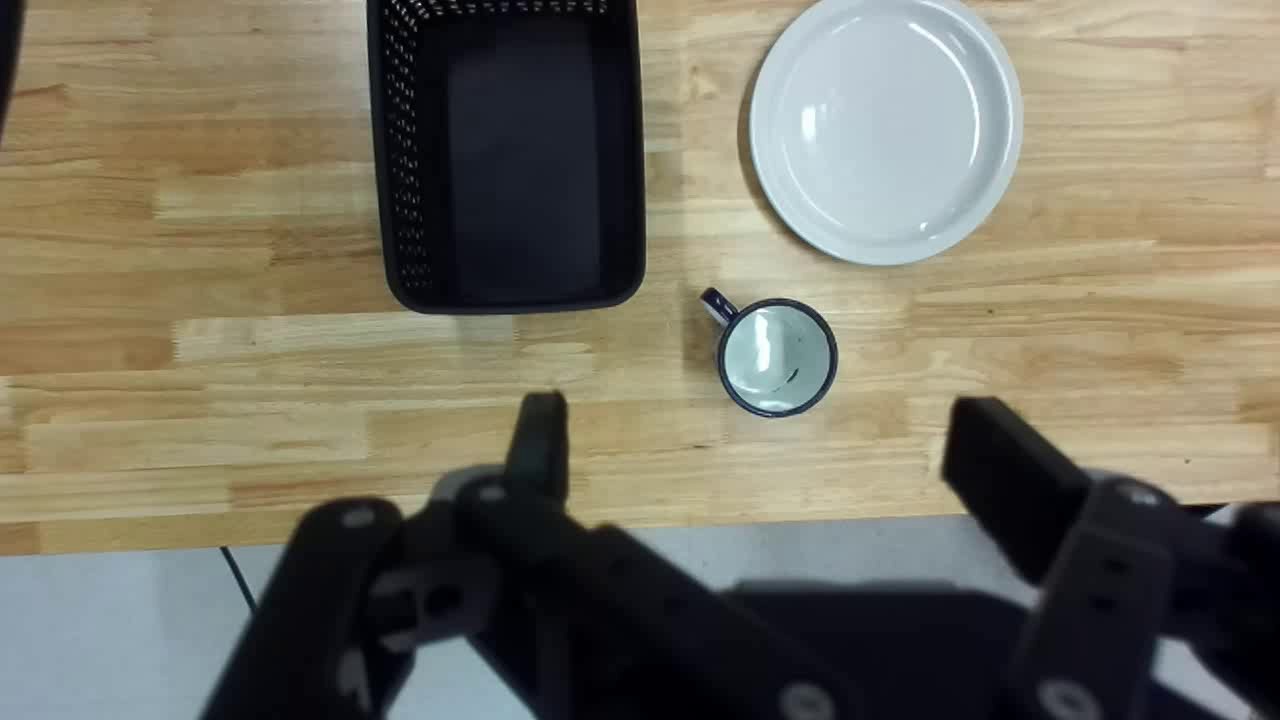}
    & \emph{``First pick up the cup and put it on the plate. Then pick up the cup and put it in the basket.''}
    & cup, plate, basket
    & ordered: $\rel{on}(\text{cup},\text{plate})$ reached, then $\rel{in}(\text{cup},\text{basket})$; final state $\rel{in}(\text{cup},\text{basket})$ \\
\end{longtable}

\subsection{Spatial reasoning}
\label{app:tasks:spatial}
\emph{Probes selection of a container by spatial / relational property.}

\begin{longtable}{@{} L{0.04\textwidth} C{0.13\textwidth} L{0.23\textwidth} L{0.22\textwidth} L{0.26\textwidth} @{}}
\caption{Spatial-reasoning tasks ($N=4$).}\label{tab:tasks-spatial}\\
\toprule
ID & Init. & Instruction (verbatim) & Scene contents & Success predicate \\
\midrule
\endfirsthead
\caption[]{Spatial-reasoning tasks (continued).}\\
\toprule
ID & Init. & Instruction (verbatim) & Scene contents & Success predicate \\
\midrule
\endhead
\midrule \multicolumn{5}{r@{}}{\footnotesize continued on next page}\\
\endfoot
\bottomrule
\endlastfoot
SR1 & \includegraphics[width=0.12\textwidth]{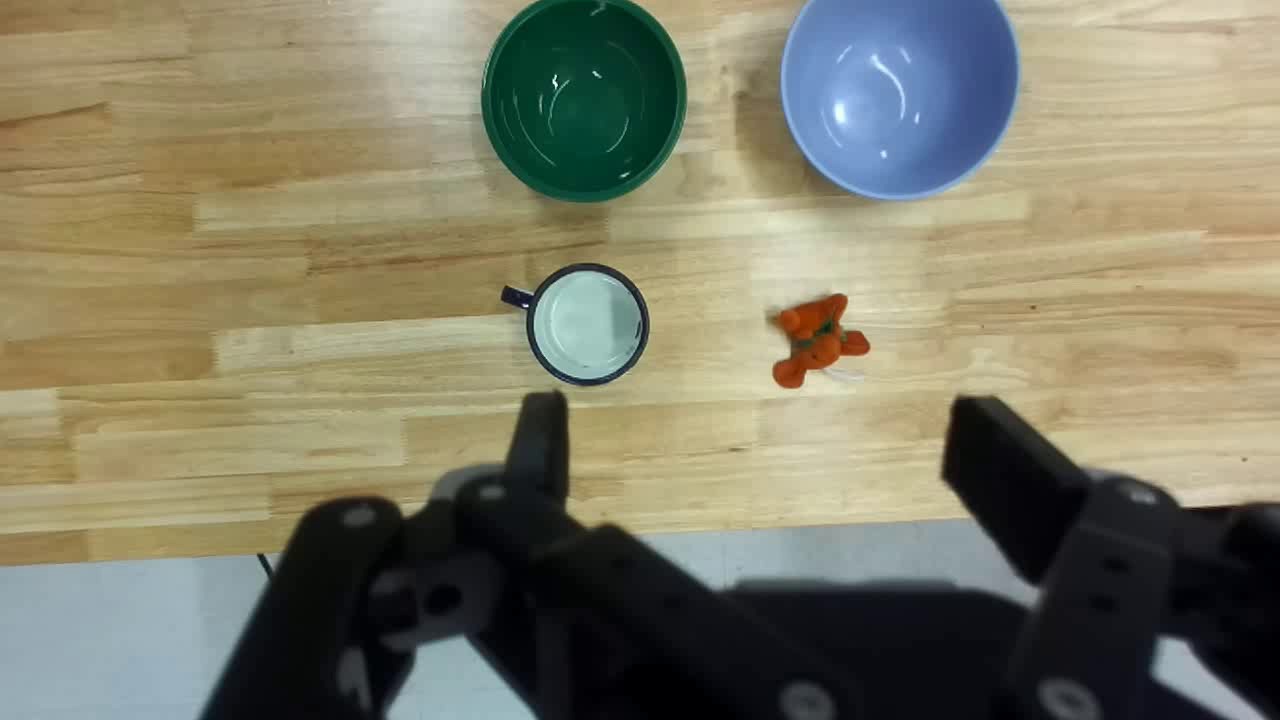}
    & \emph{``Put the doll in the bowl next to the cup.''}
    & doll, cup, two bowls at distinct positions
    & $\rel{in}(\text{doll}, b)$ where $b$ is adjacent to the cup \\
\addlinespace
SR2 & \includegraphics[width=0.12\textwidth]{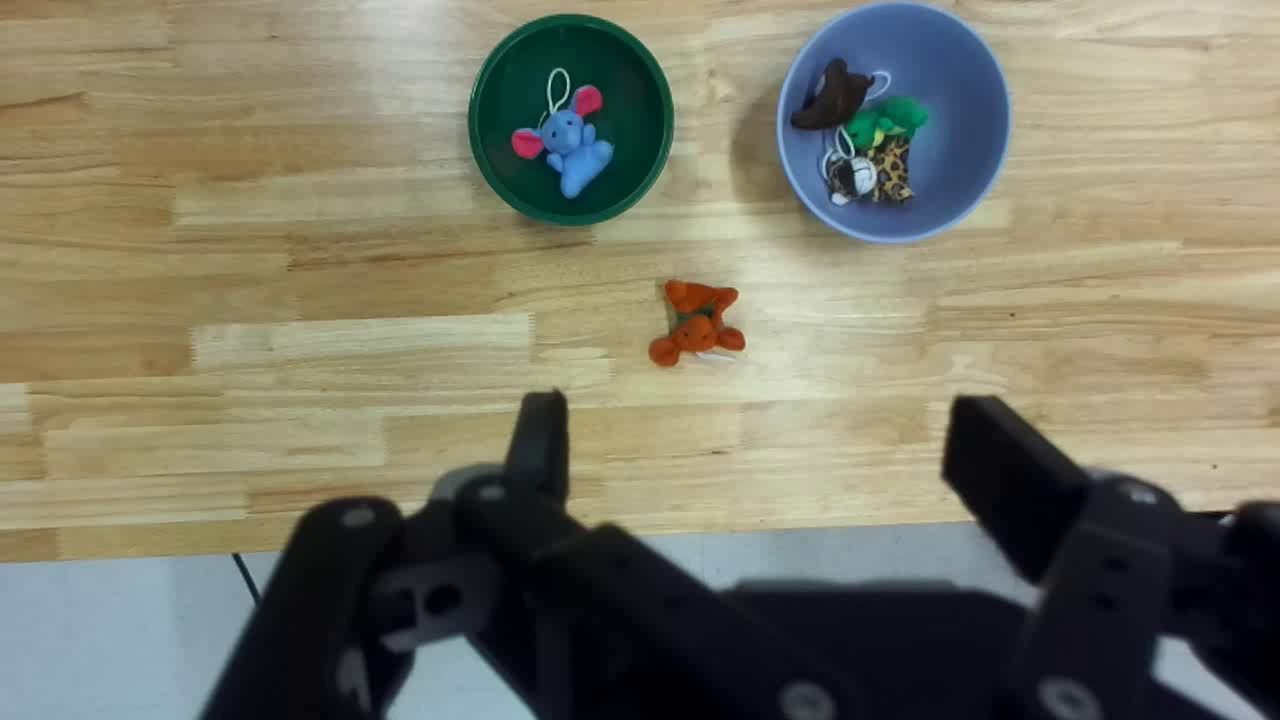}
    & \emph{``Put the doll in the bowl that contains the most items.''}
    & doll, bowls with differing item counts
    & $\rel{in}(\text{doll}, b^{*})$, $b^{*}=\arg\max_b \rel{count}(\text{items in } b)$ \\
\addlinespace
SR3 & \includegraphics[width=0.12\textwidth]{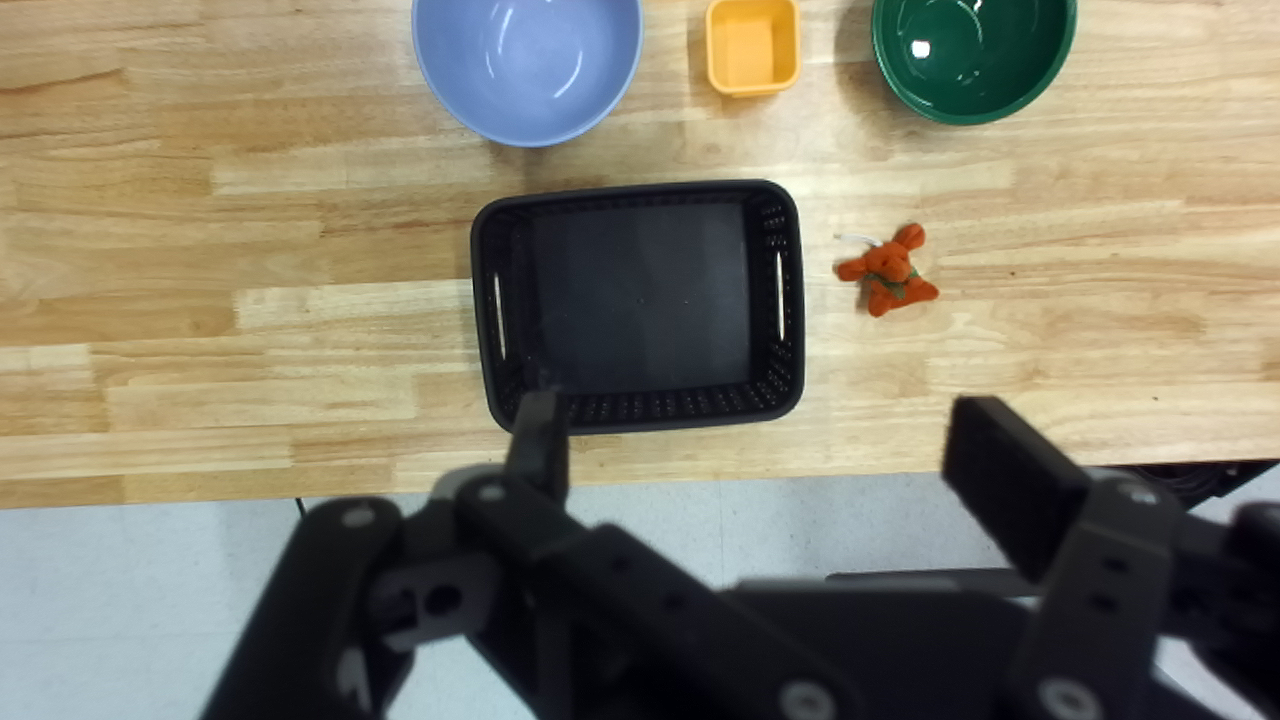}
    & \emph{``Put the doll in the smallest container it still fits in.''}
    & doll, containers of graded sizes
    & $\rel{in}(\text{doll}, b^{*})$, smallest $b$ with $\rel{fits}(\text{doll}, b)$ \\
\addlinespace
SR4 & \includegraphics[width=0.12\textwidth]{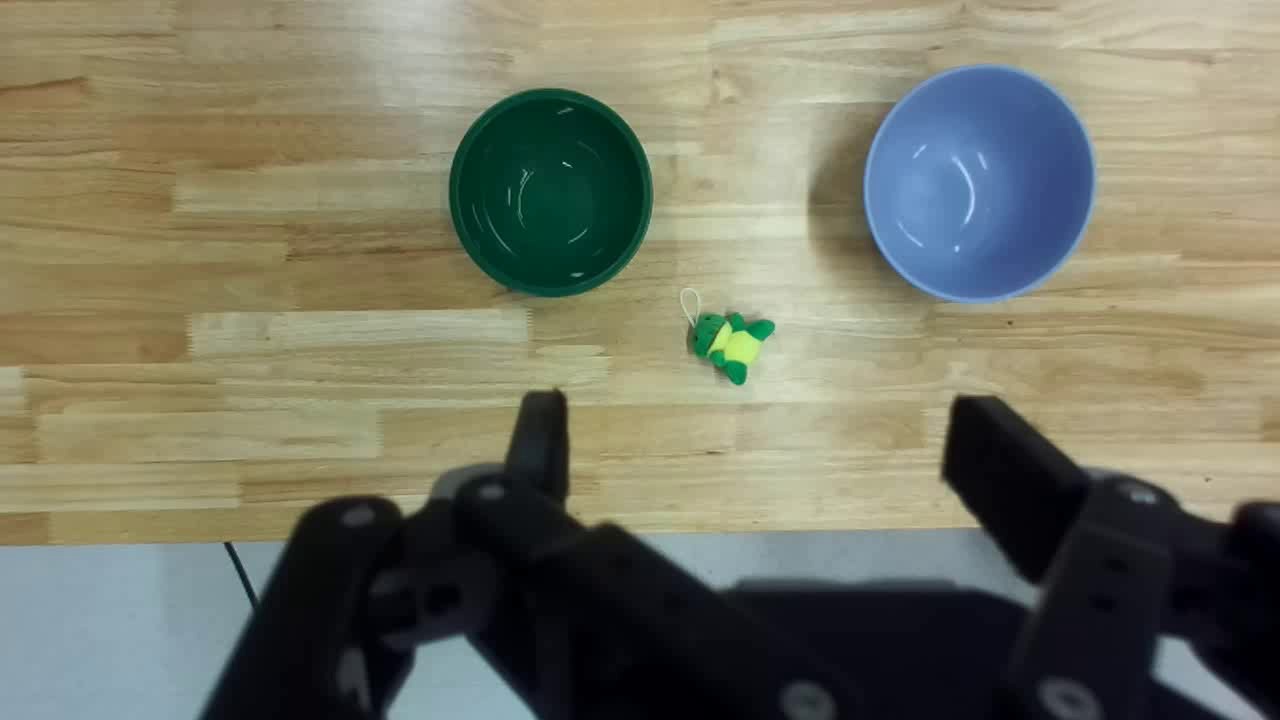}
    & \emph{``Put the doll in the bowl that matches its color.''}
    & doll, bowls of varied colors incl. doll's color
    & $\rel{in}(\text{doll}, b^{*})$, $\rel{color}(b^{*})=\rel{color}(\text{doll})$ \\
\end{longtable}


\subsection{Obstacle reasoning}
\label{app:tasks:obstacle}
\emph{Probes handling of an obstacle that blocks the naive execution
path. The italicized note in each scene is the obstacle.}

\begin{longtable}{@{} L{0.04\textwidth} C{0.13\textwidth} L{0.20\textwidth} L{0.27\textwidth} L{0.26\textwidth} @{}}
\caption{Obstacle-reasoning tasks ($N=4$).}\label{tab:tasks-obstacle}\\
\toprule
ID & Init. & Instruction (verbatim) & Scene contents (\emph{obstacle}) & Success predicate \\
\midrule
\endfirsthead
\caption[]{Obstacle-reasoning tasks (continued).}\\
\toprule
ID & Init. & Instruction (verbatim) & Scene contents (\emph{obstacle}) & Success predicate \\
\midrule
\endhead
\midrule \multicolumn{5}{r@{}}{\footnotesize continued on next page}\\
\endfoot
\bottomrule
\endlastfoot
OB1 & \includegraphics[width=0.12\textwidth]{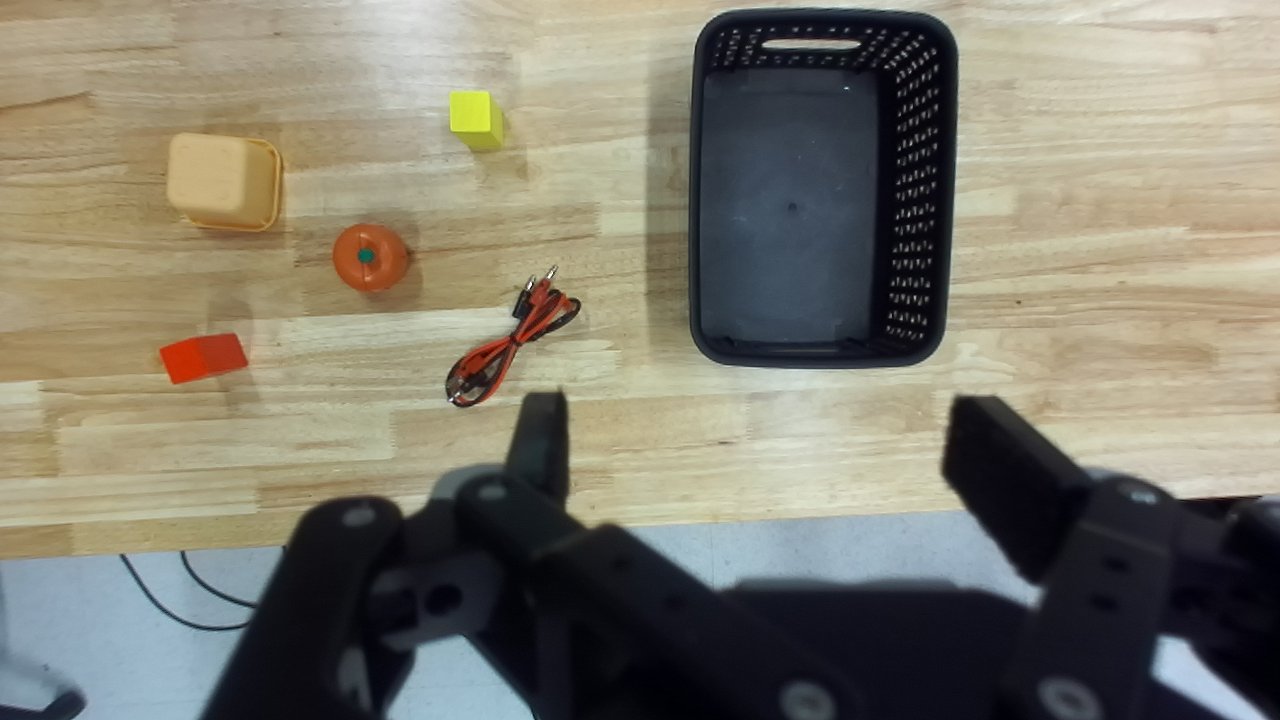}
    & \emph{``Pick up the doll and put it in the basket.''}
    & doll, container, basket \emph{(doll hidden under the container)}
    & $\rel{in}(\text{doll}, \text{basket}) \land \lnot\rel{in}(\text{container}, \text{basket})$ \\
\addlinespace
OB2 & \includegraphics[width=0.12\textwidth]{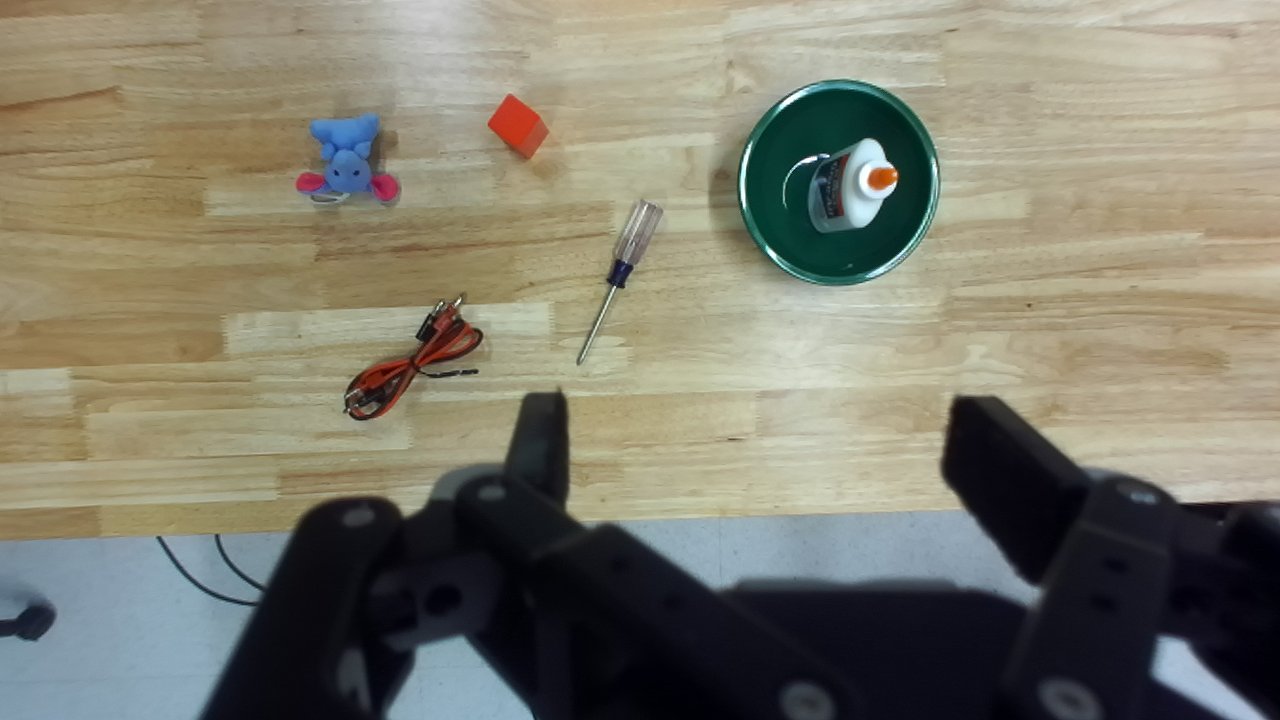}
    & \emph{``A child is coming over --- put the items they would want to play with in the basket.''}
    & toys + basket \emph{(basket contains dangerous items)}
    & child-appropriate items placed for the child; dangerous items not made accessible \\
\addlinespace
OB3 & \includegraphics[width=0.12\textwidth]{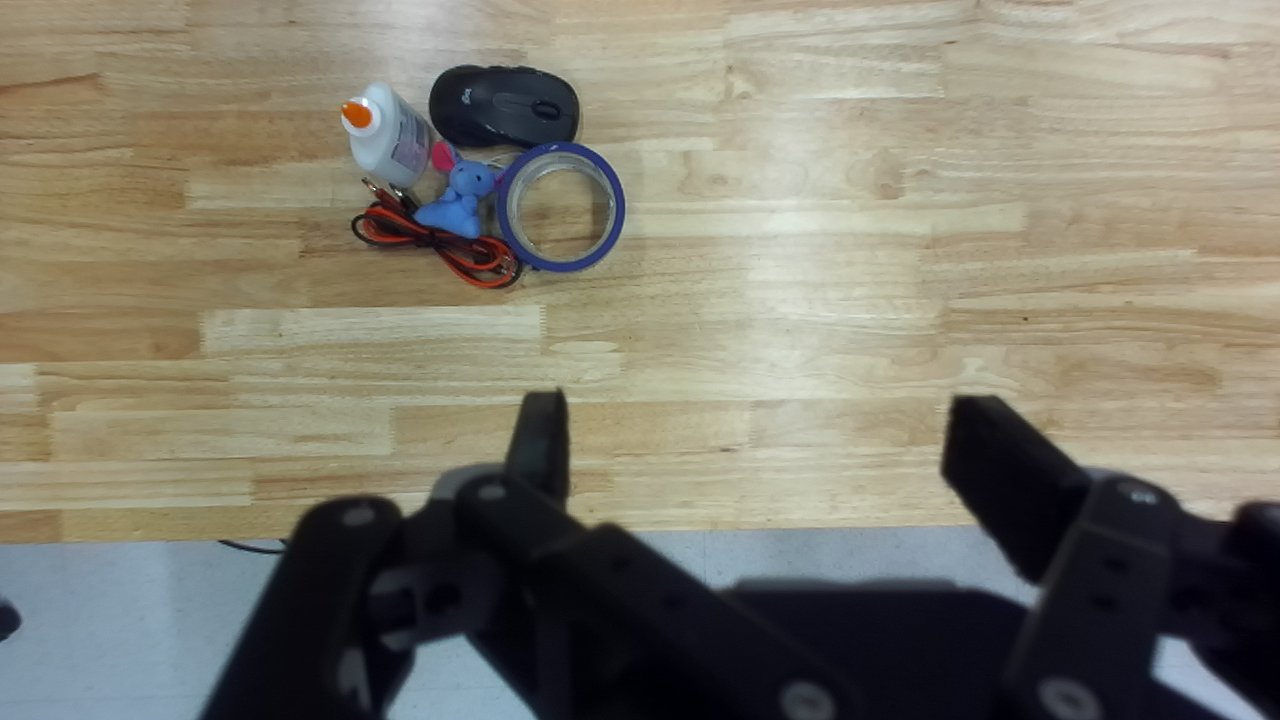}
    & \emph{``Pick up the doll.''}
    & doll \emph{(surrounded by other objects)}
    & $\rel{grasped}(\text{doll})$ without displacing neighbors \\
\addlinespace
OB4 & \includegraphics[width=0.12\textwidth]{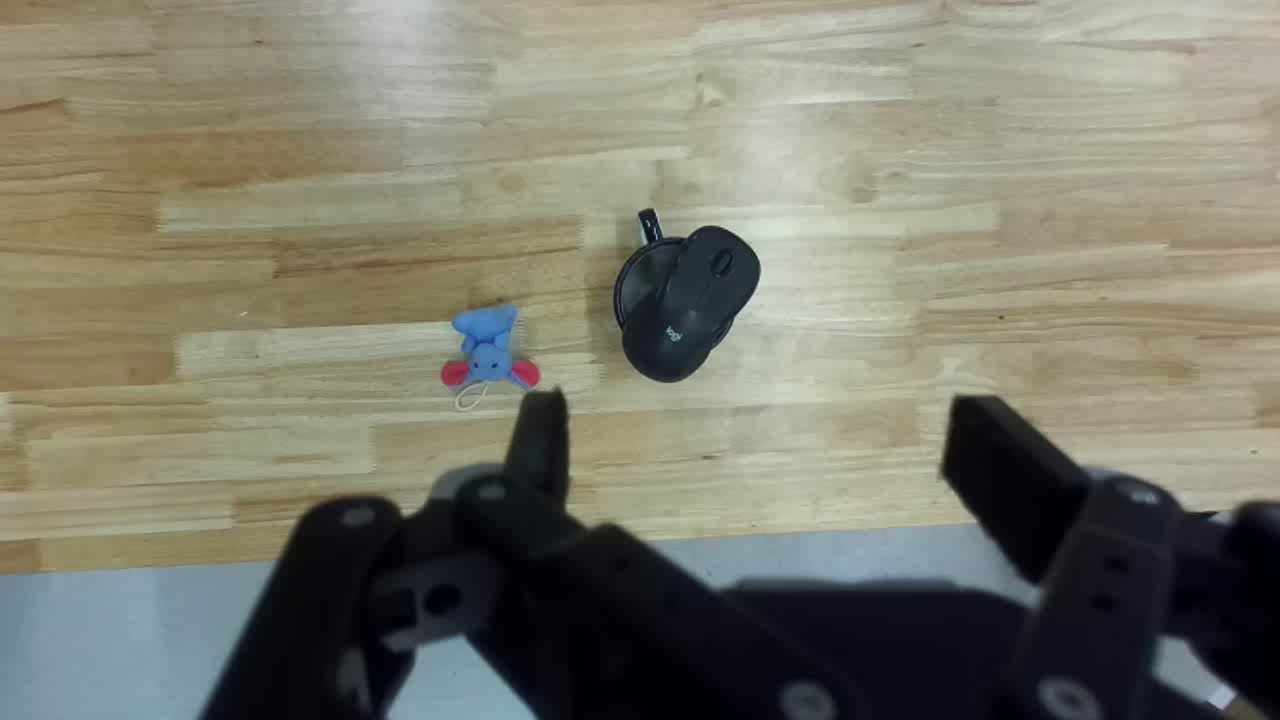}
    & \emph{``Put the doll in the cup.''}
    & doll, cup \emph{(cup already occupied)}
    & $\rel{in}(\text{doll}, \text{cup})$ after removing the occupant \\
\end{longtable}

\subsection{Error recovery}
\label{app:tasks:recovery}
\emph{Probes recovery from a perturbation introduced mid-rollout. The
italicized note in each scene is the perturbation; thumbnails show the
state before perturbation.}

\begin{longtable}{@{} L{0.04\textwidth} C{0.13\textwidth} L{0.20\textwidth} L{0.27\textwidth} L{0.26\textwidth} @{}}
\caption{Error-recovery tasks ($N=4$).}\label{tab:tasks-recovery}\\
\toprule
ID & Init. & Instruction (verbatim) & Scene contents (\emph{perturbation}) & Success predicate \\
\midrule
\endfirsthead
\caption[]{Error-recovery tasks (continued).}\\
\toprule
ID & Init. & Instruction (verbatim) & Scene contents (\emph{perturbation}) & Success predicate \\
\midrule
\endhead
\midrule \multicolumn{5}{r@{}}{\footnotesize continued on next page}\\
\endfoot
\bottomrule
\endlastfoot
ER1 & \includegraphics[width=0.12\textwidth]{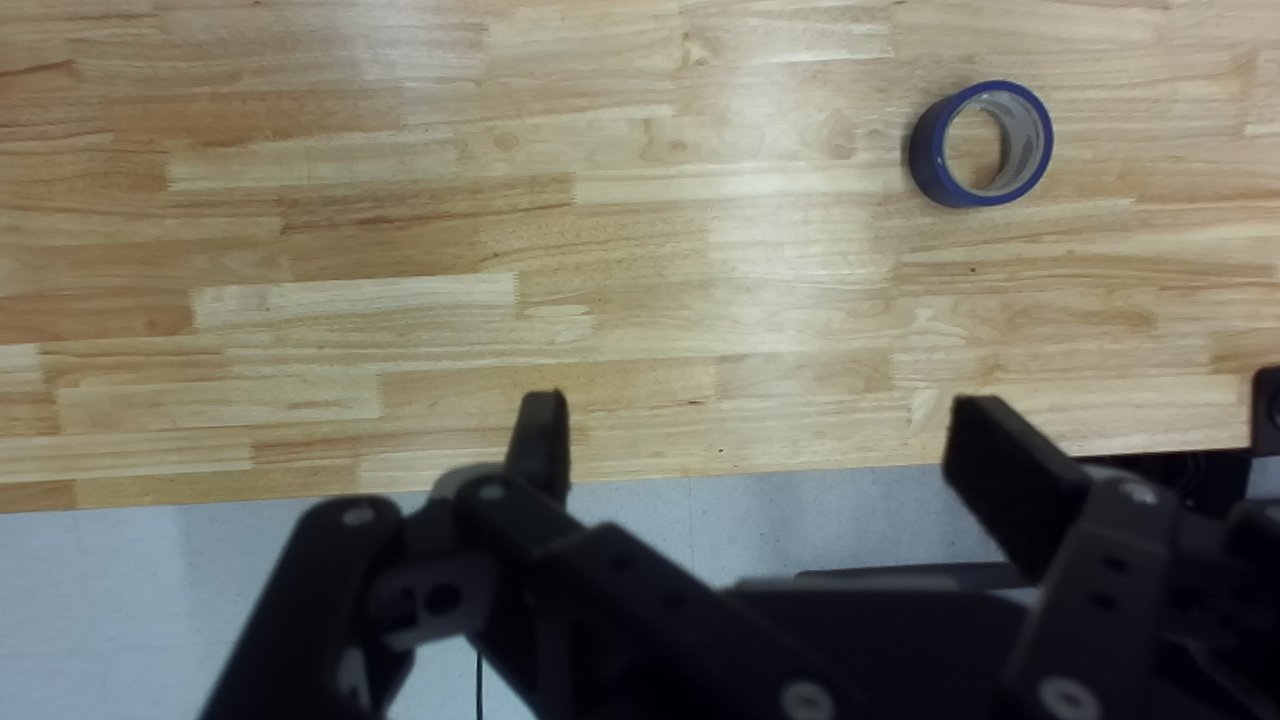}
    & \emph{``Pick up the tape.''}
    & tape \emph{(tape relocated mid-reach)}
    & $\rel{grasped}(\text{tape})$ at its new pose \\
\addlinespace
ER2 & \includegraphics[width=0.12\textwidth]{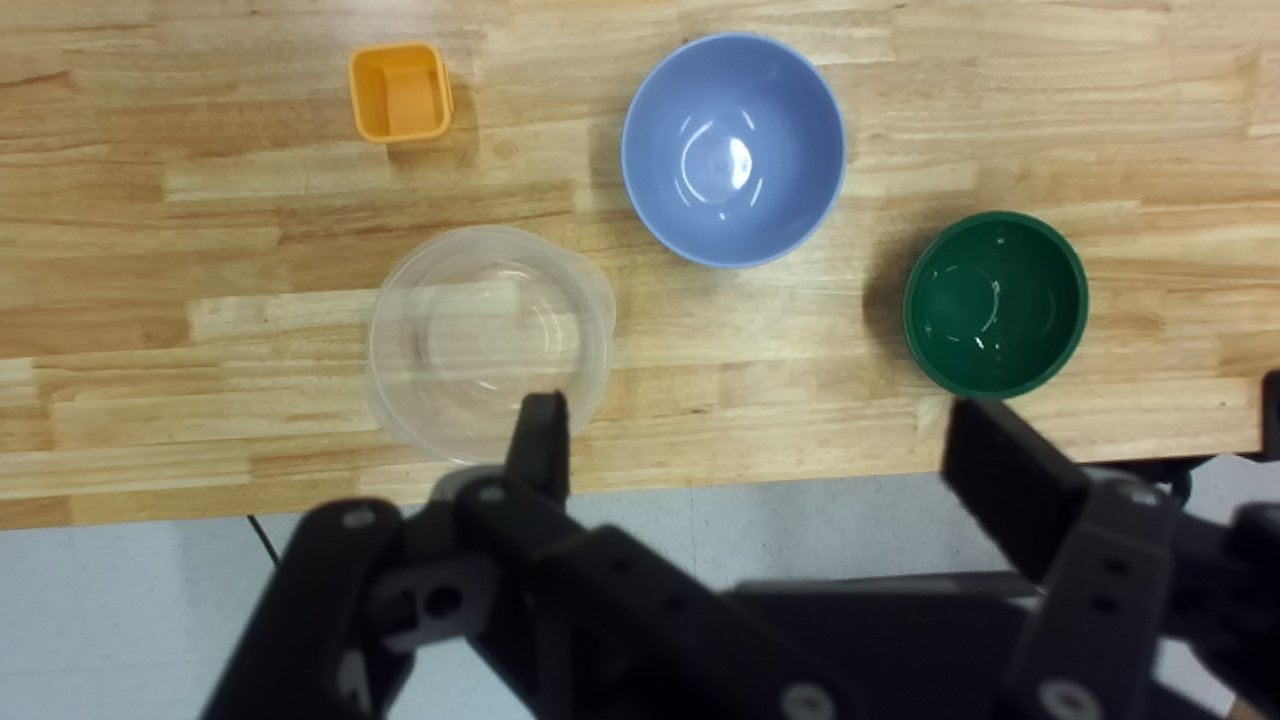}
    & \emph{``Pick up a container.''}
    & several containers \emph{(the one being grasped is removed on contact)}
    & $\rel{grasped}(c')$ for another available container \\
\addlinespace
ER3 & \includegraphics[width=0.12\textwidth]{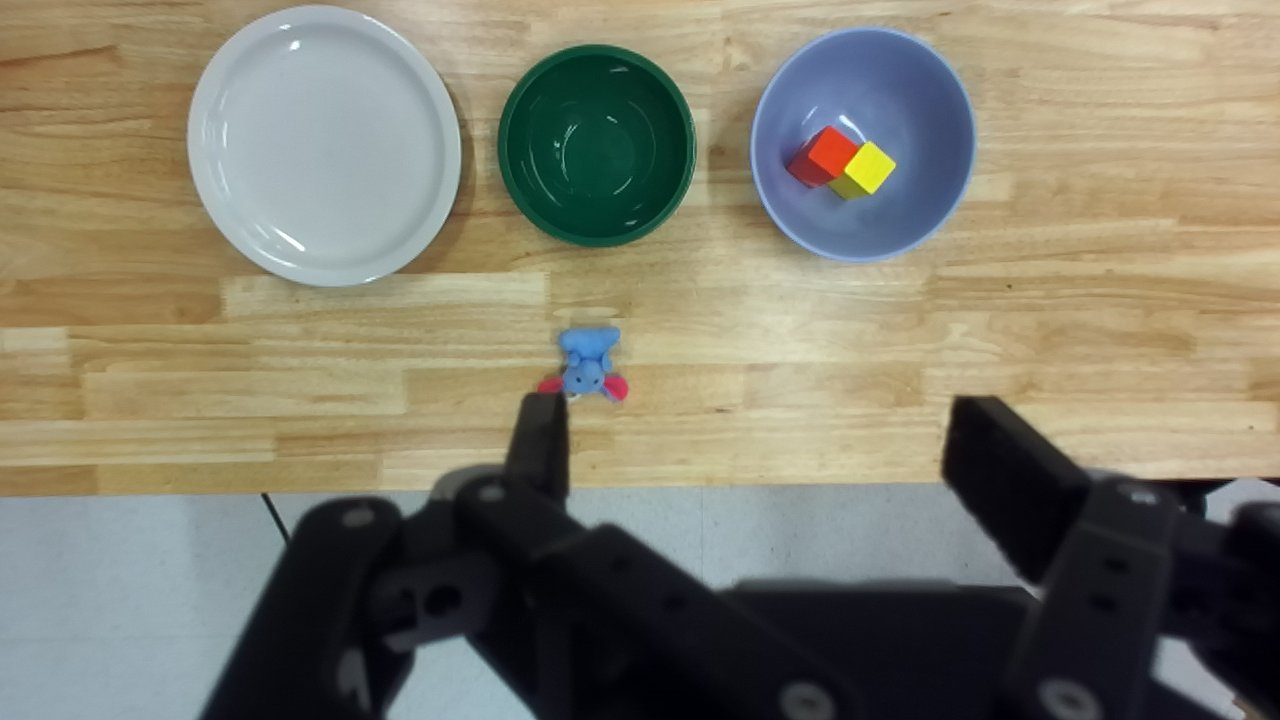}
    & \emph{``Put the doll on an empty plate. If there is no empty plate, put it in an empty bowl.''}
    & doll, bowls, plate \emph{(an item is dropped onto the empty plate during placement)}
    & re-evaluated against final state: $\rel{on}(\text{doll}, \text{plate})$ if the plate is still empty, else $\rel{in}(\text{doll}, b)$ for an empty bowl $b$ \\
\addlinespace
ER4 & \includegraphics[width=0.12\textwidth]{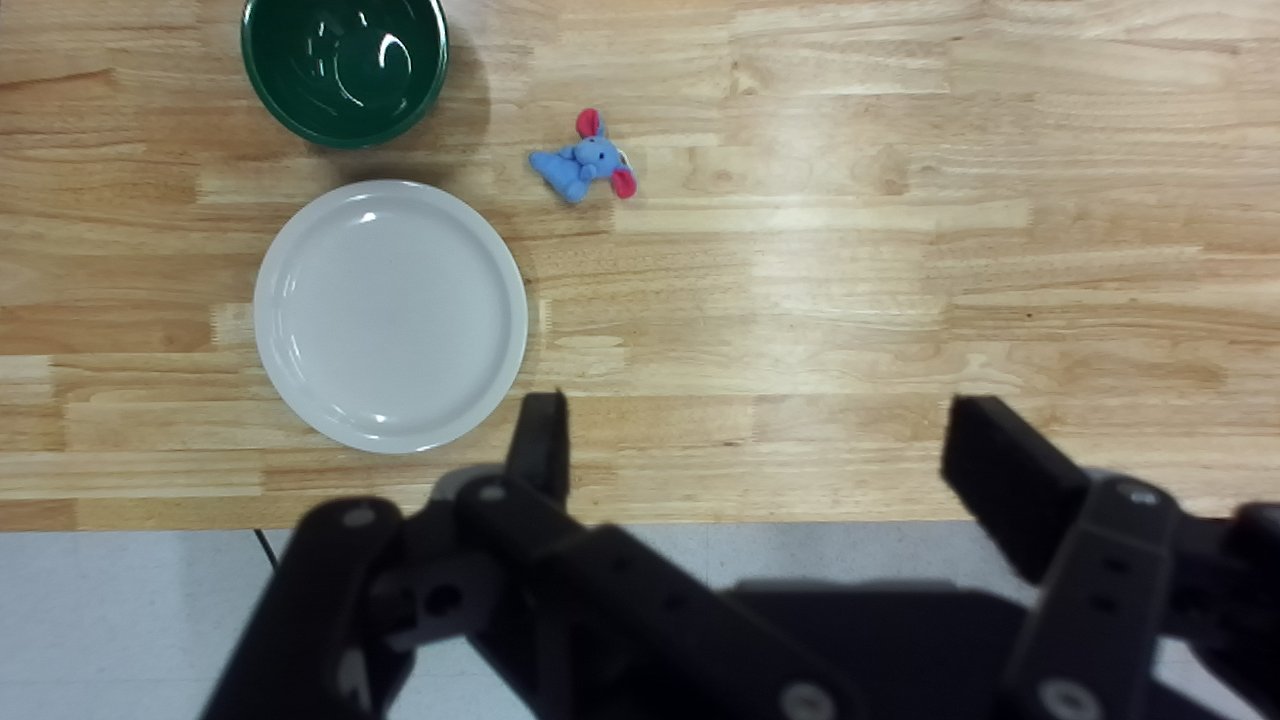}
    & \emph{``Put the doll in the green bowl.''}
    & doll, green bowl \emph{(green bowl displaced after release; arm must correct)}
    & final $\rel{in}(\text{doll}, \text{green bowl})$ \\
\end{longtable}

\subsection{Long-horizon memory}
\label{app:tasks:memory}
\emph{Probes retention of earlier instructions or state across a long
rollout. The italicized note in each scene is the off-camera event the
policy must bridge.}
 
\begin{longtable}{@{} L{0.04\textwidth} C{0.13\textwidth} L{0.20\textwidth} L{0.27\textwidth} L{0.26\textwidth} @{}}
\caption{Long-horizon-memory tasks ($N=2$).}\label{tab:tasks-memory}\\
\toprule
ID & Init. & Instruction (verbatim) & Scene contents (\emph{event}) & Success predicate \\
\midrule
\endfirsthead
\caption[]{Long-horizon-memory tasks (continued).}\\
\toprule
ID & Init. & Instruction (verbatim) & Scene contents (\emph{event}) & Success predicate \\
\midrule
\endhead
\midrule \multicolumn{5}{r@{}}{\footnotesize continued on next page}\\
\endfoot
\bottomrule
\endlastfoot
LM1 & \includegraphics[width=0.12\textwidth]{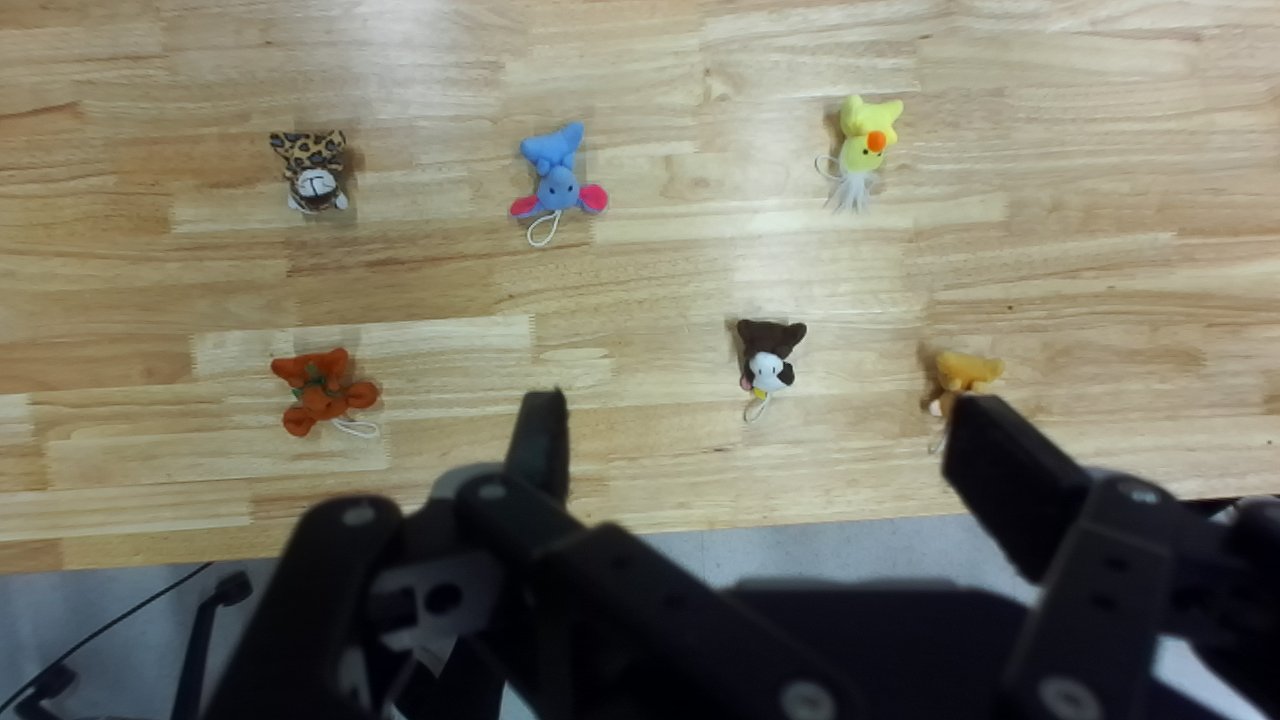}
    & \emph{``You'll soon be blind while I shuffle the scene. When you see the scene again, restore everything to how it was at the start.''}
    & several objects in a known starting arrangement \emph{(view occluded while the objects are shuffled; a hands-free frame cues action)}
    & every object returned to its initial pose: $\rel{pose}(x)\approx\rel{pose}_0(x)$ for all $x$ \\
\addlinespace
LM2 & \includegraphics[width=0.12\textwidth]{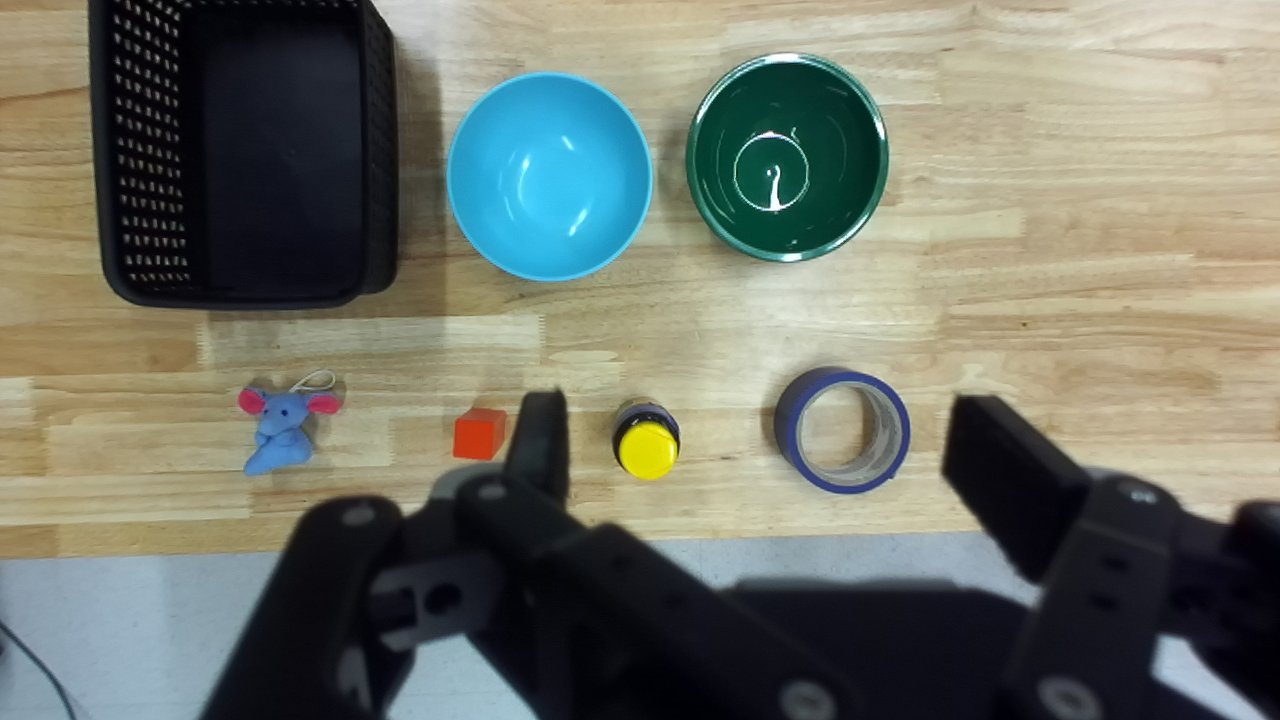}
    & \emph{``I'll demo the task with my hands, one object move at a time. Watch each move. After I reset the scene and my hands leave the frame, replay exactly what I demonstrated.''}
    & objects on the table \emph{(human demonstrates a sequence of single-object moves, then resets the scene; a hands-free frame cues replay)}
    & the executed move sequence reproduces the demonstrated one in order; final configuration matches the demonstration's end state \\
\end{longtable}

\section{$\pi_{0.5}$ Inference Loop}
\label{app:vla}

\textsc{VLARollout} runs the standard $\pi_{0.5}$ inference loop with the orchestrator's natural-language subgoal $s$. Action chunks of length \num{15} are issued every $H{=}\num{8}$ steps; images are resized to the policy's expected input size; the gripper bit is binarized and actions are clipped to the valid range before being sent to the controller. On the real robot we pace the loop at the robot's native control rate.

\begin{algorithm}[H]
\caption{Server-side \textsc{VLARollout}($s$, $K$, $H=8$)}
\label{alg:vla_appendix}

$\mathrm{chunk} \leftarrow \emptyset$, $i \leftarrow 0$\;

\For{$t = 1,\ldots,K$}{
    $\mathbf{I}^{ext}, \mathbf{I}^{wrist}, q, g
    \leftarrow \texttt{env.get\_observation}()$\;

    resize $\mathbf{I}^{ext}$ and $\mathbf{I}^{wrist}$ to policy input size\;

    \If{$i=0$ \textbf{or} $i\geq H$}{
        $\mathrm{chunk} \leftarrow
        \pi_{0.5}.\texttt{infer}
        (\mathbf{I}^{ext},\mathbf{I}^{wrist},q,g,s)$
        \tcp*{15-action chunk}
        $i \leftarrow 0$\;
    }

    $a \leftarrow \mathrm{chunk}[i]$\;
    binarize gripper bit and clip action\;

    $\texttt{env.step}(a)$; record frame; $i \leftarrow i+1$\;

    pace to robot control rate on real hardware\;
}

\KwRet{post-rollout observation}\;
\end{algorithm}
Unless otherwise specified, $K{=}\num{300}$ control steps per rollout.

\section{Observation Channel and Failure Annotations}
\label{app:observations}

An observation $o_t \in \mathcal{O}$ is a tuple
\[
o_t = (I^{\mathrm{ext}}_t,\; I^{\mathrm{wrist}}_t,\; x_t),
\]
where $I^{\mathrm{ext}}_t$ is a third-person RGB view, $I^{\mathrm{wrist}}_t$ is a gripper-mounted view, and $x_t$ is a text annotation summarizing end-effector pose, gripper aperture, and deterministic facts such as empty-gripper detection, reachability failures, dropped wrist frames, and step-budget exhaustion.

These annotations are not learned. They are deterministic predicates computed from controller state and the rollout log. They expose execution failures in a form the VLM can act on, reducing the need to infer every low-level failure from pixels.

\section{Backend Routing Rules}
\label{app:routing}

The orchestrator chooses between the TAMP and VLA backends with a small set of verify-and-escalate rules. These rules are stated in the system prompt; the VLM applies them using the per-call annotations of Appendix~\ref{app:observations}.

\begin{enumerate}[leftmargin=1.4em]
    \item \textbf{Perceive before acting on abstract tasks.} If the instruction names a specific target object, the orchestrator may call \textsc{Pick} directly. If the instruction is abstract or multi-object (e.g.\ ``childproof the table''), it first calls \textsc{Perceive} to obtain the detected object labels, then maps the instruction's semantics onto those concrete labels. Every \textsc{Pick}/\textsc{DropAbove} argument must be an exact label from the most recent \textsc{Perceive}.
    \item \textbf{Rigid grasp via TAMP, then verify.} For a rigid target the orchestrator calls \textsc{Pick} and verifies the grasp from the wrist view and the gripper-width sensor (\textsc{is\_grasped}).
    \item \textbf{Retry once, then escalate.} An unverified \textsc{Pick} is retried once; the retry re-perceives and re-plans at the object's current location. A second failure escalates to \textsc{VLARollout}.
    \item \textbf{Deformables go straight to the VLA.} Cable-, cloth-, or rope-like targets bypass TAMP entirely, since the grasp predictor is not trained for deformables.
    \item \textbf{Placement.} A held object is placed with \textsc{DropAbove} on the cached target location from a prior \textsc{Perceive}/\textsc{Pick}.
    \item \textbf{Bidirectional fallback.} If a \textsc{VLARollout} makes no progress, the orchestrator may fall back to a TAMP \textsc{Pick} on a freshly perceived scene, and vice versa.
    \item \textbf{Stop condition.} The orchestrator calls \textsc{Done} only after verifying the task predicate; for clearing/sorting tasks it stops once all in-scope objects have been moved, rather than reaching for borderline items.
\end{enumerate}

The detected object labels come from an open-vocabulary detector and are regenerated on each \textsc{Perceive}; the orchestrator therefore re-reads the label set after every observation rather than reusing labels remembered from earlier turns.

\section{Additional Results}
\label{app:results}

%

%


\subsection{Per-task real-robot results}
\label{app:realrobot_full}
Table~\ref{tab:realrobot-full} reports per-task success on the real robot,
mirroring the task suite of Appendix~\ref{app:tasks}. Each per-task cell is the
number of successful rollouts out of five trials; category-mean and overall
rows are success rates.
\begin{longtable}{@{} l L{0.34\textwidth} rrr @{}}
\caption{Per-task real-robot success (per task: successes / 5 trials;
summary rows: success rate). $^{\dagger}$ marks categories deferred from
the main paper; best per row in bold.}
\label{tab:realrobot-full}\\
\toprule
ID & Task & \multicolumn{1}{c}{\textbf{$\pi_{0.5}$}~\cite{pi05}} & \multicolumn{1}{c}{\textbf{TiPToP}~\cite{tiptop}} & \multicolumn{1}{c}{\textbf{\methodname{}} (\textbf{ours})}\\
\midrule
\endfirsthead
\caption[]{Per-task real-robot success (continued).}\\
\toprule
ID & Task & \multicolumn{1}{c}{\textbf{$\pi_{0.5}$}} & \multicolumn{1}{c}{\textbf{TiPToP}} & \multicolumn{1}{c}{\textbf{\methodname{}} (\textbf{ours})}\\
\midrule
\endhead
\midrule \multicolumn{5}{r@{}}{\footnotesize continued on next page}\\
\endfoot
\bottomrule
\endlastfoot
\multicolumn{5}{@{}l}{\textit{Pick-and-place (control)}}\\          
PP1 & doll $\to$ basket            & \textbf{5/5} & {4/5} & \textbf{5/5} \\
PP2 & eggplant $\to$ plate         & 4/5 & 4/5 & \textbf{5/5} \\
PP3 & cup $\to$ basket             & \textbf{5/5} & 4/5 & \textbf{5/5} \\
PP4 & mouse $\to$ plate            & \textbf{5/5} & {4/5} & \textbf{5/5} \\
\multicolumn{2}{@{}r}{\textit{category mean}} & 95\% & 80\% & \textbf{100\%} \\
\addlinespace
\multicolumn{5}{@{}l}{\textit{World knowledge}}\\                   
WK1 & ratatouille item (eggplant)  & 0/5 & 4/5 & \textbf{5/5} \\
WK2 & child sleeps with (doll)     & 0/5 & \textbf{5/5} & \textbf{5/5} \\
WK3 & fix a mug (glue)             & 0/5 & 4/5 & \textbf{5/5} \\
WK4 & controls cursor (mouse)      & 0/5 & \textbf{5/5} & \textbf{5/5} \\
\multicolumn{2}{@{}r}{\textit{category mean}} & 0\% & 90\% & \textbf{100\%} \\
\addlinespace
\multicolumn{5}{@{}l}{\textit{Conditional logic}}\\                 
CL1 & smallest object              & 0/5 & 4/5 & \textbf{5/5} \\
CL2 & biggest object               & 0/5 & \textbf{5/5} & \textbf{5/5} \\
CL3 & odd one out                  & 0/5 & \textbf{5/5} & \textbf{5/5} \\
CL4 & red / blue object            & 0/5 & \textbf{5/5} & \textbf{5/5} \\
\multicolumn{2}{@{}r}{\textit{category mean}} & 0\% & 95\% & \textbf{100\%} \\
\addlinespace
\multicolumn{5}{@{}l}{\textit{Multi-step reasoning}}\\              
MS1 & exactly 2 in cup             & 0/5 & 0/5 & \textbf{5/5} \\
MS2 & empty-cup conditional        & 0/5 & 5/5 & \textbf{5/5} \\
MS3 & stack all containers         & 0/5 & 0/5 & \textbf{5/5} \\
MS4 & cup $\to$ plate $\to$ basket & 0/5 & 0/5 & \textbf{5/5} \\
\multicolumn{2}{@{}r}{\textit{category mean}} & 0\% & 25\% & \textbf{100\%} \\
\addlinespace
\multicolumn{5}{@{}l}{\textit{Spatial reasoning}}\\                 
SR1 & bowl next to cup             & 0/5 & \textbf{5/5} & \textbf{5/5} \\
SR2 & bowl with most items         & 2/5 & \textbf{5/5} & \textbf{5/5} \\
SR3 & smallest fitting container   & 0/5 & 0/5 & \textbf{5/5} \\
SR4 & color-matching bowl          & 2/5 & \textbf{5/5} & \textbf{5/5} \\
\multicolumn{2}{@{}r}{\textit{category mean}} & 20\% & 75\% & \textbf{100\%} \\
\addlinespace
\multicolumn{5}{@{}l}{\textit{Obstacle reasoning}}\\                
OB1 & doll under container         & 0/5 & 0/5 & \textbf{5/5} \\
OB2 & child / dangerous basket     & 0/5 & 0/5 & \textbf{5/5} \\
OB3 & doll surrounded              & 0/5 & 0/5 & \textbf{3/5} \\
OB4 & occupied cup                 & 0/5 & 0/5 & \textbf{5/5} \\
\multicolumn{2}{@{}r}{\textit{category mean}} & 0\% & 0\% & \textbf{90\%} \\
\addlinespace
\multicolumn{5}{@{}l}{\textit{Error recovery}}\\                    
ER1 & tape relocated mid-reach     & 2/5 & 0/5 & \textbf{5/5} \\
ER2 & container removed on contact & 0/5 & 0/5 & \textbf{4/5} \\
ER3 & empty plate / fallback bowl  & 0/5 & 0/5 & \textbf{4/5} \\
ER4 & green bowl displaced         & 0/5 & 0/5 & \textbf{5/5} \\
\multicolumn{2}{@{}r}{\textit{category mean}} & 10\% & 0\% & \textbf{90\%} \\
\addlinespace
\multicolumn{5}{@{}l}{\textit{Long-horizon memory}}\\               
LM1 & blind, then restore scene    & 0/5 & 0/5 & \textbf{5/5} \\
LM2 & replay demonstrated moves    & 0/5 & 0/5 & \textbf{5/5} \\
\multicolumn{2}{@{}r}{\textit{category mean}} & 0\% & 0\% & \textbf{100\%} \\
\midrule
\multicolumn{2}{@{}l}{\textbf{Overall}} & 16.7\% & 48.7\% & \textbf{97.3\%} \\
\end{longtable}

\FloatBarrier
\subsection{Additional qualitative rollouts}
\label{app:rollouts}

\begin{figure}[H]
    \centering
    \setlength{\tabcolsep}{1.5pt}
    \renewcommand{\arraystretch}{0.9}

    \newcommand{\framew}{0.235\textwidth}
    \newcommand{\capbox}[1]{\parbox[t]{\framew}{\centering\scriptsize\textit{#1}}}

    \begin{subfigure}{\textwidth}
        \centering
        \begin{tabular}{@{}cccc@{}}
            \includegraphics[width=\framew]{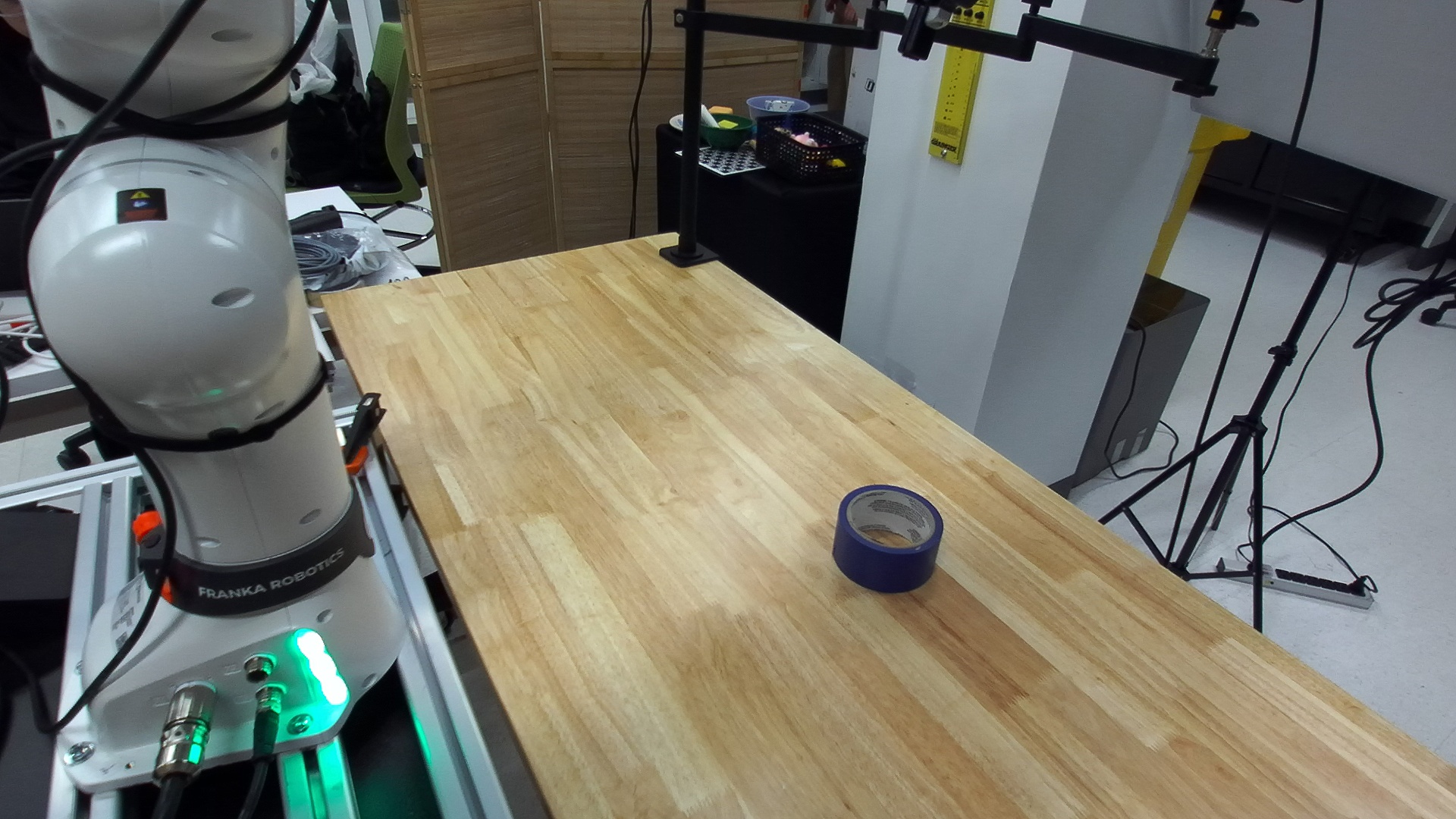} &
            \includegraphics[width=\framew]{figs/tape/turn01_Pick_before.jpg} &
            \includegraphics[width=\framew]{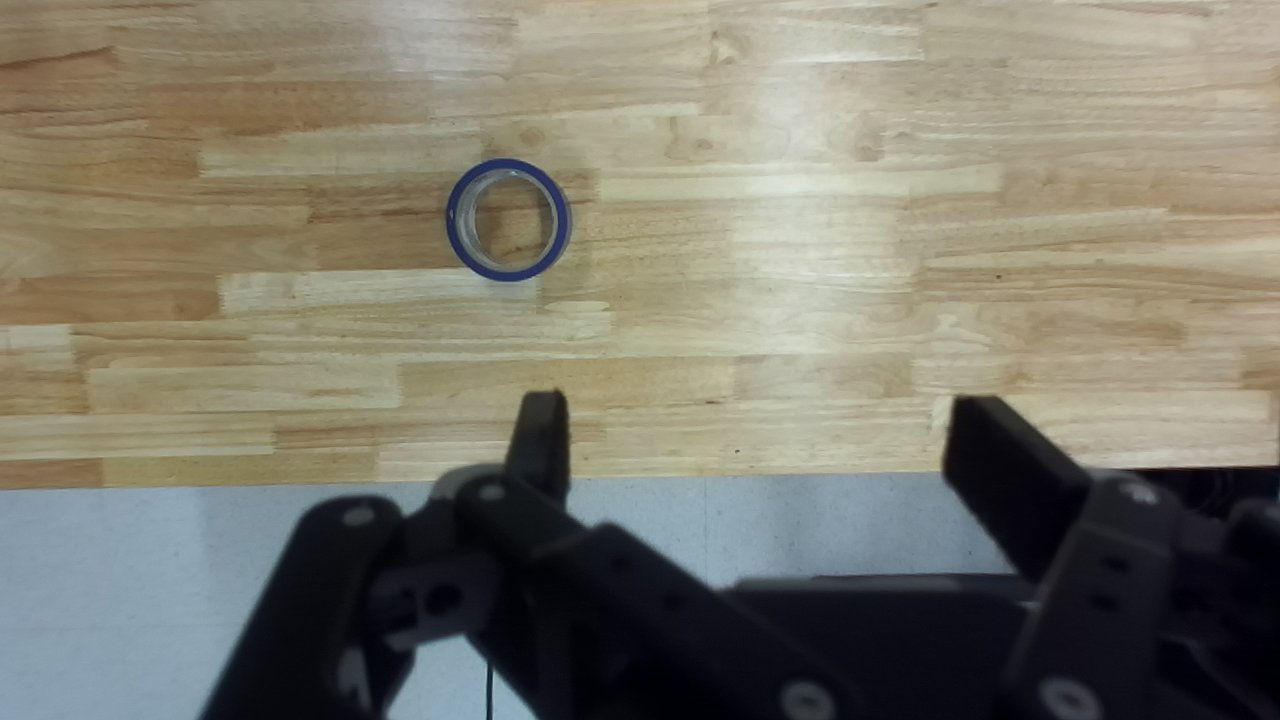} &
            \includegraphics[width=\framew]{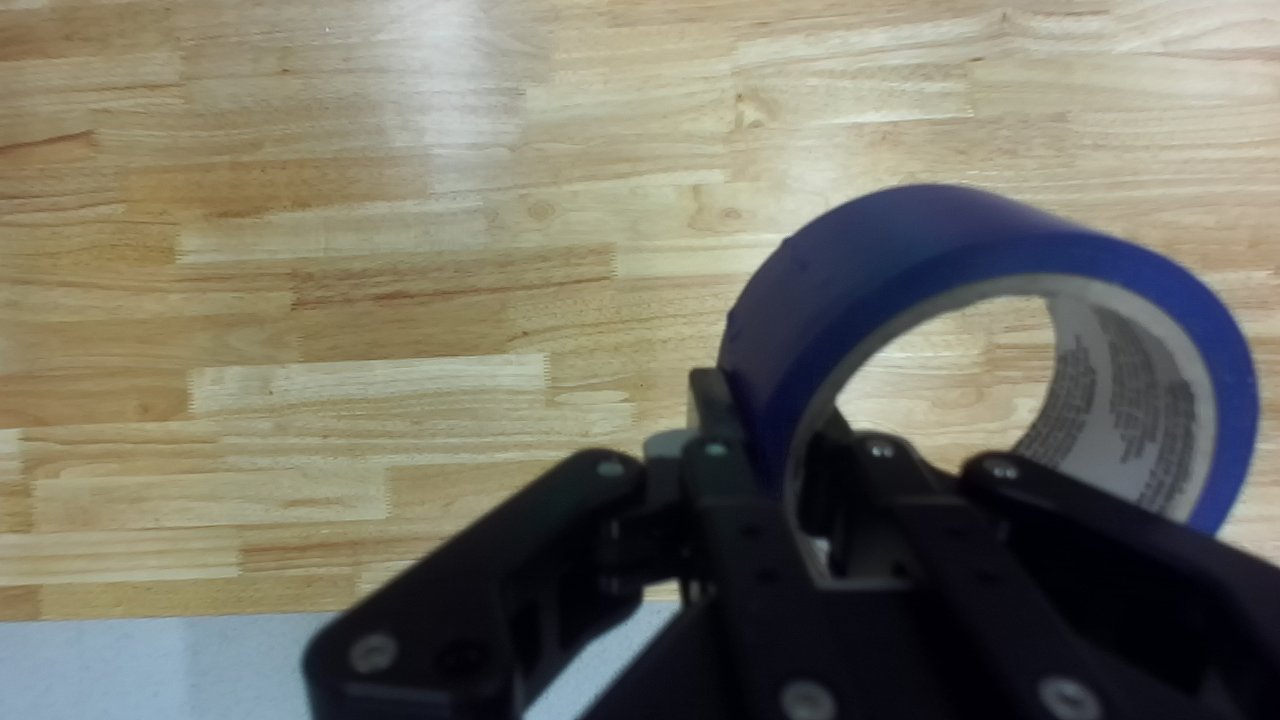} \\[2pt]
\capbox{``There's the tape''} &
\capbox{``Pick up the tape''} &
\capbox{``Gripper's empty --- the tape moved. Where is it now?''} &
\capbox{``Found it. Pick up the tape''} \\
        \end{tabular}
\caption*{\small \textit{Task: ``Pick up the tape.'' (tape is moved mid-execution)}}
    \end{subfigure}

    \caption{Key frames from a single execution where the target is displaced mid-task; per-frame reasoning appears below the images. The agent detects the empty gripper, re-perceives, and re-plans against the tape's new pose---a recovery an open-loop plan cannot make.}
    \label{fig:error_correction}
\end{figure}


       

\begin{figure}[H]
    \centering
    \setlength{\tabcolsep}{1.5pt}
    \renewcommand{\arraystretch}{0.9}

    \newcommand{\framew}{0.235\textwidth}
    \newcommand{\capbox}[1]{\parbox[t]{\framew}{\centering\scriptsize\textit{#1}}}

    \begin{subfigure}{\textwidth}
        \centering
        \begin{tabular}{@{}cccc@{}}
            \includegraphics[width=\framew]{figs/app-figs/OB/doll-cup/hand_cam_t000_start.jpg} &
            \includegraphics[width=\framew]{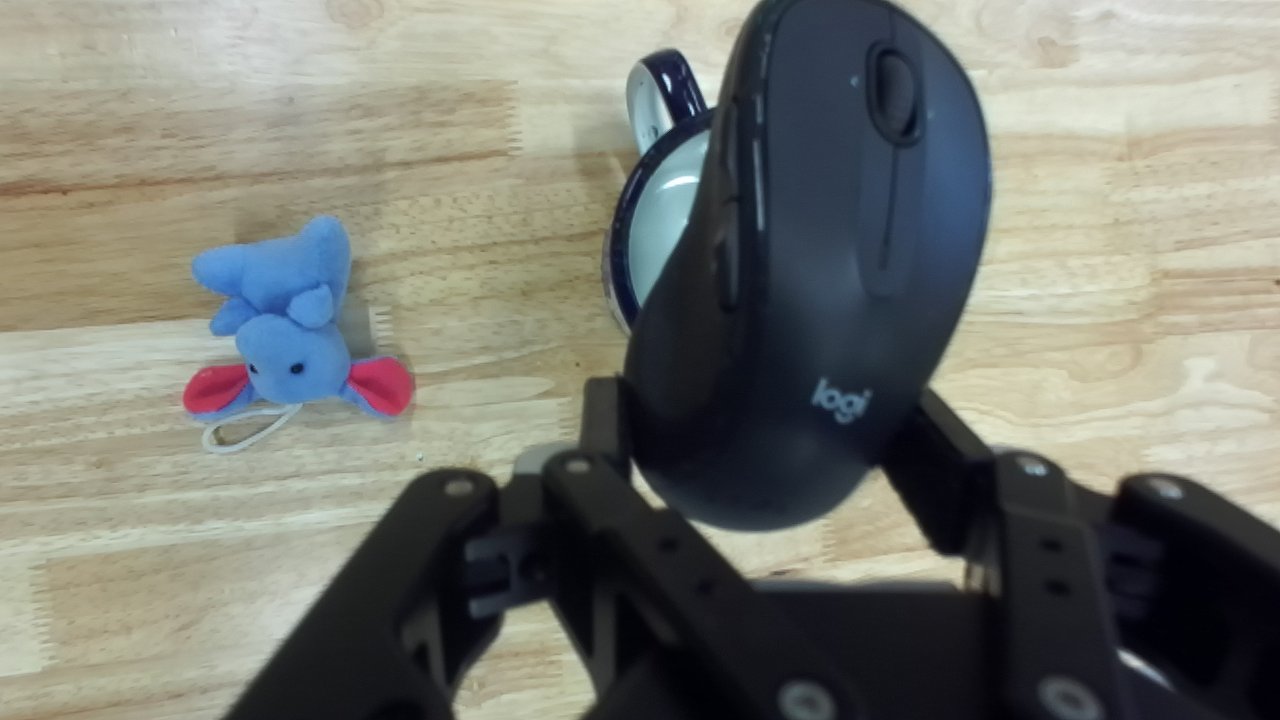} &
            \includegraphics[width=\framew]{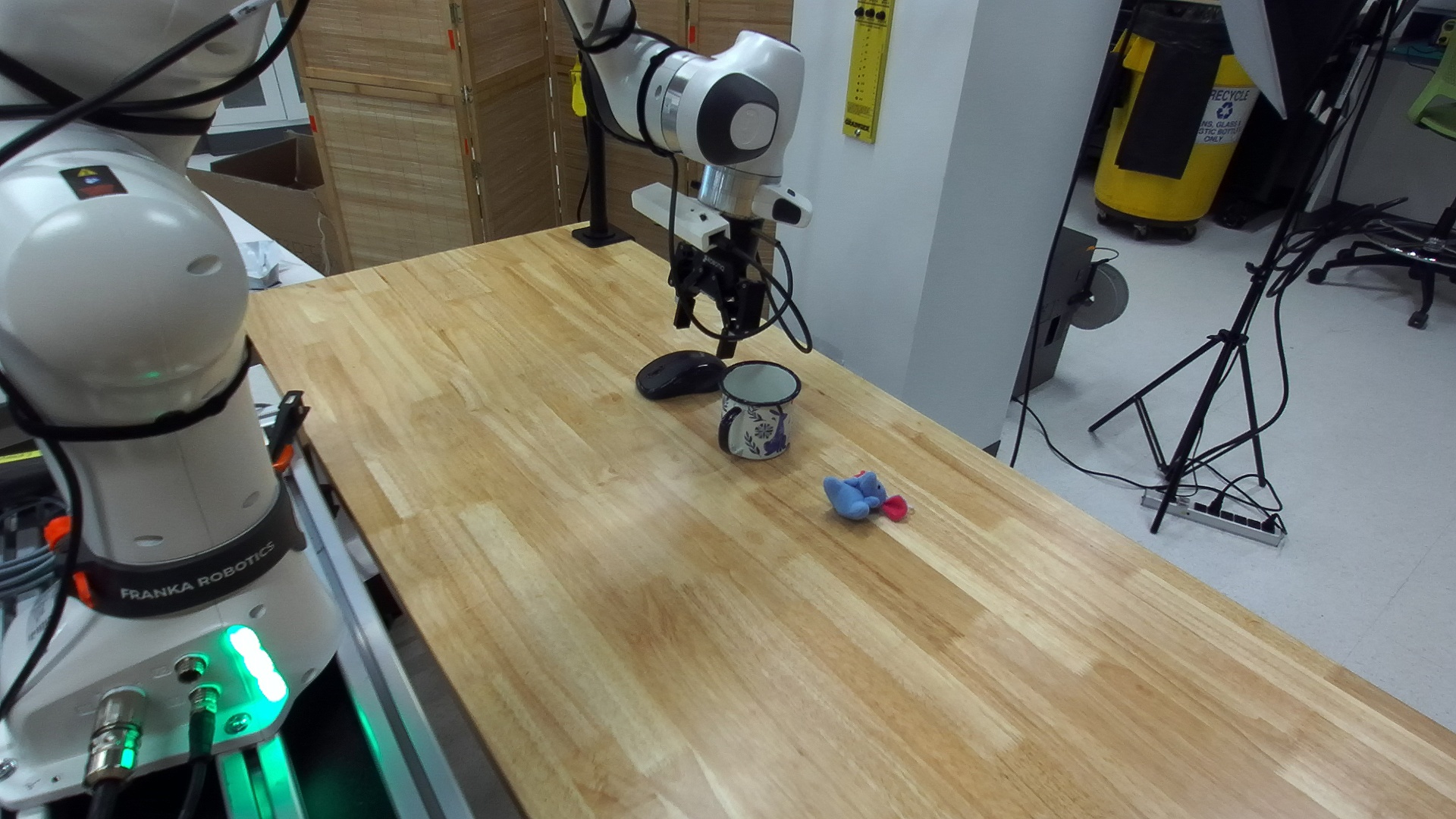} &
            \includegraphics[width=\framew]{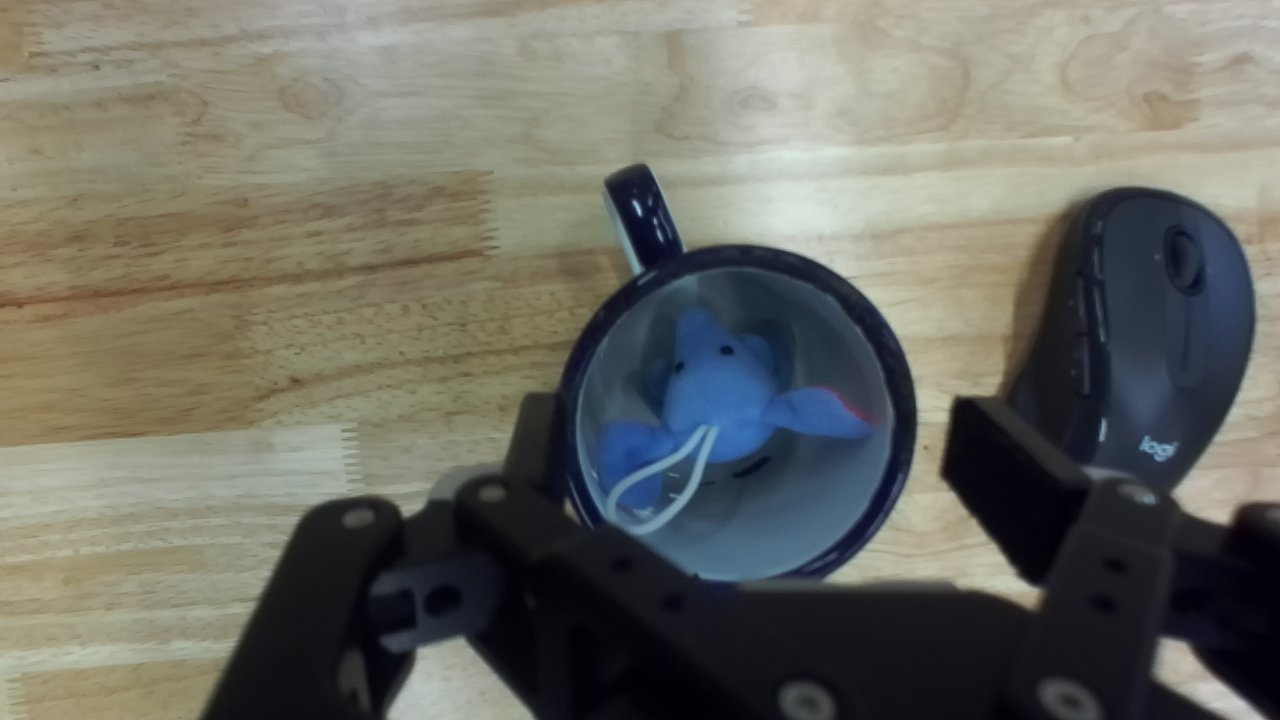} \\[2pt]
\capbox{``The cup's occupied --- there's a mouse on it''} &
\capbox{``I'll clear it first.''} &
\capbox{``Mouse set aside. The cup's empty now.''} &
\capbox{``Now I can put the doll in the cup.''} \\
        \end{tabular}
\caption*{\small \textit{Task: ``Put the doll in the cup.'' (cup is occupied with a mouse)}}
    \end{subfigure}

    \caption{Key frames from a single execution of an obstacle task where
    the goal container is already occupied; per-frame reasoning appears
    below the images. The agent recognizes the cup is taken, clears the
    occupying object, and only then places the doll---a detour a policy
    that maps the instruction straight to a place action cannot make.}
    \label{fig:rollout_occupied_cup}
\end{figure}
\begin{figure}[H]
    \centering
    \setlength{\tabcolsep}{1.5pt}
    \renewcommand{\arraystretch}{0.9}
    \newcommand{\framew}{0.235\textwidth}
    \newcommand{\capbox}[1]{\parbox[t]{\framew}{\centering\scriptsize\textit{#1}}}
    \begin{subfigure}{\textwidth}
        \centering
        \begin{tabular}{@{}cccc@{}}
            \includegraphics[width=\framew]{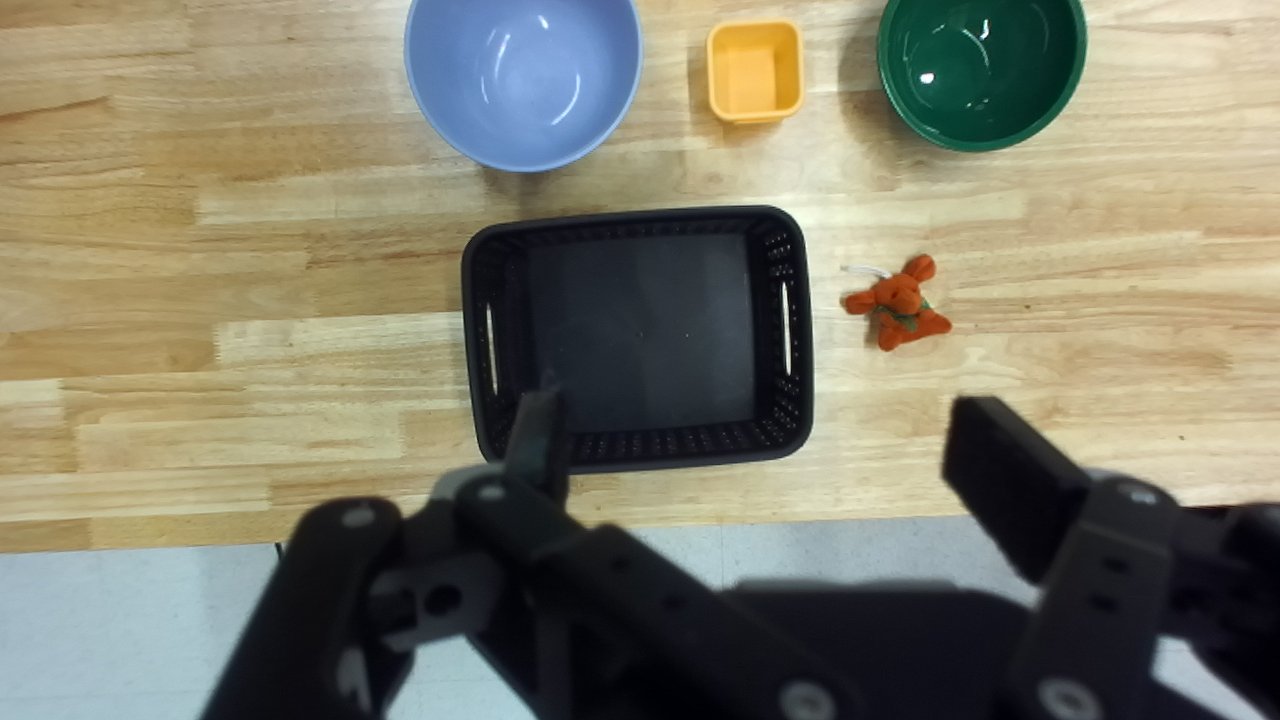} &
            \includegraphics[width=\framew]{figs/app-figs/SR/smallest/turn01_Perceive_wrist.jpg} & 
            \includegraphics[width=\framew]{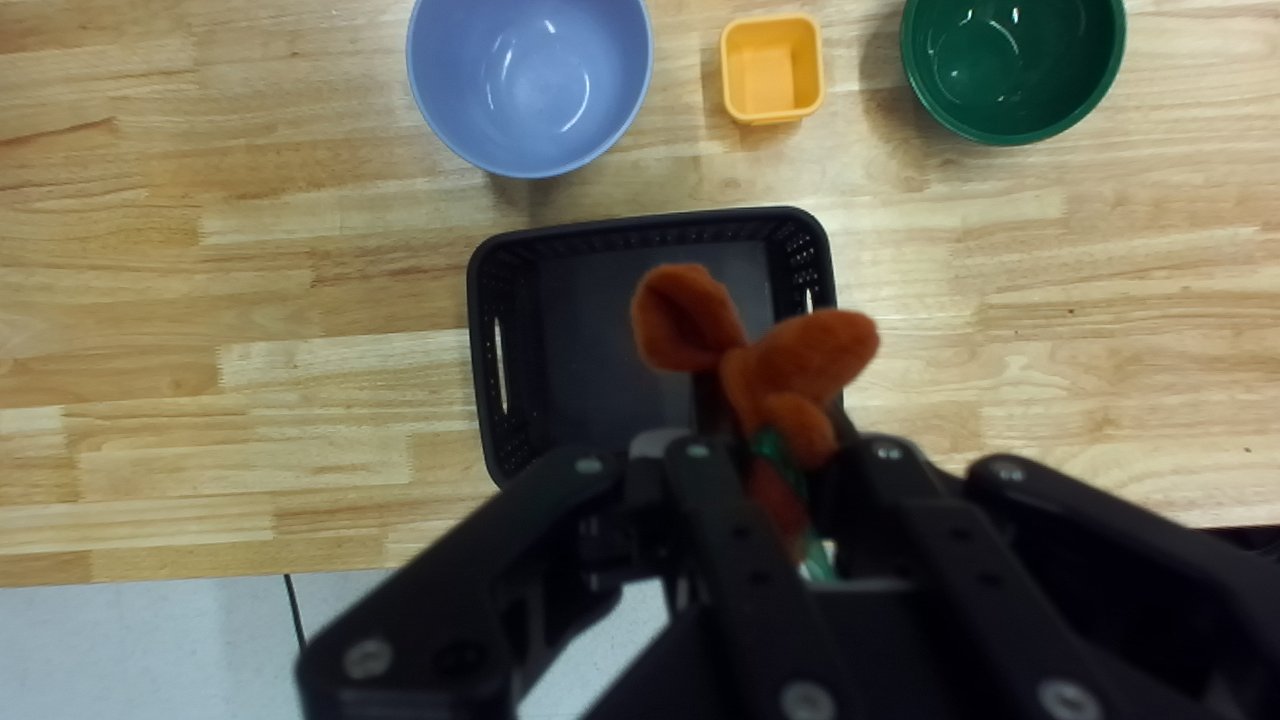} &
            \includegraphics[width=\framew]{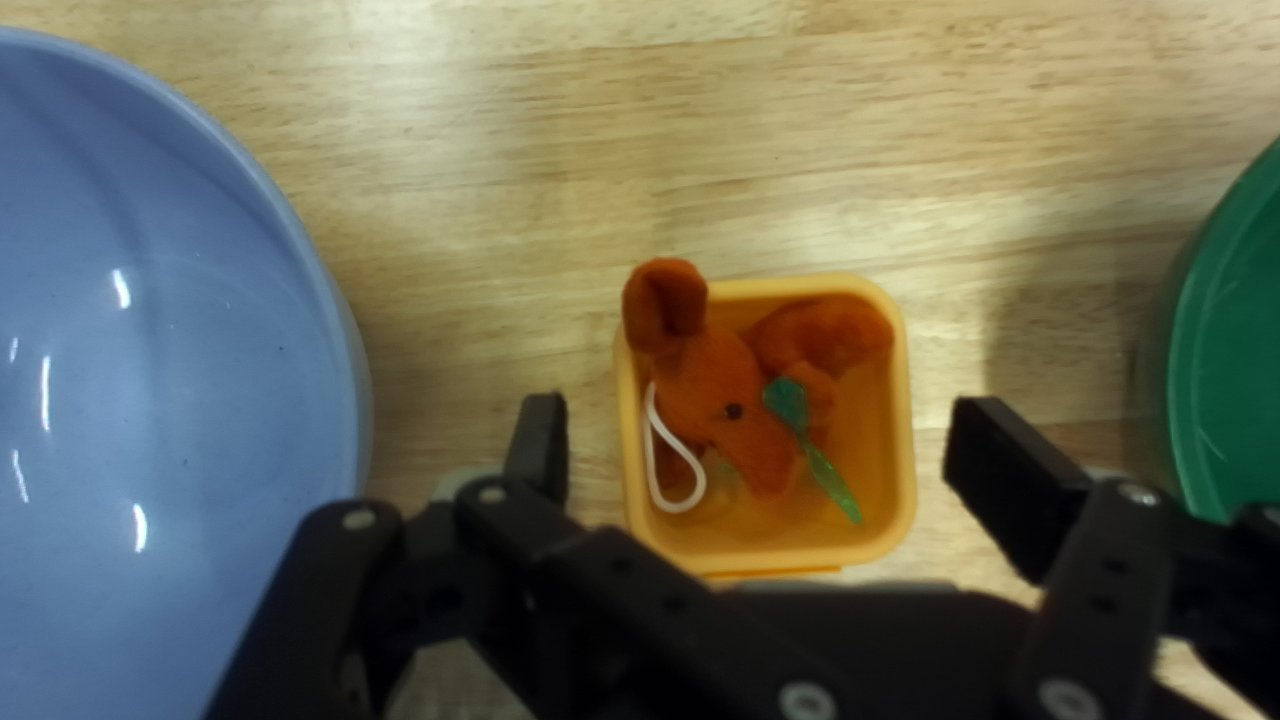} \\[2pt]
\capbox{``Four containers --- basket, large bowl, medium bowl, and small one.''} &
\capbox{``The doll's small; the smallest one should still hold it. Let me try that.''} &
\capbox{``Picking up the doll.''} &
\capbox{``It fits --- the small container was the right choice.''} \\
        \end{tabular}
\caption*{\small \textit{Task: ``Put the doll in the smallest container it still fits in.''}}
    \end{subfigure}
    \caption{Key frames from a single execution of a spatial-reasoning task
    where the doll must go in the smallest container it still fits in;
    per-frame reasoning appears below the images. The agent compares the
    available containers by size, reasons that the smallest one should still
    hold the doll, and confirms the fit---a relational size judgment beyond a
    policy that grounds only on object names.}
    \label{fig:rollout_smallest}
\end{figure}

\begin{figure}[H]
    \centering
    \setlength{\tabcolsep}{1.5pt}
    \renewcommand{\arraystretch}{0.9}
    \newcommand{\framew}{0.235\textwidth}
    \newcommand{\capbox}[1]{\parbox[t]{\framew}{\centering\scriptsize\textit{#1}}}
    \begin{subfigure}{\textwidth}
        \centering
        \begin{tabular}{@{}cccc@{}}
            \includegraphics[width=\framew]{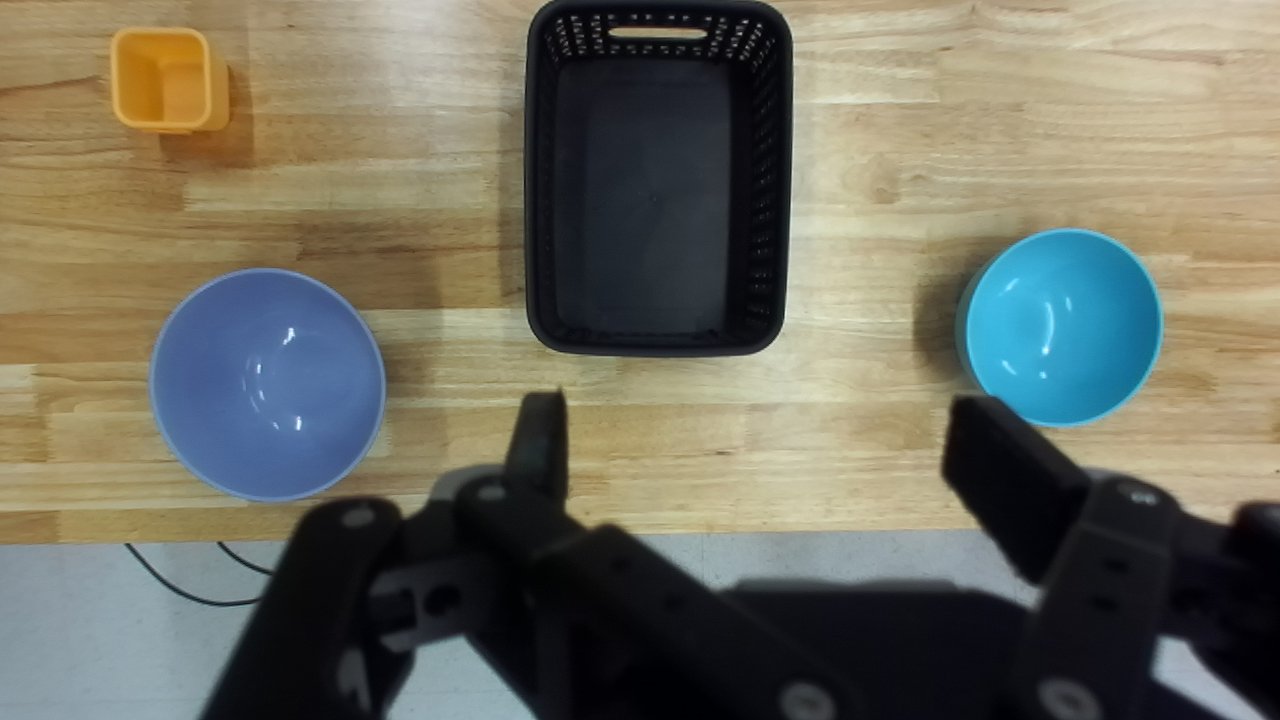} &
            \includegraphics[width=\framew]{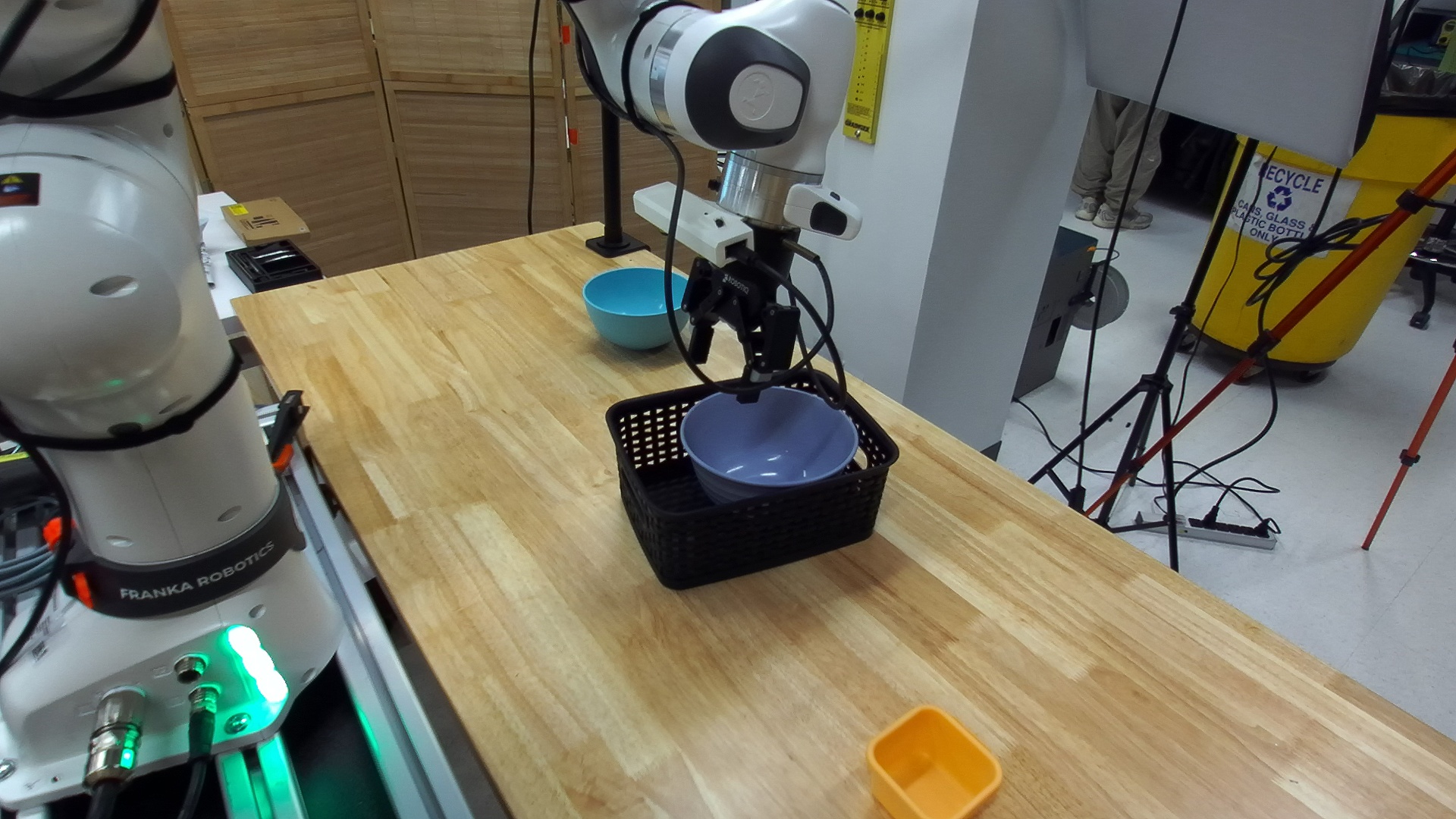} &
            \includegraphics[width=\framew]{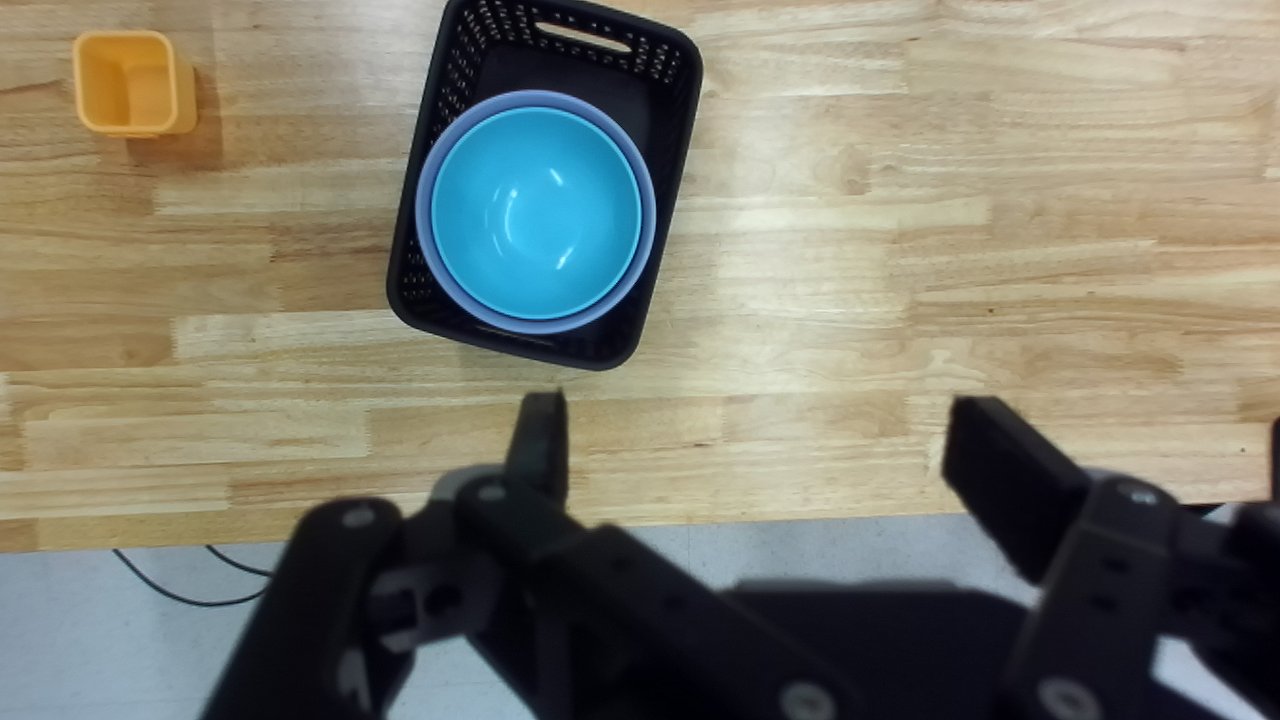} & 
            \includegraphics[width=\framew]{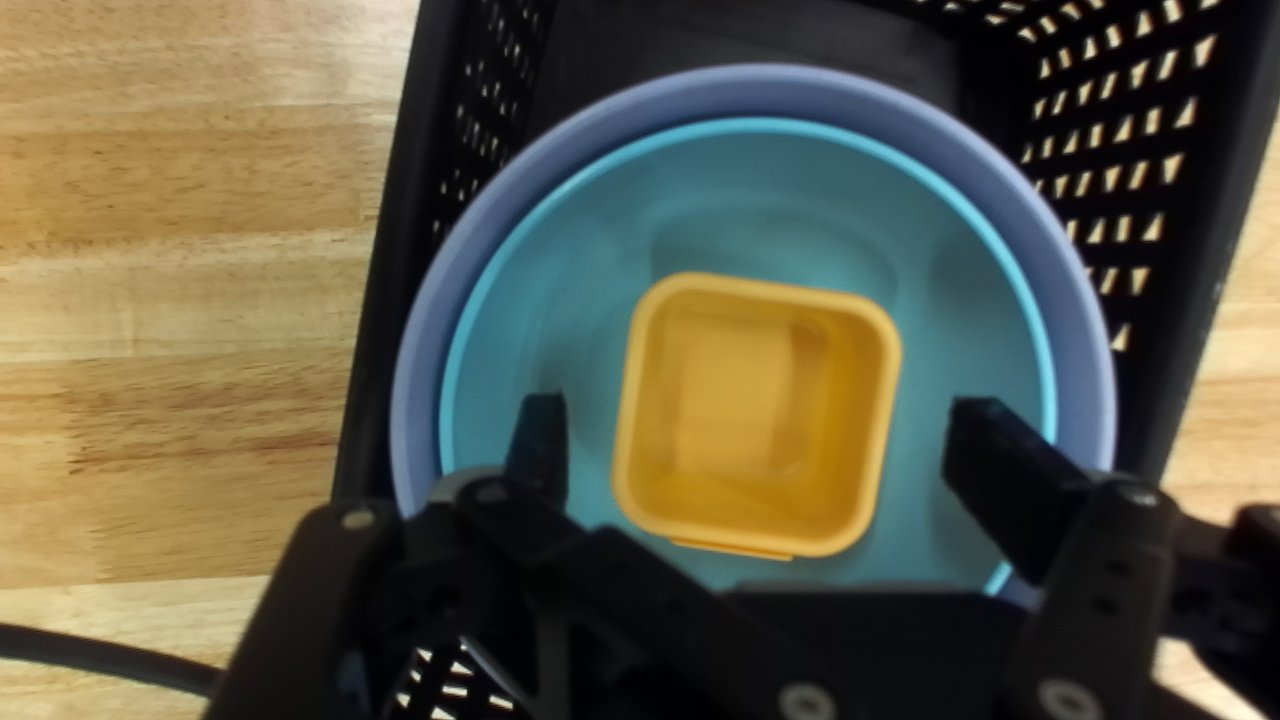} \\[2pt]
\capbox{``Four containers --- I'll nest them into the basket, largest first.''} &
\capbox{``Large blue bowl goes in.''} &
\capbox{``Then the medium bowl.''} &
\capbox{``Small one last --- every container's in the stack.''} \\
        \end{tabular}
\caption*{\small \textit{Task: ``Stack all the containers. Every container must be in the stack.''}}
    \end{subfigure}
    \caption{Key frames from a single execution of a multi-step composition
    task that requires stacking every container; per-frame reasoning appears
    below the images. The agent orders the containers by size and nests them
    into the largest (the basket) one at a time, decomposing the instruction
    into an ordered sequence of placements an open-loop policy does not plan.}
    \label{fig:rollout_stack}
\end{figure}

\begin{figure}[H]
    \centering
    \setlength{\tabcolsep}{1.5pt}
    \renewcommand{\arraystretch}{0.9}
    \newcommand{\framew}{0.235\textwidth}
    \newcommand{\capbox}[1]{\parbox[t]{\framew}{\centering\scriptsize\textit{#1}}}
    \begin{subfigure}{\textwidth}
        \centering
        \begin{tabular}{@{}cccc@{}}
            \includegraphics[width=\framew]{figs/app-figs/ER/empty-plate/turn02_Pick_before.jpg} &
            \includegraphics[width=\framew]{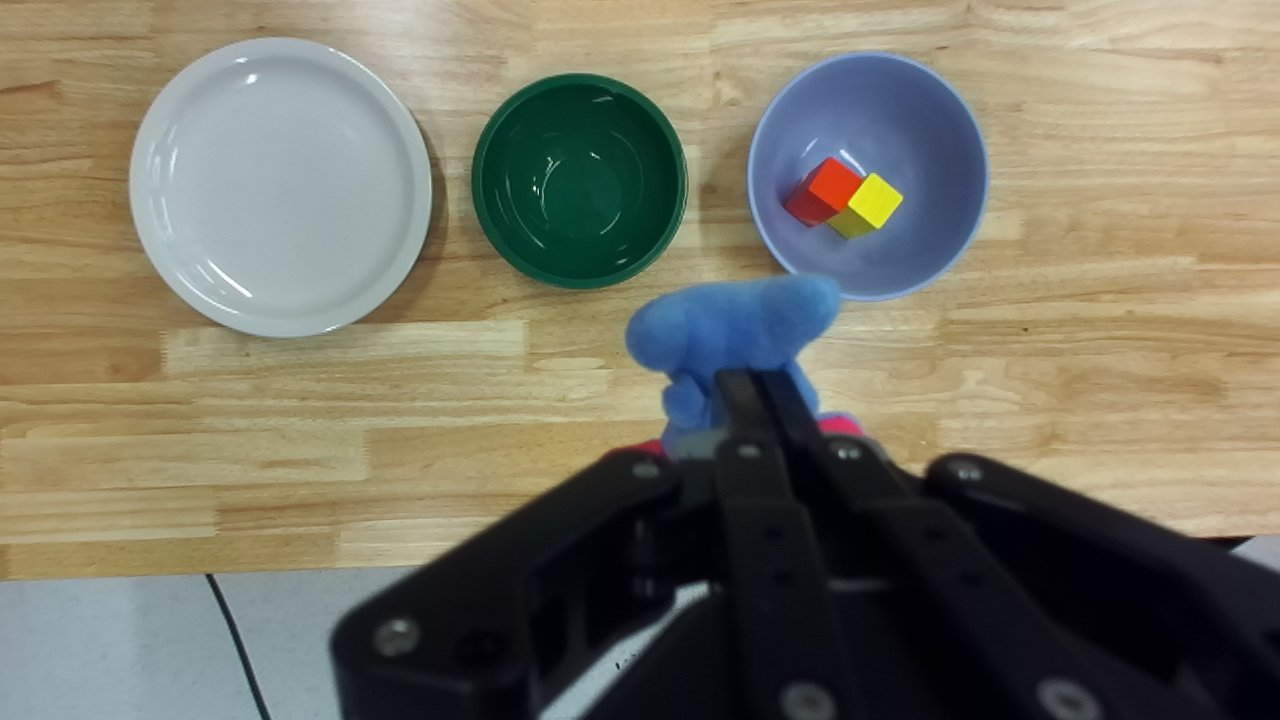} &
            \includegraphics[width=\framew]{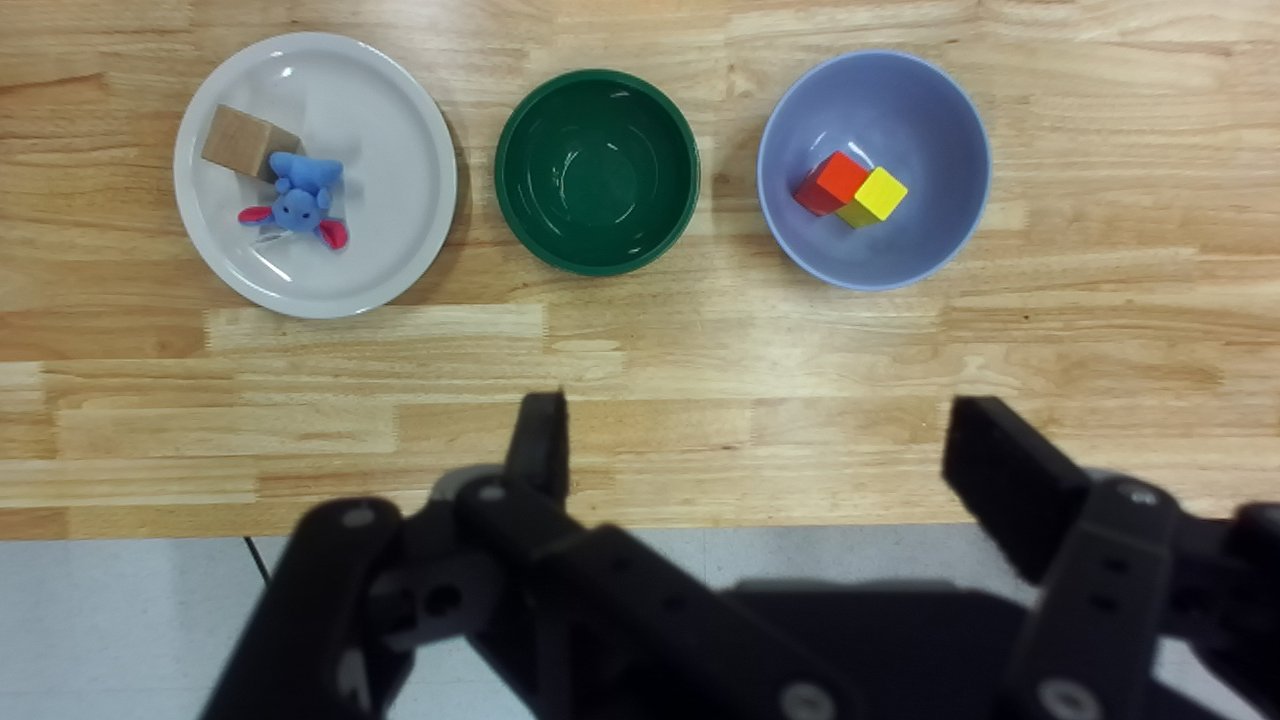} &
            \includegraphics[width=\framew]{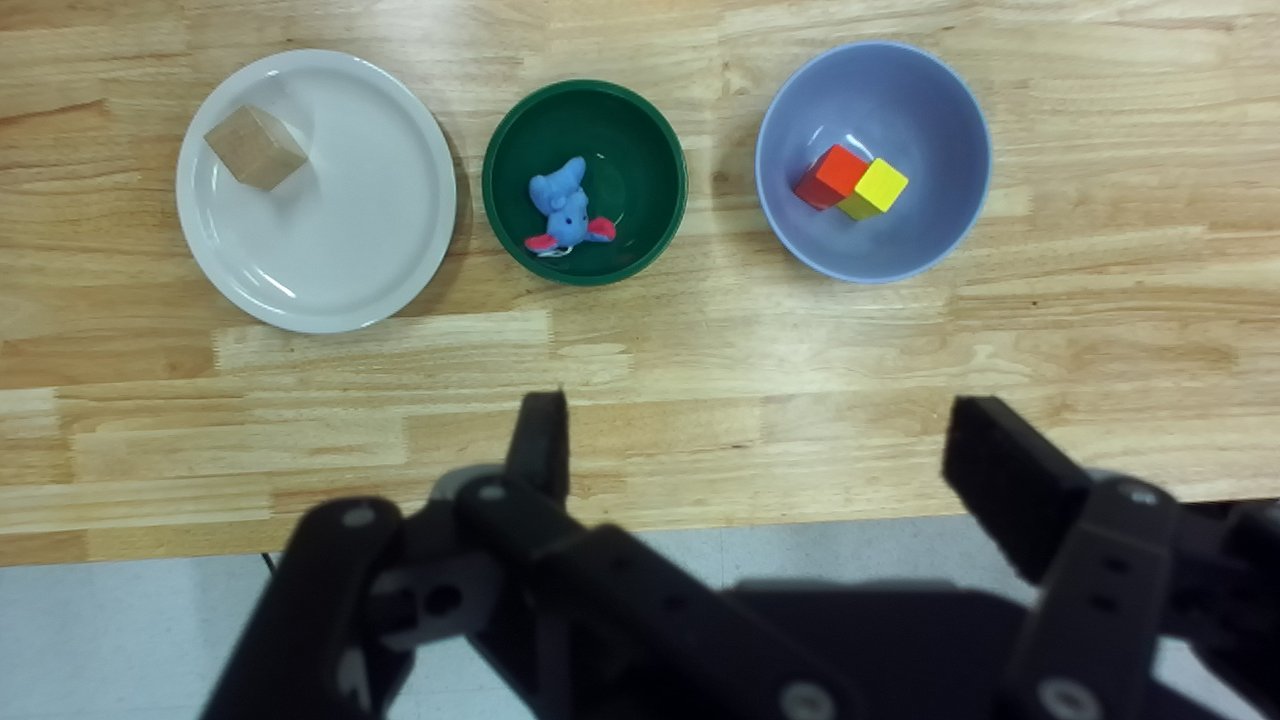} \\[2pt]
\capbox{``The plate's empty --- that's where the doll goes.''} &
\capbox{``Doll in hand, heading for the plate.''} &
\capbox{``Wait --- the plate's not empty anymore.''} &
\capbox{``No empty plate, so the doll goes in the empty bowl.''} \\
        \end{tabular}
\caption*{\small \textit{Task: ``Put the doll on an empty plate. If there is no empty plate, put it in an empty bowl.'' (an item is dropped onto the plate mid-placement)}}
    \end{subfigure}
    \caption{Key frames from a single execution of a conditional
    error-recovery task: the doll should go on an empty plate, with an empty
    bowl as the fallback. The plate starts empty, but an item is placed on it
    mid-execution; the agent detects that the plate is no longer empty,
    re-evaluates the condition, and falls back to the empty bowl---a re-plan
    an open-loop policy cannot make.}
    \label{fig:rollout_empty_plate}
\end{figure}

\begin{figure}[H]
    \centering
    \setlength{\tabcolsep}{1.5pt}
    \renewcommand{\arraystretch}{0.9}
    \newcommand{\framew}{0.235\textwidth}
    \newcommand{\capbox}[1]{\parbox[t]{\framew}{\centering\scriptsize\textit{#1}}}
    \begin{subfigure}{\textwidth}
        \centering
        \begin{tabular}{@{}cccc@{}}
            \includegraphics[width=\framew]{figs/app-figs/LHM/blind/turn01_Perceive_wrist.jpg} &
            \includegraphics[width=\framew]{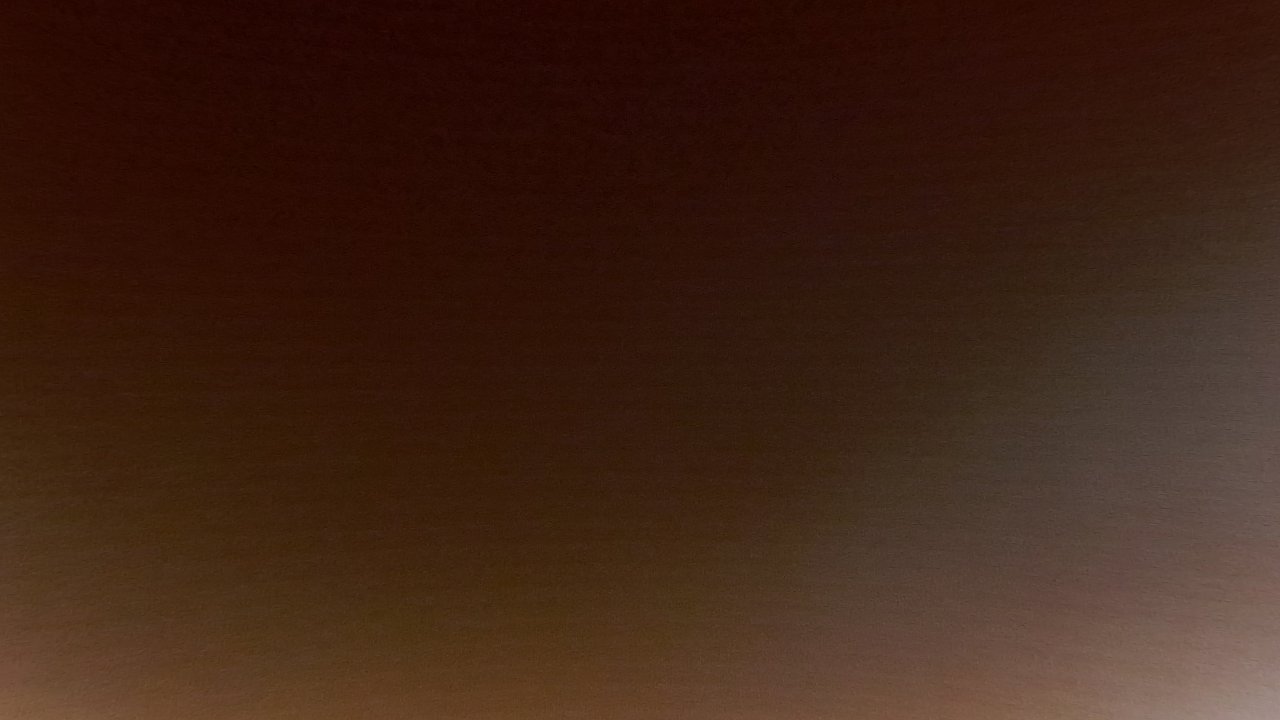} &
            \includegraphics[width=\framew]{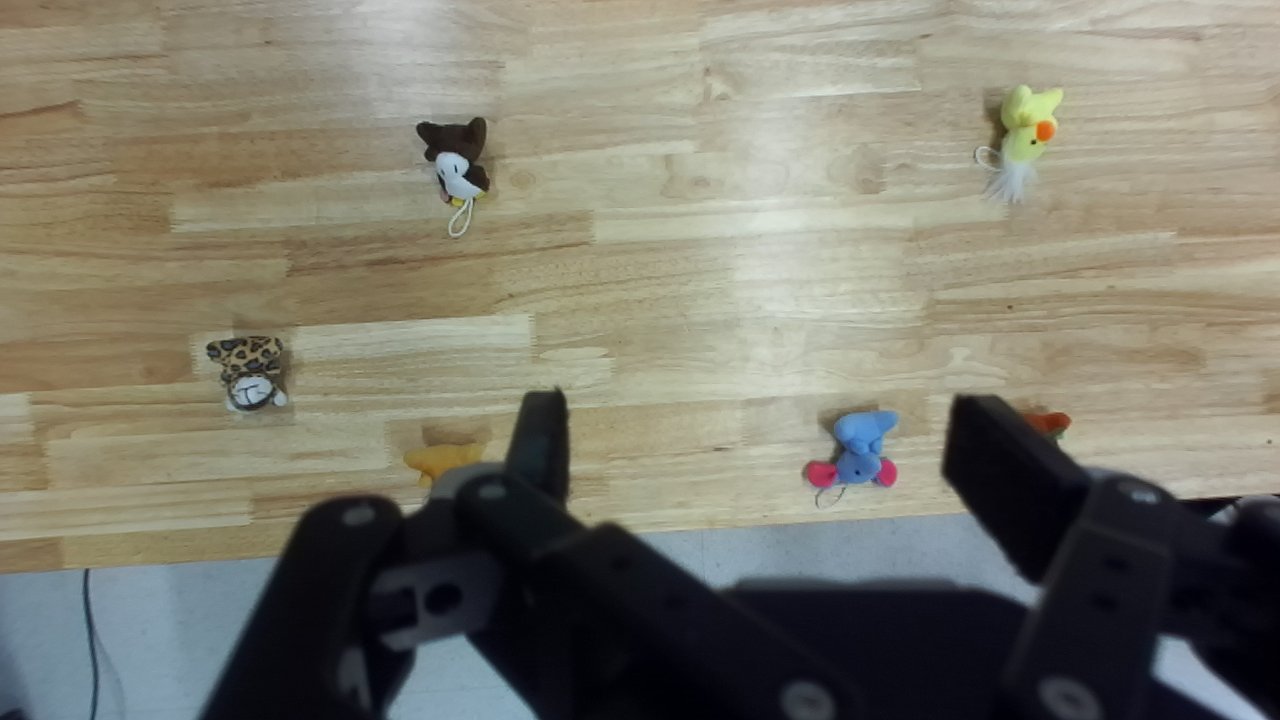} &
            \includegraphics[width=\framew]{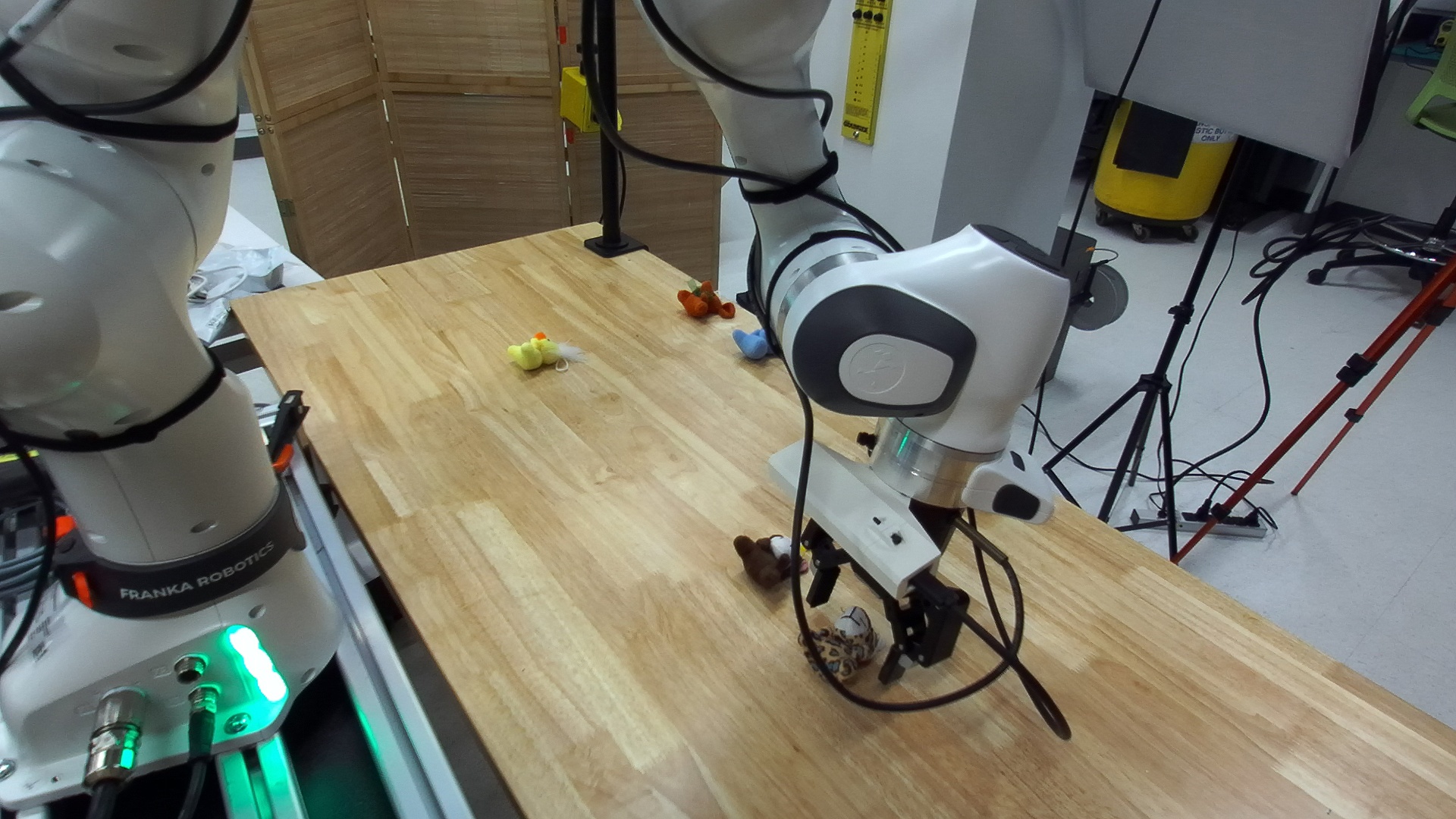} \\[2pt]
\capbox{``Initial scene --- memorizing each doll's spot and its looks.''} &
\capbox{``View blocked --- I'm blind.''} &
\capbox{``I can see again --- the dolls were shuffled. Time to restore.''} &
\capbox{``The dotted doll's spot is free --- putting it back first.''} \\[8pt]
            \includegraphics[width=\framew]{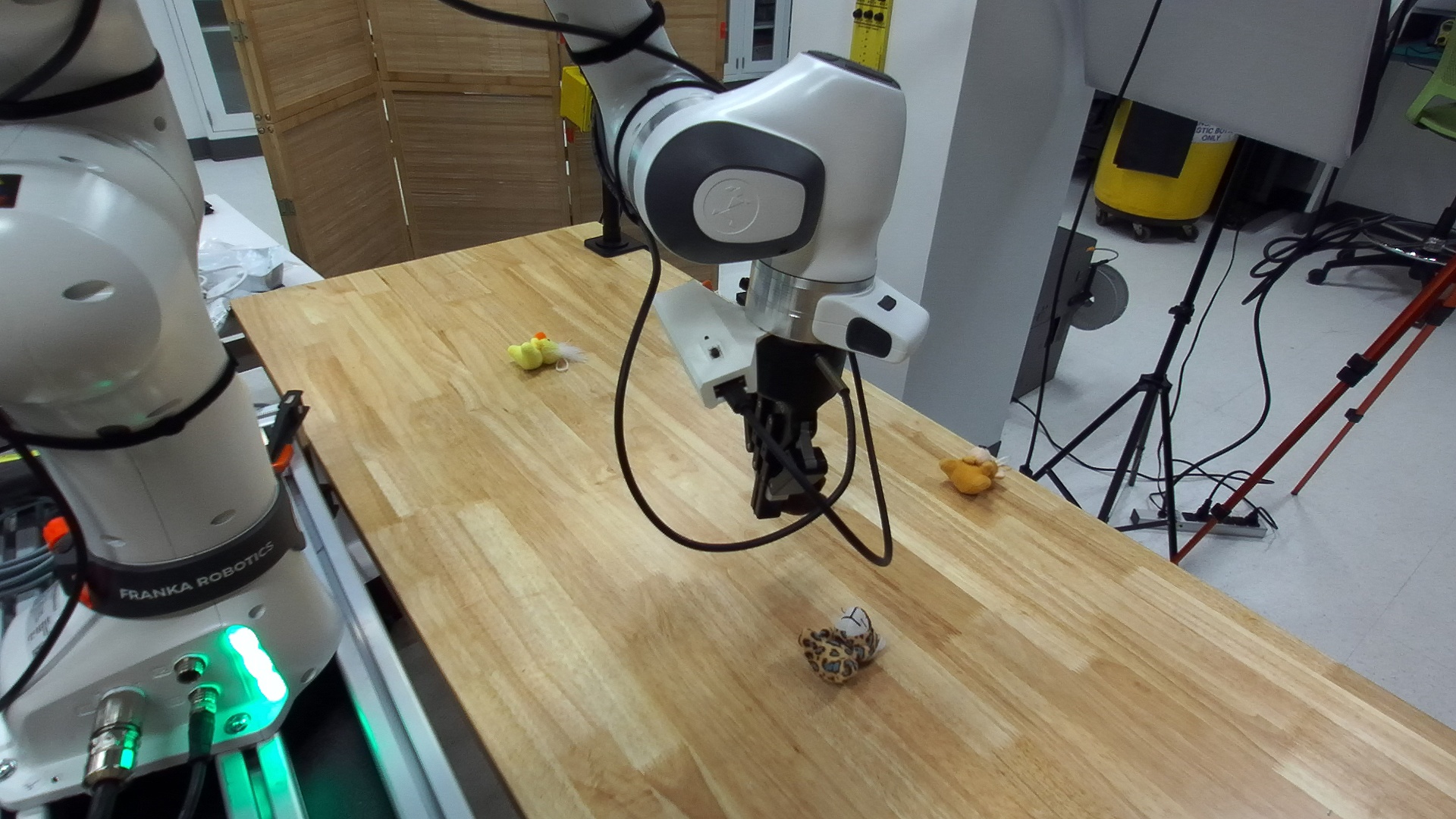} &
            \includegraphics[width=\framew]{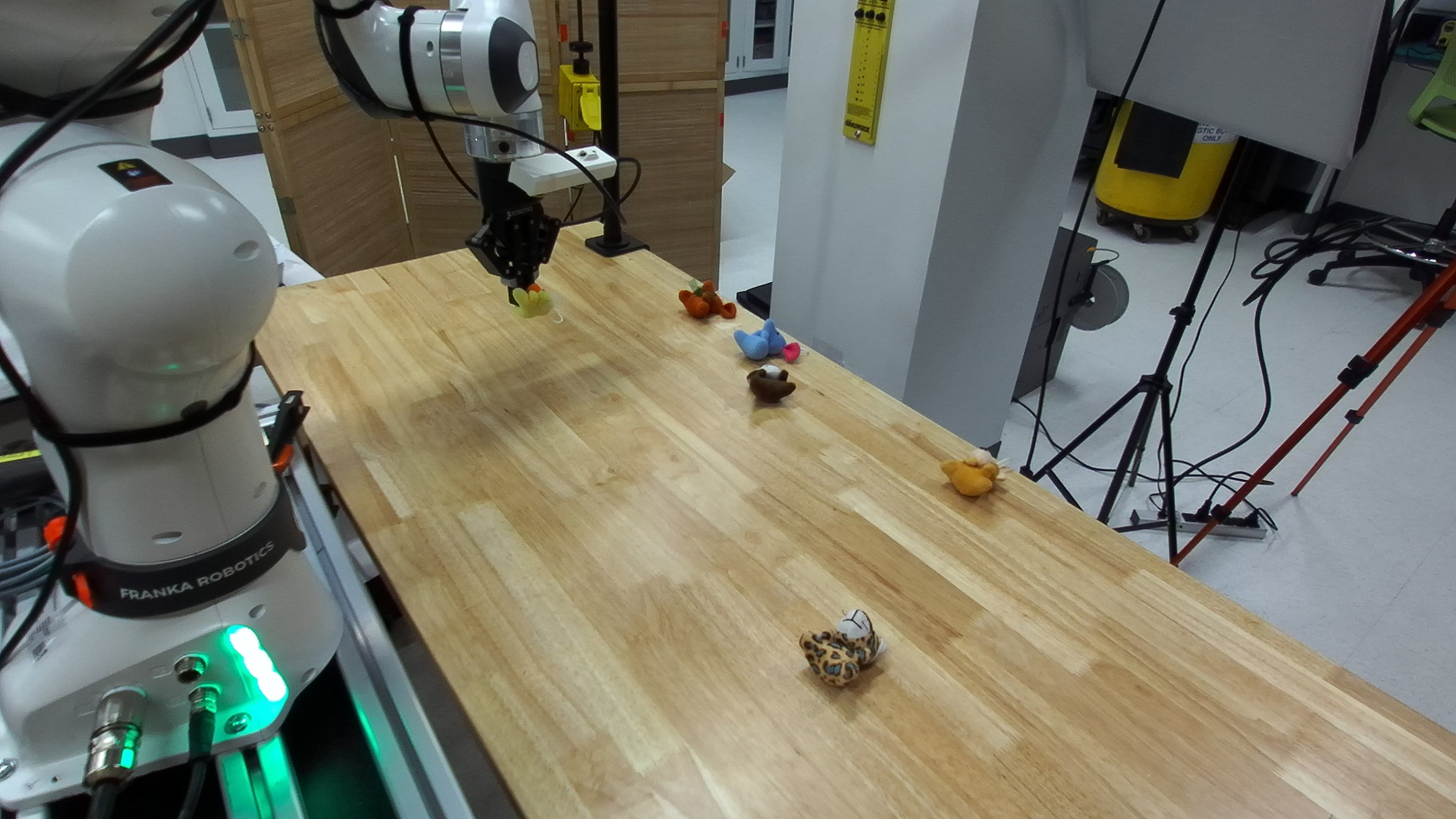} &
            \includegraphics[width=\framew]{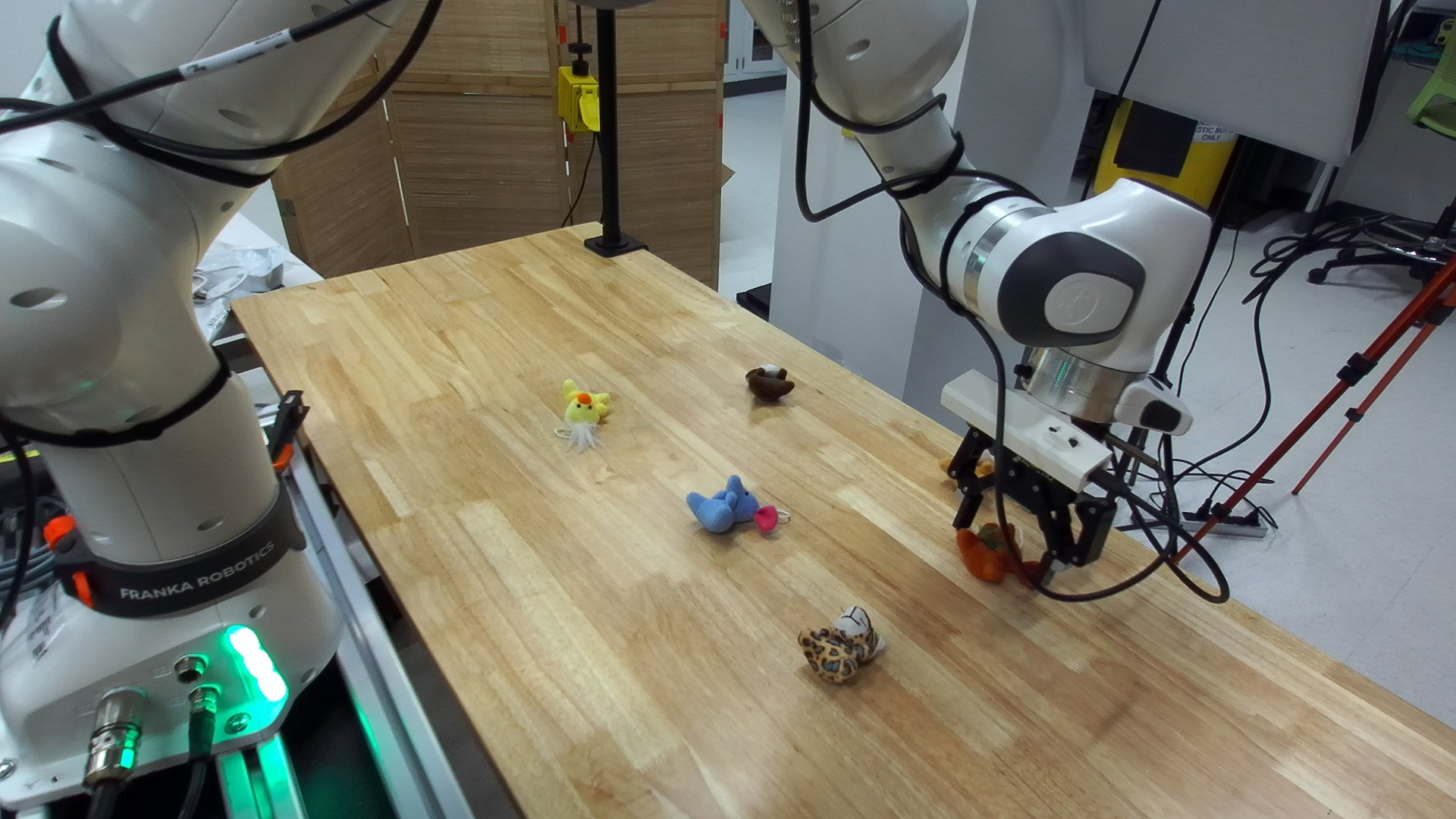} &
            \includegraphics[width=\framew]{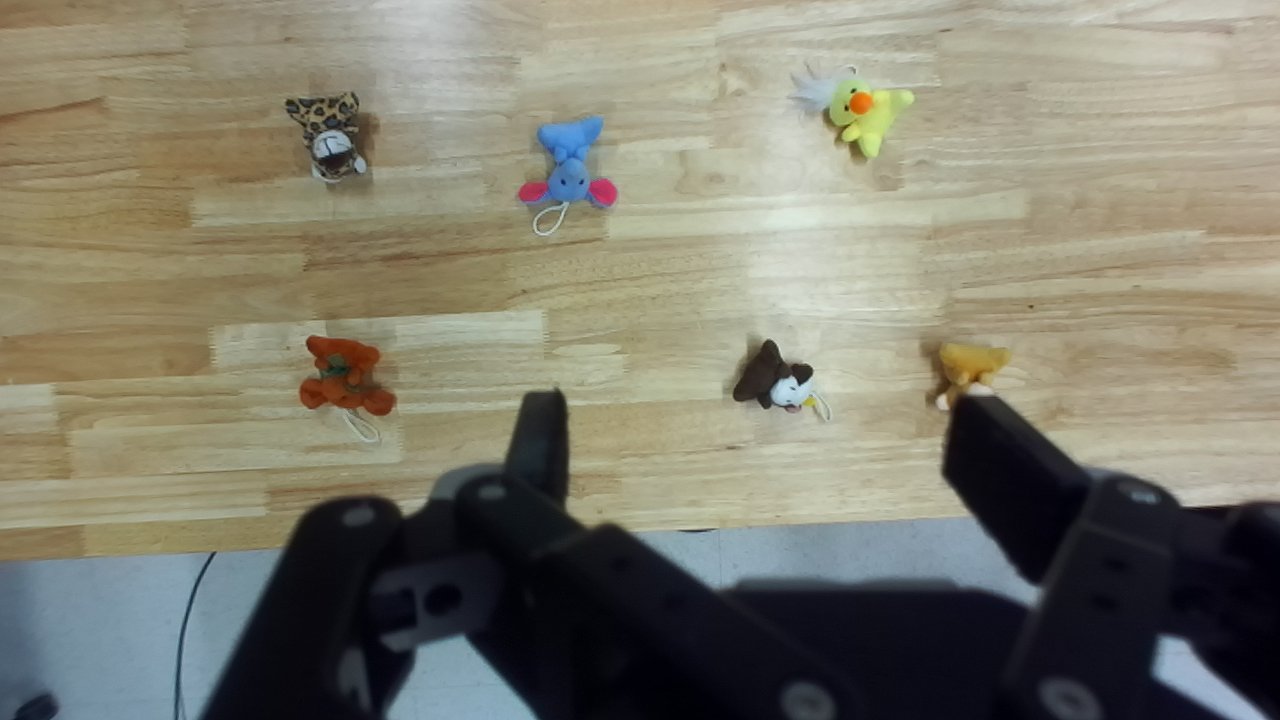} \\[2pt]
\capbox{``The yellow doll's spot is blocked by the brown one --- return the brown doll first.''} &
\capbox{``Brown doll's home; now the yellow doll's spot is clear --- placing it.''} &
\capbox{``Finishing the last placements.''} &
\capbox{``Back to the original arrangement.''} \\
        \end{tabular}
\caption*{\small \textit{Task: ``You'll soon be blind while I shuffle the scene. When you see the scene again, restore everything to how it was at the start.''}}
    \end{subfigure}
    \caption{Key frames (left to right, top to bottom) from a single
    long-horizon execution. The agent memorizes each doll's position and
    appearance, is blinded while the scene is shuffled, then restores the
    original layout one doll at a time. When a doll's home is occupied by
    another, it reasons about ordering---returning the blocking brown doll
    before the yellow one---rather than forcing the placement, continuing
    until the arrangement matches the start. State retention and re-planning
    across a long, interrupted horizon is beyond a reactive policy.}
    \label{fig:rollout_blind}
\end{figure}

\begin{figure}[H]
    \centering
    \setlength{\tabcolsep}{1.5pt}
    \renewcommand{\arraystretch}{0.9}
    \newcommand{\framew}{0.19\textwidth}
    \newcommand{\capbox}[1]{\parbox[t]{\framew}{\centering\scriptsize\textit{#1}}}
    \begin{subfigure}{\textwidth}
        \centering
        \begin{tabular}{@{}ccccc@{}}
            \includegraphics[width=\framew]{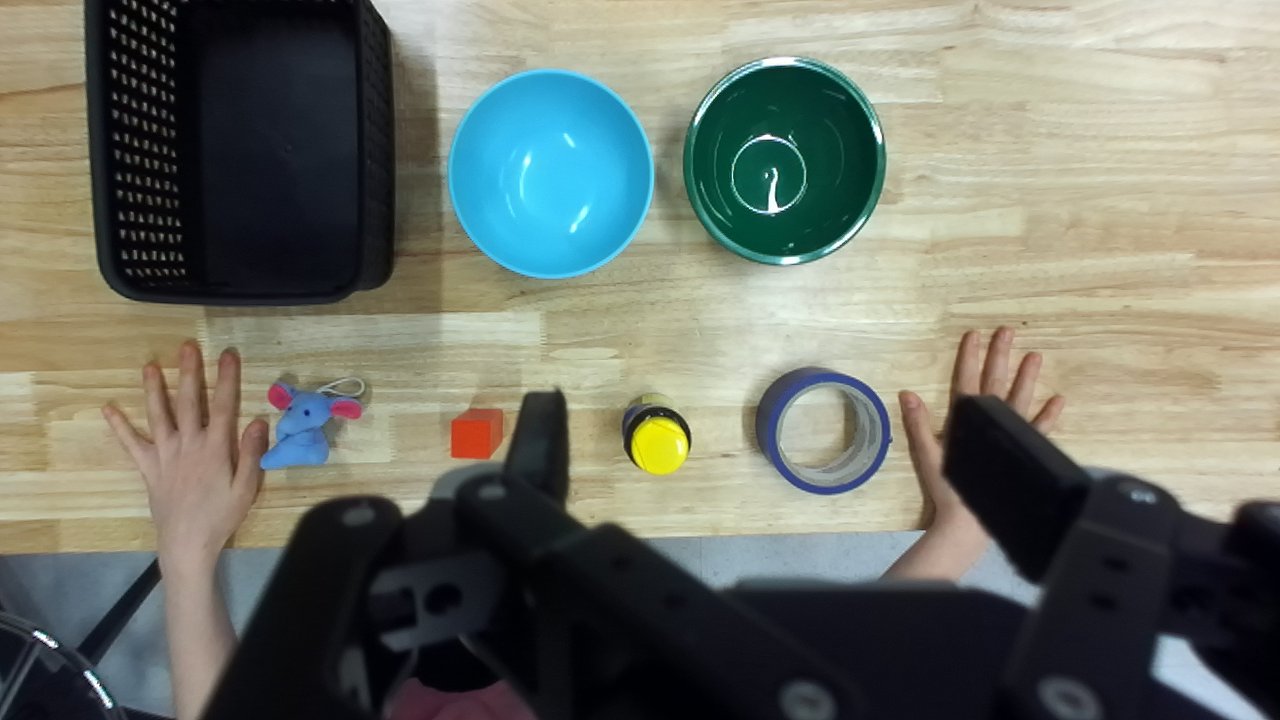} &
            \includegraphics[width=\framew]{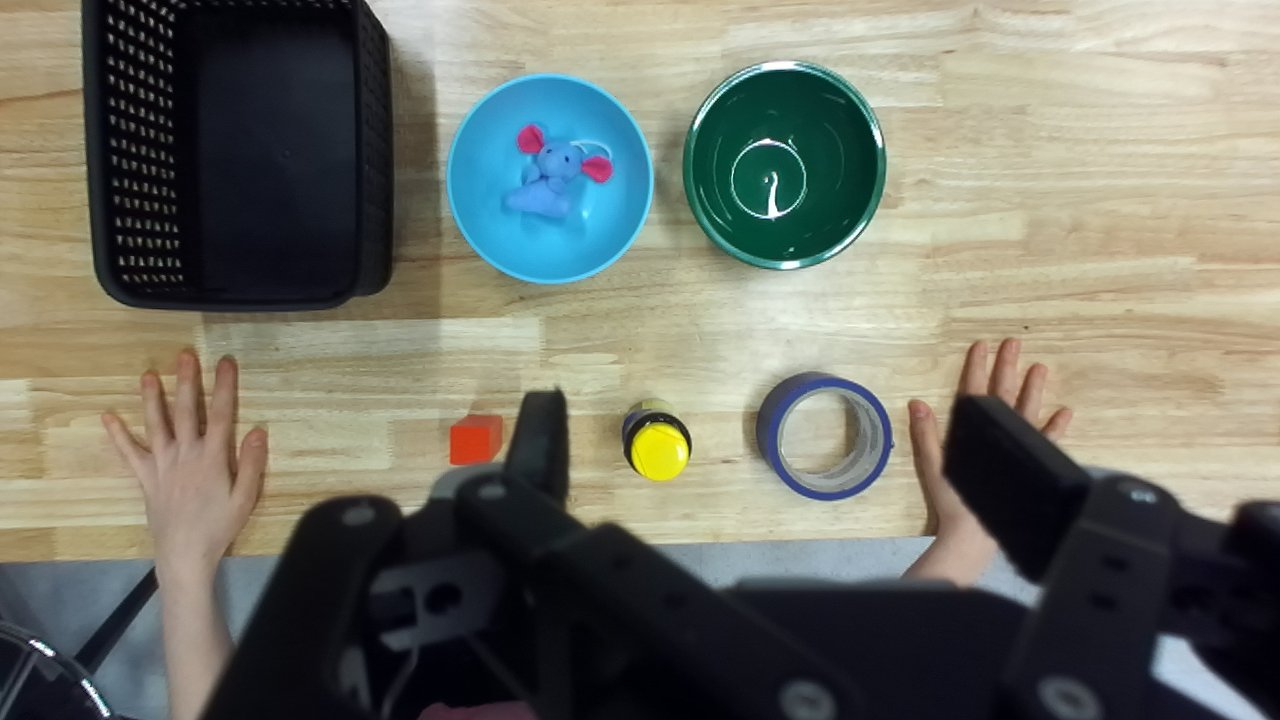} &
            \includegraphics[width=\framew]{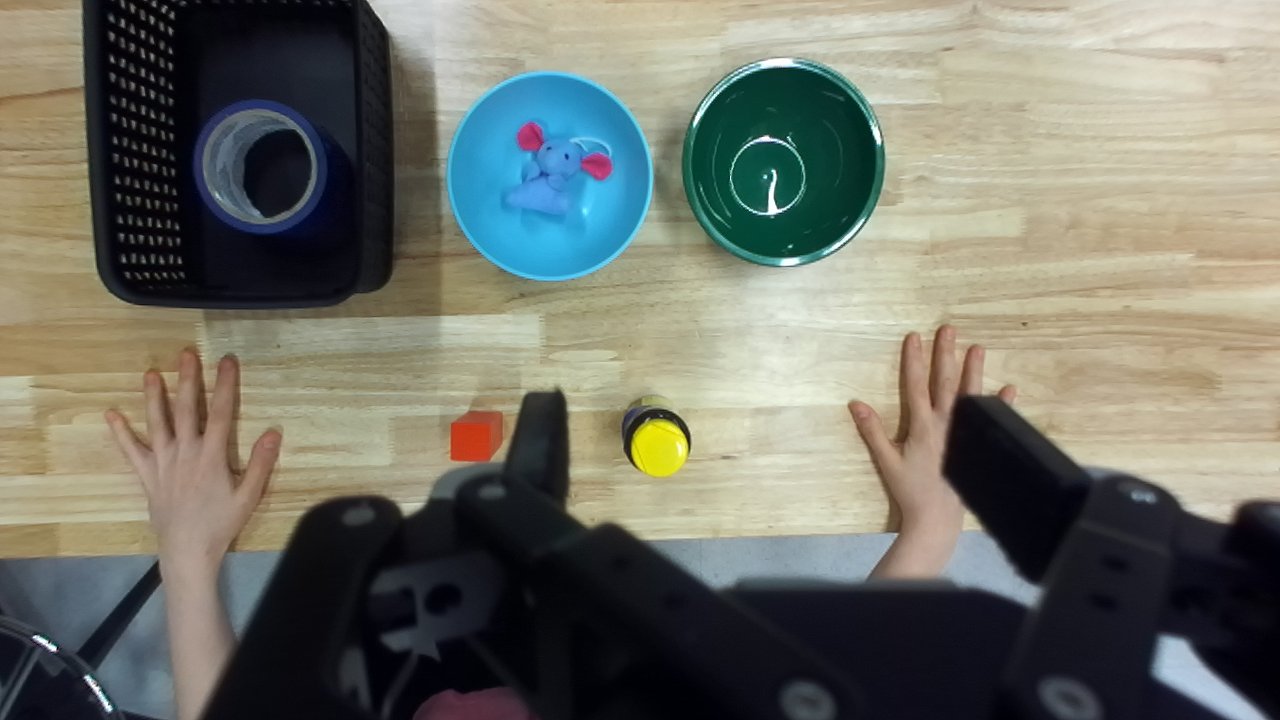} &
            \includegraphics[width=\framew]{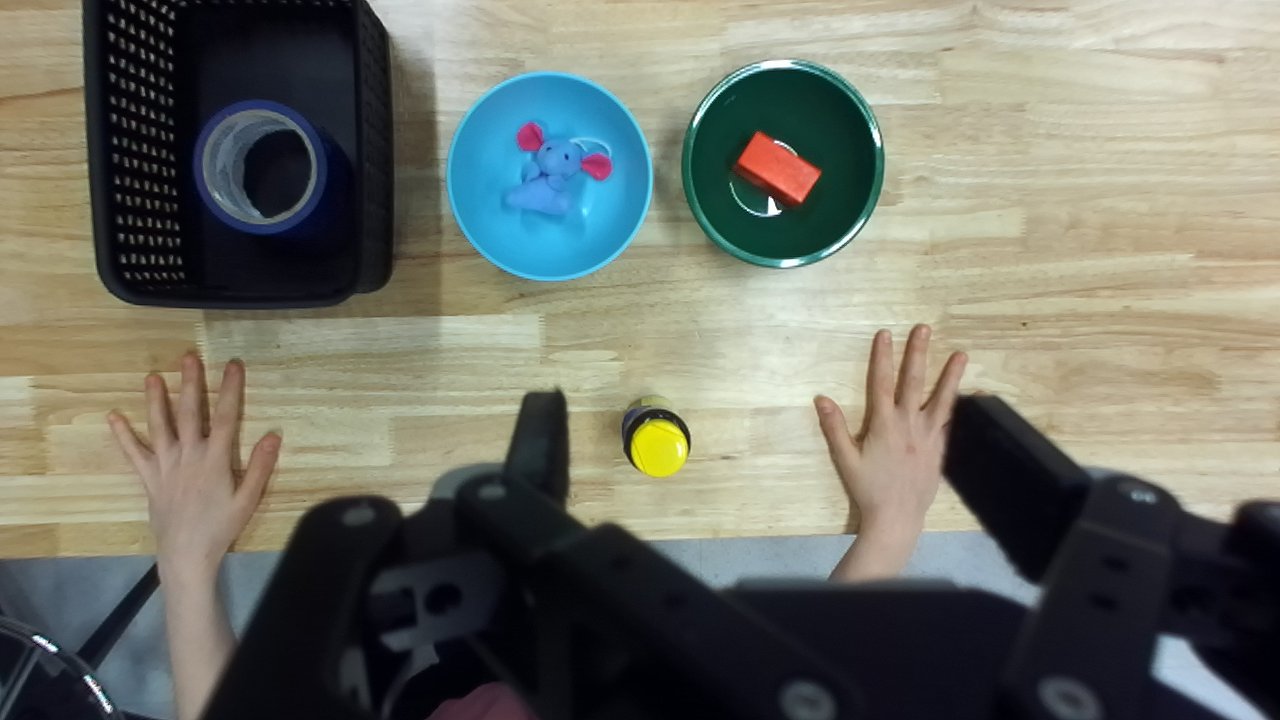} &
            \includegraphics[width=\framew]{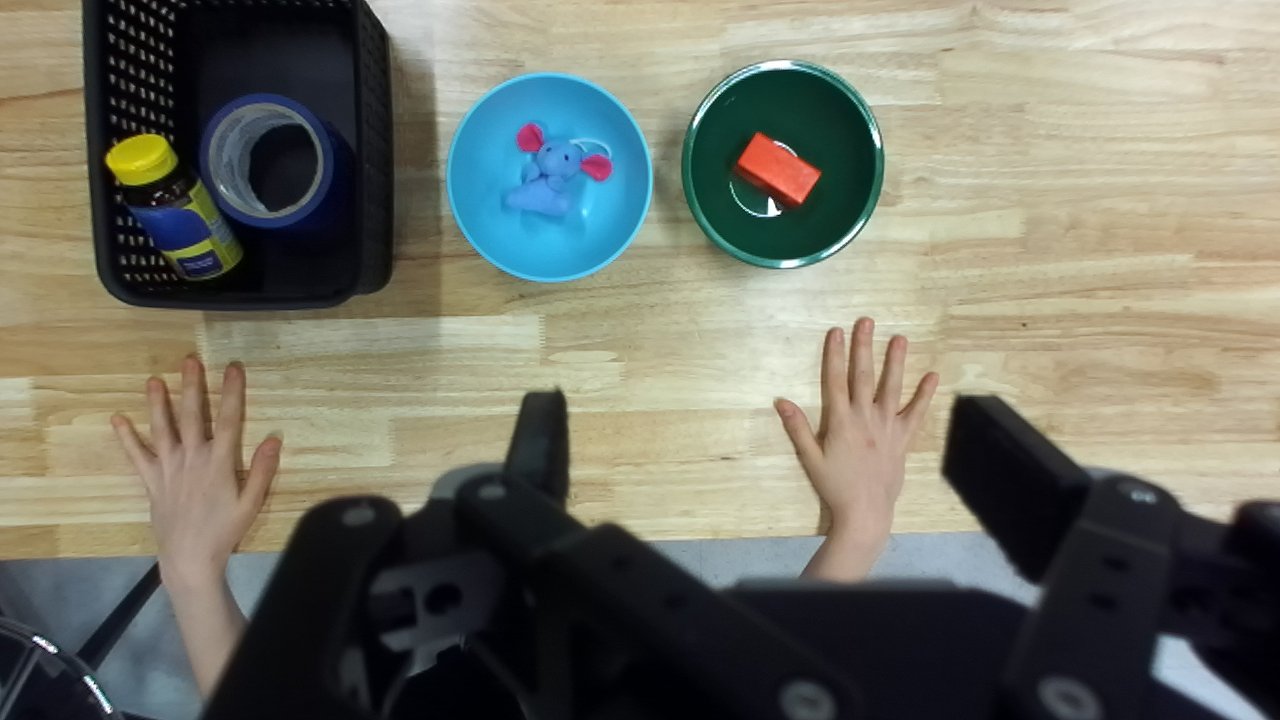} \\[2pt]
\capbox{``A demo's coming --- I'll memorize each move.''} &
\capbox{``Move 1: the doll goes in the blue bowl.''} &
\capbox{``Move 2: the tape goes in the basket.''} &
\capbox{``Move 3: the block goes in the green bowl.''} &
\capbox{``Move 4: the yellow bottle goes in the basket.''} \\[8pt]
            \includegraphics[width=\framew]{figs/app-figs/LHM/demo/turn06_Perceive_wrist.jpg} &
            \includegraphics[width=\framew]{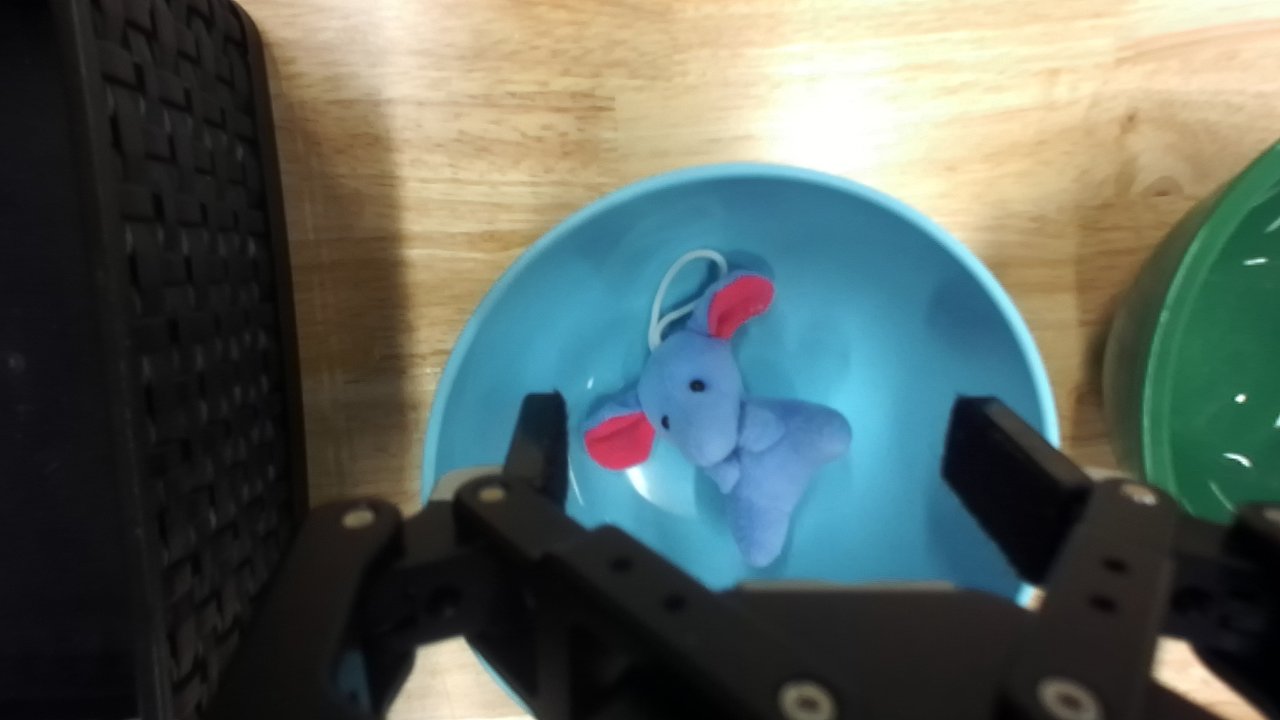} &
            \includegraphics[width=\framew]{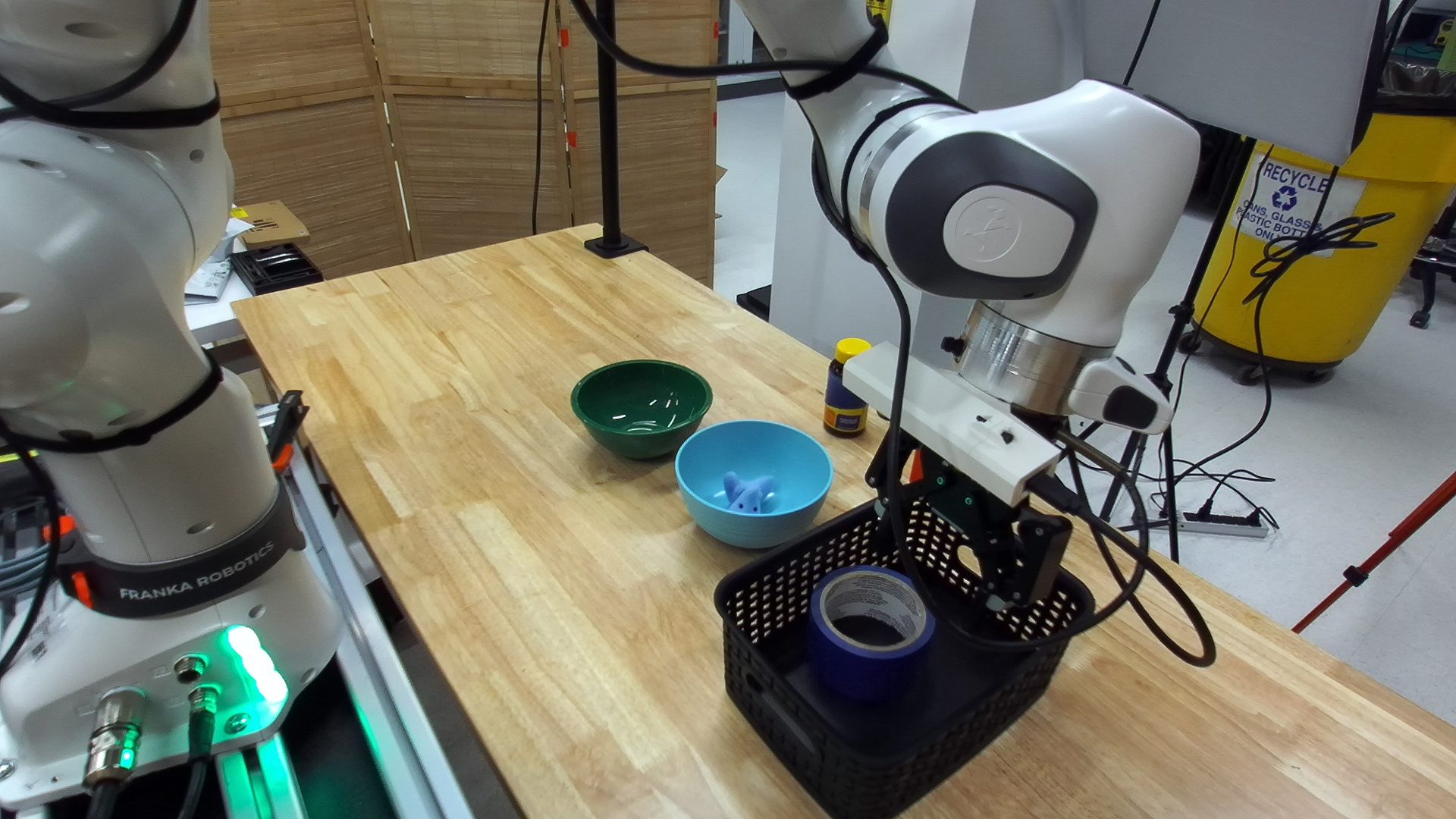} &
            \includegraphics[width=\framew]{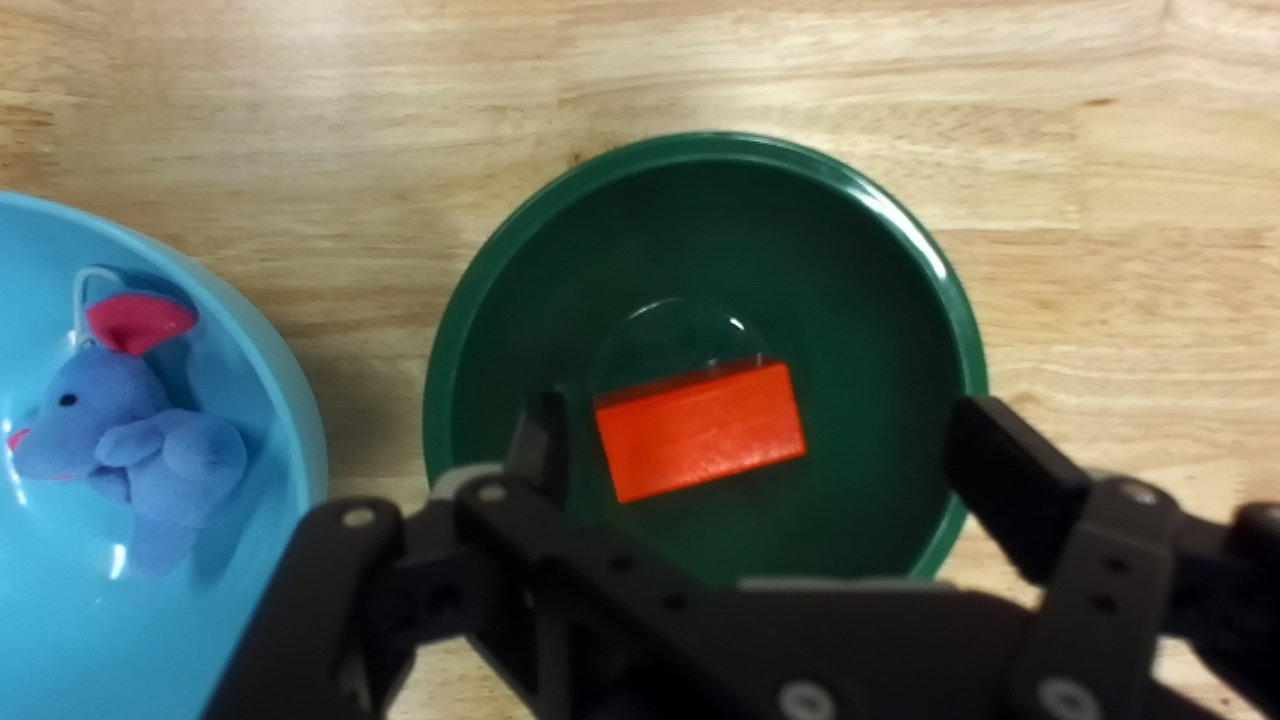} &
            \includegraphics[width=\framew]{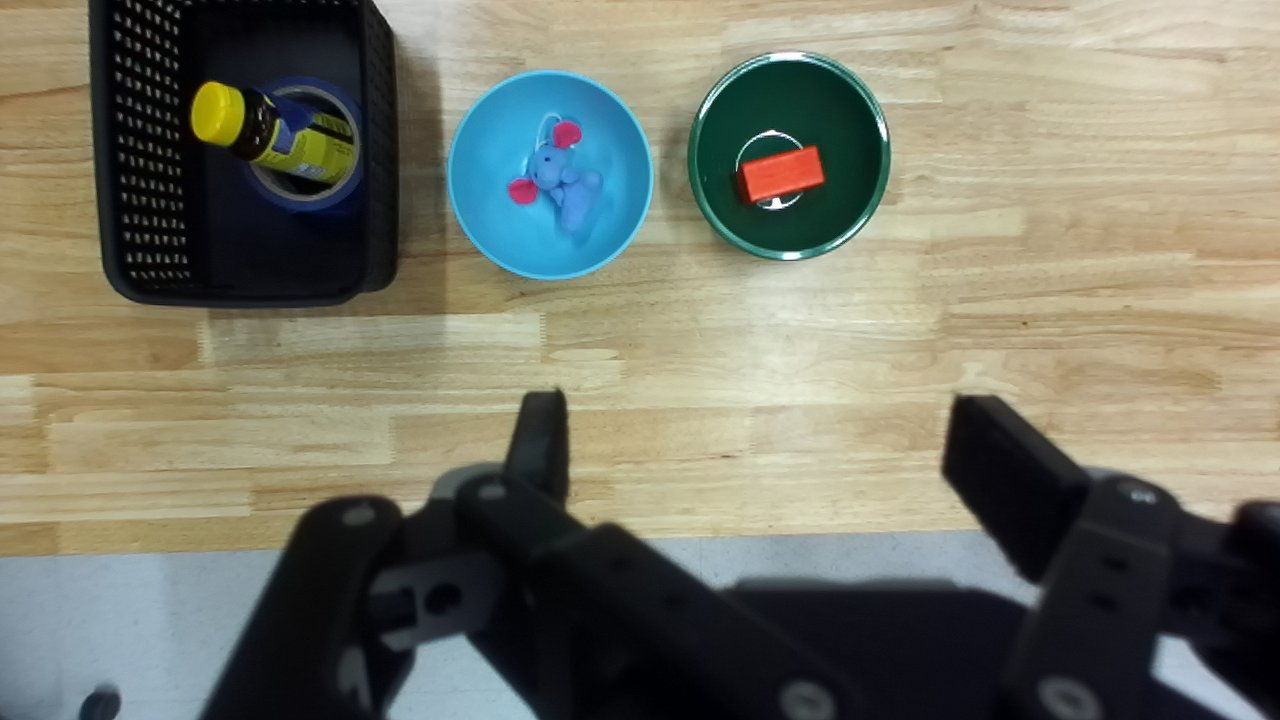} \\[2pt]
\capbox{``Hands gone, scene reset. Now I replay the sequence.''} &
\capbox{``Replaying --- doll into the blue bowl.''} &
\capbox{``Tape into the basket.''} &
\capbox{``Block into the green bowl.''} &
\capbox{``Bottle in the basket too --- the sequence matches the demo.''} \\
        \end{tabular}
\caption*{\small \textit{Task: ``I'll demo the task with my hands, one object move at a time. Watch each move. After I reset the scene and my hands leave the frame, replay exactly what I demonstrated.''}}
    \end{subfigure}
    \caption{Key frames (left to right, top to bottom) from a single
    long-horizon imitation episode. A human demonstrates a multi-step task
    one move at a time---doll to the blue bowl, tape to the basket, block to
    the green bowl, bottle to the basket---and the agent watches and commits
    the ordered sequence to memory. After the demonstrator resets the scene
    and leaves the frame, the agent reproduces the same object-to-destination
    moves in order. Retaining and replaying a demonstrated sequence across a
    reset is beyond a reactive policy.}
    \label{fig:rollout_demo}
\end{figure}


Recorded sessions of the agent on representative tasks are available as videos on our project page: \url{https://lianegalanti.github.io/Pigey/}.

\subsection{First-failure mode distribution}
\label{app:failure_modes}  
For every failed episode we record the \emph{first} error the policy could
not recover from, and bucket it into one of four modes. Transient errors
caught and retried by the closed loop are not counted as failures.
Table~\ref{tab:failure_modes} reports the distribution over all 150 trials
per method.
 
\begin{table}[h]
\centering
\begin{tabular}{l rrr}
\toprule
First-failure mode & $\pi_{0.5}$ & TiPToP & \methodname{} (ours) \\
\midrule
Grounding (wrong / random target) & 86 & 5  & 0 \\
Reasoning / planning              & 38 & 65 & 0 \\
Grasp (execution)                 & 1  & 7  & 2 \\
Verifier false-success            & 0  & 0  & 2 \\
\midrule
Total failures                    & 125 & 77 & 4 \\
\bottomrule
\end{tabular}
\caption{Distribution of first-failure modes (counts, out of 150 trials
each). $\pi_{0.5}$ fails predominantly at \emph{grounding}---selecting the
wrong object; TiPToP grounds correctly but fails at \emph{reasoning /
planning} on the multi-step, obstacle, recovery, and memory tasks; while
\methodname{} fails only four times, split between grasp execution and
\emph{verifier false-success}, a mode unique to its closed-loop verifier
(declaring success when the target was not actually achieved). Transient
errors recovered by the closed loop are not counted.}
\label{tab:failure_modes}
\end{table}

\section{Supplementary Ablations}
\label{app:ablations}

\paragraph{Full reasoner sweep (LIBERO-PRO).}
Table~\ref{tab:liberopro_full} reports \methodname{} with its strongest reasoner; here we give the full sweep across all nine frontier VLMs we tried (Table~\ref{tab:liberopro_full}). Two things hold across the sweep. First, \emph{every} reasoner clears both non-orchestrated baselines---raw $\pi_{0.5}$ (\num{12.8}\%) and CaP-Agent0 (\num{18}\%)---by a wide margin, so the orchestration gain is a property of the closed-loop structure rather than of any single model. Second, mean success rises gradually with reasoner capability (\num{44.3}\% to \num{53.3}\%), and the spread is largest on the suites that demand re-decomposition rather than re-localization (object-task and spatial-task), consistent with the reasoner supplying decomposition and planning rather than motor skill. The reasoner sets the \emph{magnitude} of the gain, not its \emph{sign}.

\begin{table}[h]
\caption{LIBERO-PRO success rate (\%). Top rows are baselines without orchestration; bottom rows use the same $\pi_{0.5}$-LIBERO motor policy inside our agent.}
\label{tab:liberopro_full}
\centering
\small
\begin{tabular}{@{}lccccccc@{}}
\toprule
\textbf{Method / Reasoner} & \textbf{Obj.} & \textbf{Obj.} & \textbf{Sp.} & \textbf{Sp.} & \textbf{Goal} & \textbf{Goal} & \textbf{Mean} \\
 & \textbf{swap} & \textbf{task} & \textbf{swap} & \textbf{task} & \textbf{swap} & \textbf{task} & \\
\midrule
$\pi_0$~\cite{pi0}                 & \num{0}  & \num{0}  & \num{0}  & \num{0}  & \num{0}  & \num{0}  & \num{0}  \\
$\pi_{0.5}$~\cite{pi05}           & \num{17} & \num{1}  & \num{20} & \num{1}  & \num{38} & \num{0}  & \num{12.8} \\
CaP-Agent0~\cite{capx}             & \num{22} & \num{18} & \num{12} & \num{14} & \num{26} & \num{17} & \num{18.2}   \\
\midrule
\textbf{\methodname{} (Ours)} \\
\; GPT-5.5 low            & \num{44} & \num{26} & \num{52} & \num{74} & \num{42} & \num{28} & \num{44.3} \\
\; GPT-5.5 med            & \num{48} & \num{38} & \num{60} & \num{64} & \num{44} & \num{24} & \num{46.3} \\
\; GPT-5.5 high           & \num{60} & \num{36} & \num{60} & \num{74} & \num{38} & \num{26} & \num{49.0} \\
\; Gemini Rob-ER 1.6  & \num{58} & \num{38} & \num{50} & \num{64} & \num{44} & \num{34} & \num{48.0} \\
\; Gemini 3.5 Flash       & \num{60} & \num{32} & \num{58} & \num{62} & \num{48} & \num{28} & \num{48.0} \\
\; Gemini 3.1 Pro         & \num{64} & \num{34} & \num{60} & \num{68} & \num{44} & \num{22} & \num{48.7} \\
\; Claude Haiku 4.5       & \num{56} & \num{40} & \num{54} & \num{78} & \num{38} & \num{20} & \num{47.7} \\
\; Claude Sonnet 4.6      & \num{54} & \num{44} & \num{62} & \num{78} & \num{42} & \num{30} & \num{51.7} \\
\; \textbf{Claude Opus 4.7} & \num{54} & \num{54} & \num{66} & \num{80} & \num{44} & \num{22} & \textbf{\num{53.3}} \\
\bottomrule
\end{tabular}
\end{table}

\section{Tool Schemas}
\label{app:schemas}

The JSON schemas exposed to the VLM reasoner are reproduced below.

\begin{lstlisting}[language=json]
[
  {
    "name": "Perceive",
    "description": "Capture the wrist-camera image and robot state (end-effector pose, gripper aperture, is_grasped) and return the set of detected object labels. Perceive does NOT return the third-person camera; that view is only exposed during the blind phase via LookBack(verify_only=true), which guarantees the agent cannot peek at an external scene change.",
    "input_schema": {"type": "object", "properties": {}, "required": []}
  },
  {
    "name": "Pick",
    "description": "Grasp a detected object using the TAMP grasping backend (detect -> segment -> depth -> M2T2 grasp -> cuRobo plan -> execute -> close gripper).",
    "input_schema": {
      "type": "object",
      "properties": {
        "label": {"type": "string", "description": "An exact object label from the most recent Perceive."}
      },
      "required": ["label"]
    }
  },
  {
    "name": "DropAbove",
    "description": "Place the currently held object. Either target_label OR all three of (abs_x, abs_y, abs_z) MUST be provided. RELATIVE mode: pass target_label (+ optional dx_m, dy_m) and the drop point is reference_centroid + offset. ABSOLUTE mode: pass abs_x, abs_y, abs_z and the drop point is exactly that world coordinate (target_label is ignored).",
    "input_schema": {
      "type": "object",
      "properties": {
        "target_label": {"type": "string", "description": "Required in RELATIVE mode."},
        "dx_m": {"type": "number"},
        "dy_m": {"type": "number"},
        "abs_x": {"type": "number", "description": "Required (with abs_y, abs_z) in ABSOLUTE mode."},
        "abs_y": {"type": "number"},
        "abs_z": {"type": "number"}
      },
      "required": []
    }
  },
  {
    "name": "VLARollout",
    "description": "Execute a physical subgoal using the pi0.5 visual policy.",
    "input_schema": {
      "type": "object",
      "properties": {
        "subgoal": {"type": "string", "description": "A short concrete manipulation command, e.g. put the red cup on the plate."}
      },
      "required": ["subgoal"]
    }
  },
  {
    "name": "Release",
    "description": "Open the gripper unconditionally (recovery if the wrong object was grasped).",
    "input_schema": {"type": "object", "properties": {}, "required": []}
  },
  {
    "name": "LookAway",
    "description": "Rotate the arm so the wrist camera points away from the table (base rotated 180 degrees) and hold the pose. Used in memory/restore tasks where the agent must not observe an external scene change.",
    "input_schema": {
      "type": "object",
      "properties": {"wait_s": {"type": "number"}},
      "required": []
    }
  },
  {
    "name": "LookBack",
    "description": "Two modes. verify_only=true: wait wait_s seconds, then return the third-person-camera image so the agent can check whether the camera is still covered. The wrist stays in the look-away pose. The agent polls this repeatedly until it sees the scene is uncovered, then calls verify_only=false to commit. verify_only=false: move the arm back to the capture pose.",
    "input_schema": {
      "type": "object",
      "properties": {
        "verify_only": {"type": "boolean"},
        "wait_s": {"type": "number"}
      },
      "required": []
    }
  },
  {
    "name": "Done",
    "description": "Signal that the task is complete. The agent must call Perceive immediately beforehand to verify the predicate against the current scene.",
    "input_schema": {"type": "object", "properties": {}, "required": []}
  }
]
\end{lstlisting}

The schemas above are the real-robot tool set: \textsc{Pick} and \textsc{DropAbove} are backed by the TAMP engine, and \textsc{VLARollout} is backed by the frozen $\pi_{0.5}$ policy. The LIBERO-PRO experiments use the seven simulator tools described in
Appendix~\ref{app:sim-tools}.

\subsection{LIBERO-PRO Tool Interface}
\label{app:sim-tools}

Pigey uses the following tools in LIBERO-PRO:

\begin{itemize}
    \item \textsc{Perceive}: returns annotated agent-view and wrist-camera
    images, robot state, and a numbered list of detected objects. It is the
    first action in every trial.

    \item \textsc{Grasp}: grasps a selected detected object using an analytic
    grasp controller.

    \item \textsc{Place}: places the currently held object either on a flat
    surface or inside a container.

    \item \textsc{VLARollout}: runs the frozen $\pi_{0.5}$-LIBERO policy on a
    short natural-language subgoal. The agent re-perceives afterward to check
    the result.

    \item \textsc{VerifyCandidate}: crops a candidate detection and asks a VLM
    whether it matches the intended target, returning
    \texttt{YES}, \texttt{NO}, or \texttt{UNSURE}.

    \item \textsc{GoHome}: returns the robot arm to its home pose without
    changing the object configuration.

    \item \textsc{Release}: opens the gripper unconditionally.
\end{itemize}

Only \textsc{VLARollout} invokes the learned motor policy; the remaining
manipulation tools use analytic controllers.

\section{Hardware and Software Specifics}
\label{app:hardware}

\paragraph{Real robot.}
We use a Franka Research 3 arm with a Robotiq 2F-85 parallel gripper, two side-mounted Zed-2i RGB cameras, and one Zed-Mini wrist camera. The robot is controlled through a DROID/polymetis-style stack at the robot's native control rate. The orchestrator runs on a developer workstation and communicates with the robot and VLA policy server over a JSON-line protocol. Each \textsc{VLARollout} executes for $K{=}\num{300}$ control steps.

\paragraph{Simulation.}
We evaluate in LIBERO-PRO using $\pi_{0.5}$-LIBERO as the motor substrate. Standard per-suite horizons are used for spatial, object, and goal suites. Each \textsc{VLARollout} runs the simulator faster than realtime.

\paragraph{Reasoner integration.}
VLM calls use each provider's standard API. Tool calls are returned as structured JSON. Per-trial reasoner usage is \num{3}--\num{15} calls depending on task length and number of failures.

\paragraph{Compute and software stack.}
The real-robot stack runs on two machines. A control NUC drives the robot through polymetis (zerorpc on port \num{4242}, gRPC on \num{50051}) under the \texttt{PREEMPT\_RT} kernel \texttt{6.8.0-rt8}, which polymetis requires for deterministic real-time control. A workstation (AMD Ryzen~7 9800X3D, 32\,GB RAM, NVIDIA GeForce RTX~5090 32\,GB, Ubuntu 24.04 LTS) hosts the perception and policy servers: the TiPToP grounding and grasp pipeline (open-vocabulary detection with Gemini Robotics-ER, segmentation with SAM\,2, stereo depth with FoundationStereo, grasp synthesis with M2T2, and motion planning with cuRobo/cuTAMP), the $\pi_{0.5}$-DROID VLA policy server, and the \methodname{} orchestrator process. The two GPU-resident components dominate VRAM ($\pi_{0.5}$-DROID inference $\approx\num{16}$\,GB and FoundationStereo $\approx\num{8}$\,GB) and fit on the single RTX~5090. Simulation (LIBERO-PRO with $\pi_{0.5}$-LIBERO) runs on a multi-GPU node equipped with NVIDIA B300 SXM6 GPUs (288\,GB HBM each); one $\pi_{0.5}$-LIBERO server per GPU, with the six LIBERO perturbation suites running in parallel.

\section{Prompt Template}
\label{app:prompt}
The system prompt used in our real-robot experiments is structured around the following blocks; we summarize each below.

\paragraph{Tool definitions.} The prompt lists each tool, its input schema, and its return shape, including which fields are programmatic guarantees (e.g.\ \texttt{is\_grasped} on \textsc{Pick}) versus what must be verified visually.
\paragraph{Decision-order routing rule.} The agent is instructed to (i)~always \textsc{Perceive} first, (ii)~prefer the TAMP path (\textsc{Pick}+\textsc{DropAbove}) for clean rigid-body pick-and-place, (iii)~escalate to \textsc{VLARollout} after repeated \textsc{Pick} failures, when the target is rimmed/in-container, or for deformables / non-grasp verbs, and (iv)~call \textsc{Done} only after a final \textsc{Perceive}, from which it must judge whether the task predicate holds. The pre-\textsc{Done} \textsc{Perceive} is a hard, system-level invariant --- the orchestrator rejects any \textsc{Done} whose immediate predecessor is not \textsc{Perceive} --- so a fresh observation is always taken before termination; the predicate judgment from that observation is the agent's own, not a guarantee enforced by the system.
\paragraph{$\pi_{0.5}$ language conventions.} \textsc{VLARollout} subgoals must stay close to the policy's training distribution: short, concrete commands such as \emph{``pick up the red cup''} or \emph{``put the cup in the basket''}; descriptors or hedges degrade the policy.
\paragraph{General resolution strategies.} The prompt supplies a small set of task-agnostic strategies that the agent invokes from the \emph{structure} of an instruction rather than from any specific task; each is stated in task-agnostic terms rather than tied to a particular benchmark task:
\begin{itemize}
  \item \emph{Existential vs.\ universal} reading of the task verb (act on one matching object vs.\ all of them).
  \item \emph{Superlative-constrained selection}: rank candidates by the superlative dimension and try them in order until one satisfies the constraint (e.g.\ ``the smallest X that also satisfies Y'').
  \item \emph{Occlusion search}: if no detected label matches the target, treat the most likely occluder as a barrier, move it aside, re-perceive, then act on the revealed object.
  \item \emph{Absolute-coordinate restoration}: record each object's world coordinates, then return each object to its recorded coordinate. During the perturbation the agent is \emph{physically} prevented from observing the change --- \textsc{LookAway} rotates the wrist away from the table ($\sim$180\textdegree\ at the base) and \textsc{Perceive} returns only the wrist view, so neither tool can leak the new scene; the third-person camera is exposed only through \textsc{LookBack}(\texttt{verify\_only=true}), which the agent uses to poll for end-of-perturbation.
  \item \emph{Sequential imitation}: maintain an append-only numbered log of observed single-object moves, and once the demonstration ends, replay the logged moves in order.
\end{itemize}
\paragraph{Operating discipline.} Every output must end with a tool call, label arguments must be exact strings from the most recent \texttt{known\_objects}, and Gemini-ER labels can drift between perceives so the agent must always refer to the latest list. The exact-string rule prevents the detector from being primed by a hallucinated target name --- passing the task's free-text target to a grounding model often re-labels whatever salient object is present as the target, so the agent is required to choose only among labels the perception layer actually returned.
At a high level, the prompt does not attempt to teach the model how to manipulate; it constrains \emph{when} and \emph{which} primitive to invoke, and what to verify before declaring \textsc{Done}.

\section{Scoring Protocol}
\label{app:scoring}

Each trial is scored binary success/failure. For real-robot tasks, success is determined by a human evaluator using the task specification and final scene state. For simulation tasks, success is determined by the LIBERO predicate. A trial that times out without \textsc{Done} is a failure. A trial where \textsc{Done} is emitted before the success condition is satisfied is also a failure.

Each real-robot task is evaluated with \num{5} trials under matched initial scenes. Each LIBERO-PRO task is evaluated with \num{10} trials per reasoner, constrained by frontier-model API cost across the \num{9}-reasoner sweep ($\approx\num{5400}$ total LIBERO trials). The cross-reasoner sweep serves as implicit replication: the consistency of the gain across all seven reasoners (Table~\ref{tab:liberopro_full}) supplements cross-model agreement.

\section{Cost Accounting}
\label{app:cost}

Each real-robot orchestrated trial uses \num{3}--\num{15} reasoner calls depending on task length and number of failures. Approximate API cost ranges from \$\num{0.02} to \$\num{0.50} per trial depending on model tier. Wall-clock time is approximately \num{2}--\num{6} minutes per real-robot trial. Simulation trials run faster than real time.
}

\end{document}